\documentclass[lettersize,journal]{IEEEtran}
\usepackage{amsmath,amsfonts}
\usepackage{algorithmic}
\usepackage{algorithm}
\usepackage{array}
\usepackage[caption=false,font=normalsize,labelfont=sf,textfont=sf]{subfig}
\usepackage{textcomp}
\usepackage{stfloats}
\usepackage{url}
\usepackage{verbatim}
\usepackage{graphicx}
\usepackage{cite}
\usepackage{capt-of}
\usepackage{makecell}
\usepackage{adjustbox}
\usepackage{nicematrix,tikz}
\usepackage{arydshln}
\usepackage{booktabs}
\usepackage{multirow}
\usepackage{cuted}
\usepackage{tabularray}
\usepackage{xspace}
\usepackage{colortbl}
\usepackage{newfloat}
\usepackage{listings}
\usepackage{eso-pic}
\usepackage[breaklinks=true,colorlinks,
citecolor=cyan,linkcolor=red,
bookmarks=false]{hyperref}

\newcommand{\Tref}[1]{Table~\ref{#1}}
\newcommand{\eref}[1]{Eq.~\eqref{#1}}

\newcommand{\fref}[1]{Fig.~\ref{#1}}
\newcommand{\Fref}[1]{Figure~\ref{#1}}

\newcounter{todos}
\AtEndDocument{\ifnum\value{todos}>0 \PackageWarning{TODOS}{There are \arabic{todos} todos left in this paper! Fix them before submitting the paper!} \fi}

\newcommand{\concat}[1]{\mathop{\mathrm{concat}}\left(#1\right)}

\makeatletter
\DeclareRobustCommand\onedot{\futurelet\@let@token\@onedot}
\def\@onedot{\ifx\@let@token.\else.\null\fi\xspace}

\def\wrt{w.r.t\onedot} 
\def\etal{\emph{et al}\onedot}
\makeatother

\newcommand{\unips}{UniPS~\cite{ikehata2022unips}\xspace}
\newcommand{\sdmunips}{SDM-UniPS~\cite{ikehata2023sdmunips}\xspace}
\newcommand{\stablenormal}{StableNormal~\cite{Ye2024StableNormalRD}\xspace}
\newcommand{\geowizard}{GeoWizard~\cite{fu2024geowizard}\xspace}
\newcommand{\mvm}{MV20~\cite{2020Two}\xspace}
\newcommand{\lxm}{LX18~\cite{Li2018LearningTR}\xspace}
\newcommand{\ee}{E2E-FT~\cite{martingarcia2024diffusione2eft}\xspace}
\newcommand{\me}{Metric3Dv2~\cite{hu2024metric3dv2}\xspace}
\newcommand{\ours}{FlashNormal\xspace}
\newcommand{\ourdata}{EvalFlash\xspace}
\newcommand{\traindata}{Flash100K\xspace}
\newcommand{\sd}{SD~\cite{Rombach_2022_CVPR}\xspace}

\newcommand{\fnf}{flash/no-flash images}

\definecolor{revisionyellow}{RGB}{176,124,0}

\begin{document}

\AddToShipoutPictureFG*{%
	\AtPageLowerLeft{%
		\raisebox{0.16in}{%
			\makebox[\paperwidth]{%
				\parbox{0.92\paperwidth}{\centering\fontsize{6}{7}\selectfont
					Copyright \textcopyright{} 2026 IEEE. Personal use of this material is permitted. However, permission to use this material for any other purposes must be obtained from the IEEE by sending an email to pubs-permissions@ieee.org.}%
			}%
		}%
	}%
}

\title{FlashNormal: Detailed Surface Normal Estimation from Flash and No-Flash Images}

\author {Ruiyang Chen\textsuperscript{$\dagger$},
	Feiran Li\textsuperscript{$\dagger$},
	Heng Guo\textsuperscript{$\ast$}, 
	and Zhanyu Ma,~\IEEEmembership{Senior Member,~IEEE} 
\IEEEcompsocitemizethanks{\IEEEcompsocthanksitem Ruiyang Chen, Heng Guo, and Zhanyu Ma are with the Pattern Recognition and Intelligent System Laboratory, School of Artificial Intelligence, Beijing University of Posts and Telecommunications (BUPT), Beijing 100080, China (e-mail: \{chenruiyang, guoheng, mazhanyu\}@bupt.edu.cn). }
\IEEEcompsocitemizethanks{\IEEEcompsocthanksitem Feiran Li is independent researcher (e-mail: frl1994@hotmail.com).}
\thanks{\textsuperscript{$\dagger$}These authors contributed equally to this work. \textsuperscript{$\ast$}Corresponding author.}
}

\markboth{IEEE TRANSACTIONS ON CIRCUITS AND SYSTEMS FOR VIDEO TECHNOLOGY}%
{Shell \MakeLowercase{\textit{et al.}}: A Sample Article Using IEEEtran.cls for IEEE Journals}
\maketitle
\begin{strip}
	\begin{adjustbox}{max width=\linewidth}
		\addtolength{\tabcolsep}{-0.4em}
		\begin{NiceTabular}{cccccc}
			\makecell{Flash image} & 
			\makecell{No-flash image} & 
			\makecell{FlashNormal (Ours) \\surface normal} & 
			\makecell{FlashNormal (Ours) \\integrated mesh} & 
			\makecell{\ee \\surface normal} &
			\makecell{\ee \\integrated mesh} \\
			
			\includegraphics[width=0.2\linewidth]{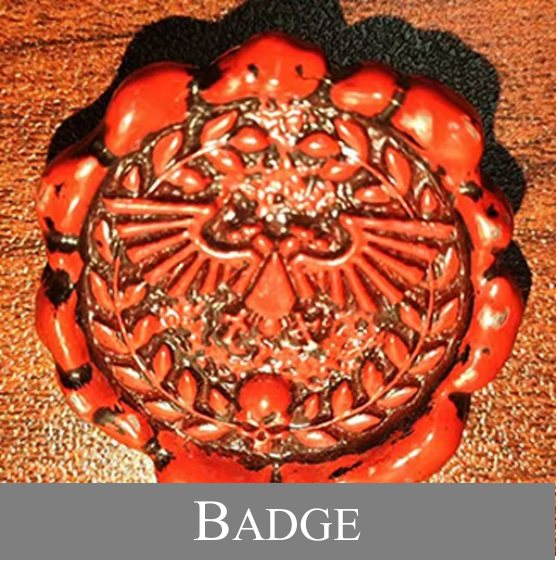} & 
			\includegraphics[width=0.2\linewidth]{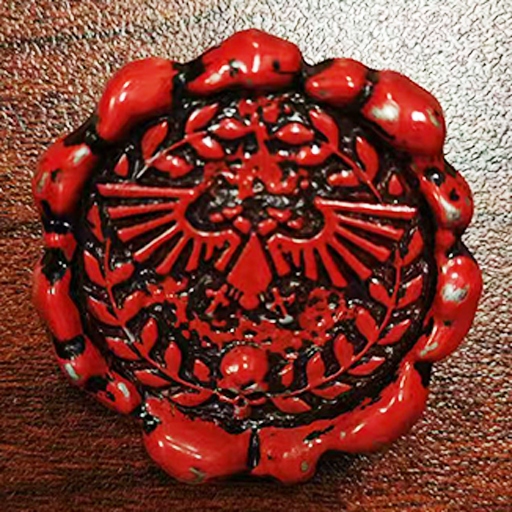}	&
			\includegraphics[width=0.2\linewidth]{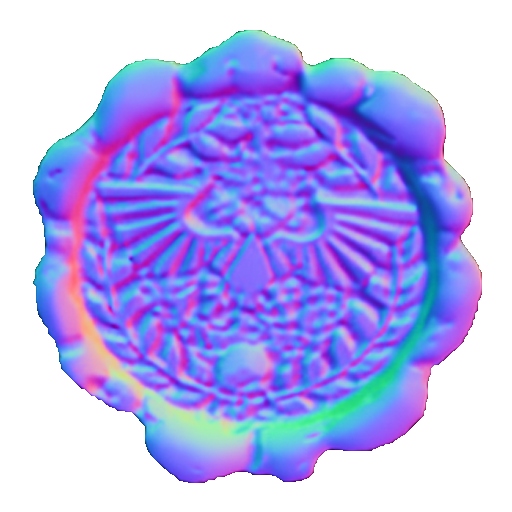} &
			\includegraphics[width=0.2\linewidth]{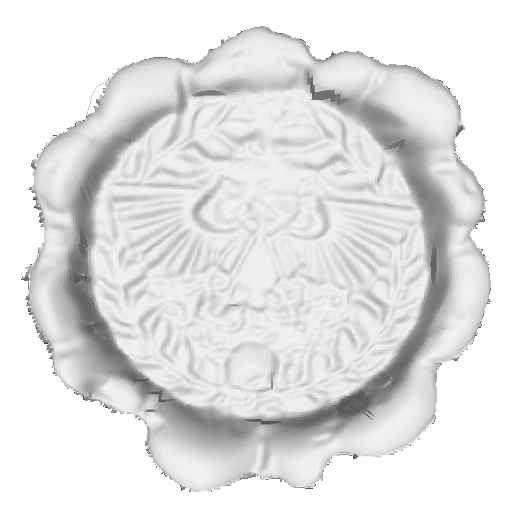} &
			\includegraphics[width=0.2\linewidth]{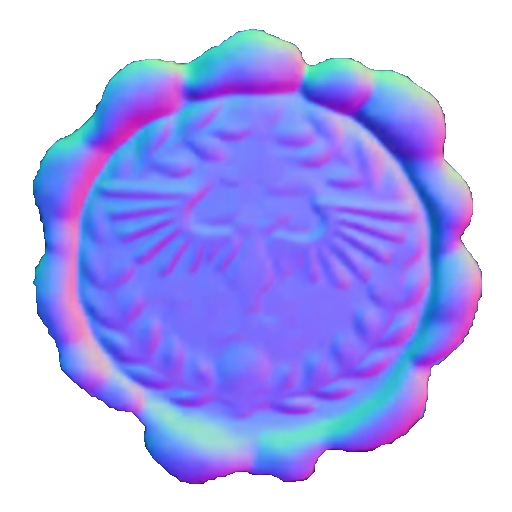} &
			\includegraphics[width=0.2\linewidth]{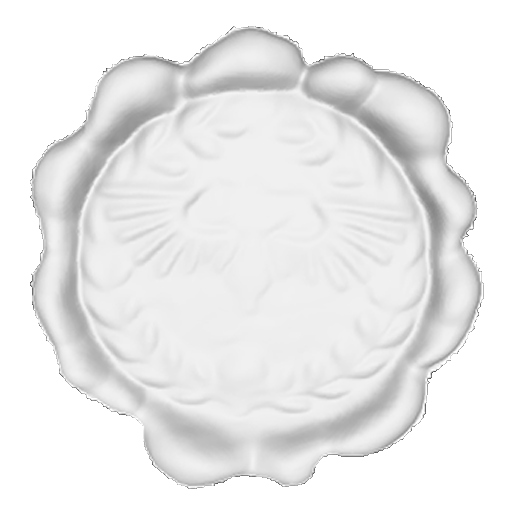} \\
			
			\includegraphics[width=0.2\linewidth]{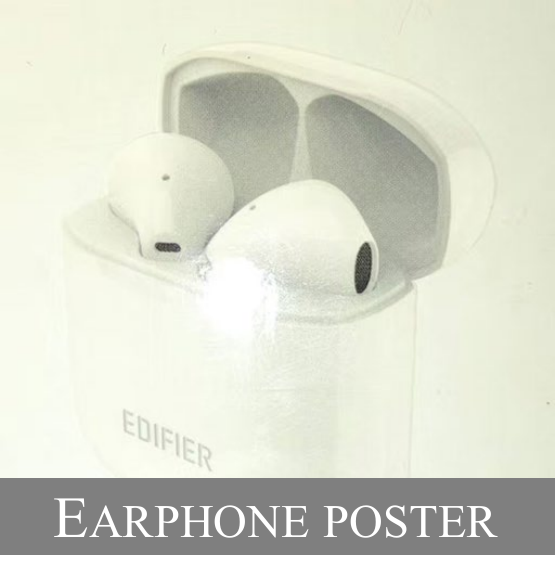} & 
			\includegraphics[width=0.2\linewidth]{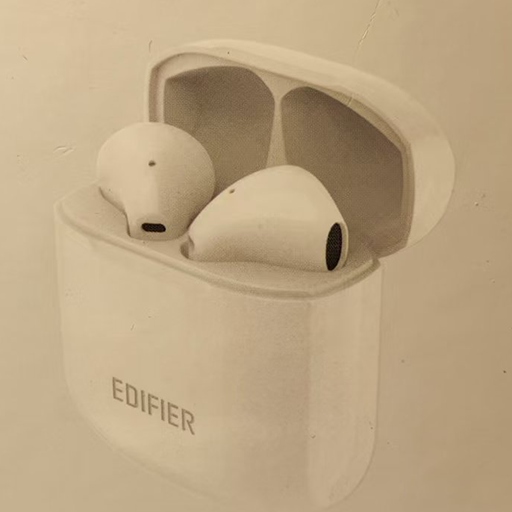} &
			\includegraphics[width=0.2\linewidth]{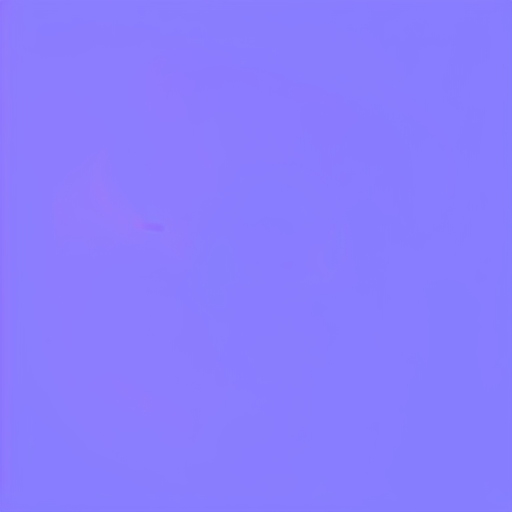} &
			\includegraphics[width=0.2\linewidth]{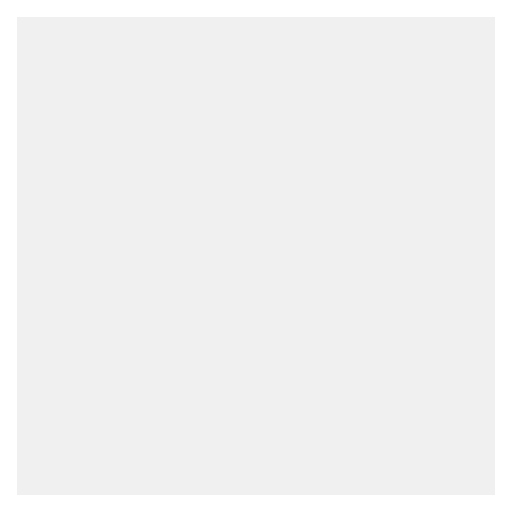} &
			\includegraphics[width=0.2\linewidth]{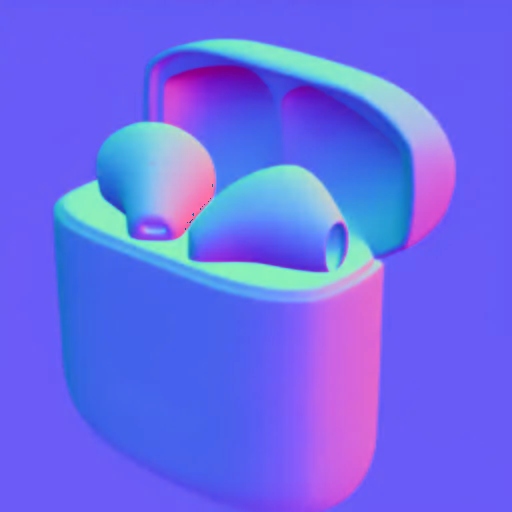} &
			\includegraphics[width=0.2\linewidth]{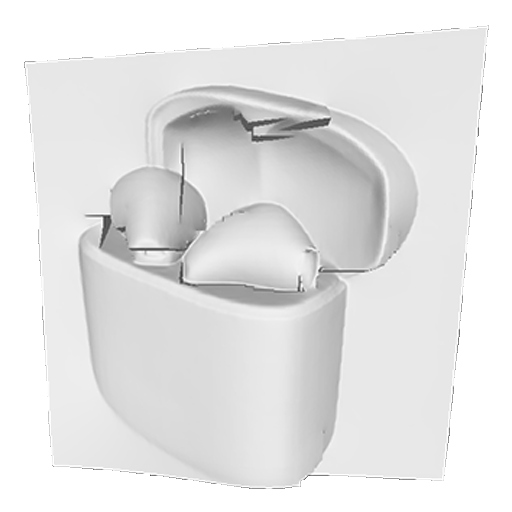} \\
		\end{NiceTabular}
	\end{adjustbox}
	\captionof{figure}{We propose \ours, a diffusion-based surface normal estimator taking flash/no-flash image pairs as input. Compared with single image-based normal estimation method~\cite{martingarcia2024diffusione2eft}, FlashNormal can recover high-quality surface normals with rich details, while reducing the shape-reflectance ambiguity as shown in the case of {\sc Earphone poster}, whose actual shape is a plane.}
	\label{teaser}
\end{strip}
\begin{abstract}

High-quality surface normal estimation is preferred for detailed surface shape recovery and image editing. Existing single image-based methods, though being a practical setup, often struggle to recover fine surface details and are sensitive to inherent shape-reflectance ambiguity. While photometric stereo achieves high-fidelity surface normal estimation from images under varying lights, its applicability is strictly limited by requiring a multi-illumination capture setup. To this end, we propose \ours, a diffusion-based surface normal estimator from flash/no-flash image pairs. While retaining high practicability on modern smartphones, our proposal takes advantage of flash-induced shading variations, and leverages curvature-guided detail enhancement strategy, improving surface detail recovery and mitigating shape-reflectance ambiguity effectively. To evaluate our proposed method, we further present \ourdata, the first real-world flash/no-flash evaluation dataset containing $20$ objects aligned with ground-truth surface normals for quantitative benchmarking. Extensive experiments demonstrate the effectiveness of \ours over state-of-the-art single image-based methods and show a significant out-performance over flash/no-flash-based normal estimation method on \ourdata.

\end{abstract}
  
\section{Introduction}

\IEEEPARstart{S}{urface} normal maps represent a $2.5$-dimensional expression of geometry. High-quality surface normals are required by various computer vision tasks, such as object relighting, augmented reality applications, and advanced rendering techniques~\cite{kuang2022neroic, supernormal2024cao, wang2022neuris, Yu2022MonoSDF, zhu2025relighting, ding2024ray, zhang2025gbr, bao20253d}. However, producing detail-rich surface normal maps usually requires either a 3D scanning process with high-accuracy scanners~\cite{shi2016benchmark}, or the collection of light-controlled images for photometric stereo~\cite{ikehata2018cnn,chen2018ps,ikehata2022unips,ikehata2023sdmunips}, both of which can be cumbersome and resource-intensive.

As a simpler alternative, recent advances in learning-based surface normal estimation~\cite{fu2024geowizard, Ye2024StableNormalRD, yin2023metric3d, qiu2023richdreamer, long2024wonder3d} have demonstrated the capability of surface normal estimation from a single RGB image. Despite their popularity, these methods struggle to handle geometric details and may produce ambiguous shape recoveries, as a single image input cannot effectively distinguish between geometric and texture details due to the inherent shape-reflectance ambiguity. For example, as shown in the top row of \fref{teaser}, the geometric details of estimated surface normals are insufficient and over-smoothed. Also, the bottom row of \fref{teaser} shows that a single image alone cannot provide enough information to determine if the depicted scene is a truly 3D shape or merely a 2D plane.

In this work, we introduce \ours, a practical and effective approach for surface normal estimation. Drawing inspiration from photometric stereo—where shading differences across images under varying lighting help resolve ambiguities between shape and reflectance—we utilize flash/no-flash image pairs as a minimal yet accessible source of shading cues to reduce shape estimation ambiguity. Unlike prior two-shot surface normal estimation methods~\cite{2020Two,cao2020stereoscopic}, our method further benefits from the zero-shot generalization capabilities of diffusion priors~\cite{Rombach_2022_CVPR}, which help alleviate shape-reflectance ambiguity and improve estimation accuracy.

In addition to addressing shape-reflectance ambiguity, we further enhance geometric details in surface normal estimation. Specifically, we introduce a curvature-guided detail enhancement strategy that directs more focus to regions with pronounced geometric features during training. Moreover, inspired by techniques from small object detection~\cite{2021AdaZoom,2019autofocus}, we propose a zoomed-pixels strategy to refine fine-scale surface details. Together, these two strategies effectively contribute to producing detail-rich surface normal maps.

To evaluate the effectiveness of \ours in real-world scenarios, we introduce \ourdata, the first real-world flash/no-flash image dataset with labeled ground-truth~(GT) surface normals. \ourdata comprises $20$ objects with diverse shapes and materials, where object meshes are scanned and carefully aligned with the captured views to generate high-quality GT surface normal maps, allowing for quantitative evaluation in real-world scenes. 

To summarize, our main contributions are as follows:
\begin{itemize}
\item We propose \ours, a diffusion-based surface normal estimator taking in flash/no-flash images, boosting the shape recovery accuracy by reducing shape-reflectance ambiguity while maintaining a practical setup;
\item We introduce curvature-guided detail enhancement strategy and zoomed-pixels strategy, shown to be effective in improving geometric details of surface normal estimates;
\item Building upon precise 3D scanning and manual alignment, we build a real-world evaluation dataset named \ourdata. To the best of our knowledge, this is the first dataset for benchmarking flash/no-flash-based surface normal estimation with the corresponding GT surface normal maps.
\end{itemize} 

\section{Related Works}
In this section, we briefly review recent advances in single-view normal estimation based on their settings.

\paragraph{Surface normal estimation from single image} 
Surface normal estimation from a single image is an ill-posed problem due to shape-reflectance ambiguity. Both Li~\etal~\cite{Li2018LearningTR} and Sang~\etal~\cite{Sang_Chandraker_2020} adopt CNN-based cascade networks to recover spatially-varying BRDFs (SVBRDF) and surface normals from a flash image. Tiwari~\etal~\cite{tiwari2024merlinsingleshotmaterialestimation} further introduce an attention-based hourglass network to refine the quality of surface normals and reflectance. Most recent works explore Stable Diffusion~(SD)~\cite{Rombach_2022_CVPR} prior in surface normal estimation. Specifically, \geowizard fine-tunes \sd model on a large-scale dataset with paired depth, normal, and images, where a geometry switcher is applied to address scene-level and object-level inputs. Since surface normal estimation is a deterministic task, GenPercept~\cite{xu2024diffusion} and \ee reduce the inherent stochasticity of diffusion models by single-step denoising process. \stablenormal further introduces a semantic-guided refinement process to improve the accuracy and sharpness of estimated surface normals. Metric3Dv2~\cite{hu2024metric3dv2} brings a versatile geometric foundation model for depth and surface normal estimation trained on $16M$ indoor and outdoor scenes. Despite the practical setting of single-shot surface normal estimation, the above methods still struggle to resolve the inherent shape-reflectance ambiguity, leading to unsatisfied shape recovery accuracy and blurry surface normal estimation.

\paragraph{Surface normal estimation with photometric stereo} 
Photometric stereo is well-posed for surface normal estimation given images captured under varying illuminations of a Lambertian surface~\cite{woodham1980photometric, ju2023estimating, luo2014accurate, li2023neural, meng2025lac, ikehata2018cnn, chen2018ps, zheng2019spline}. To address general non-Lambertian reflectance, learning-based methods such as CNN-PS~\cite{ikehata2018cnn}, PS-FCN~\cite{chen2018ps}, SPLINE-Net~\cite{zheng2019spline} are proposed. However, these methods are restricted to darkroom settings due to their requirements for images taken under point lights. To tackle this limitation, Guo~\etal~\cite{guo2021patch} introduce a patch-based illumination strategy to extend the application range of photometric stereo to unknown environment light. \unips and \sdmunips, as the most recent photometric stereo methods, achieve surface normal estimation for general-reflectance surfaces under unknown natural illuminations. Despite the high-quality normal estimates from photometric stereo, multi-light images are required for the input, which limits their application range, especially for casual capture scenarios.
\begin{figure*}[!t]
	\begin{center}
	\includegraphics[width=\linewidth]{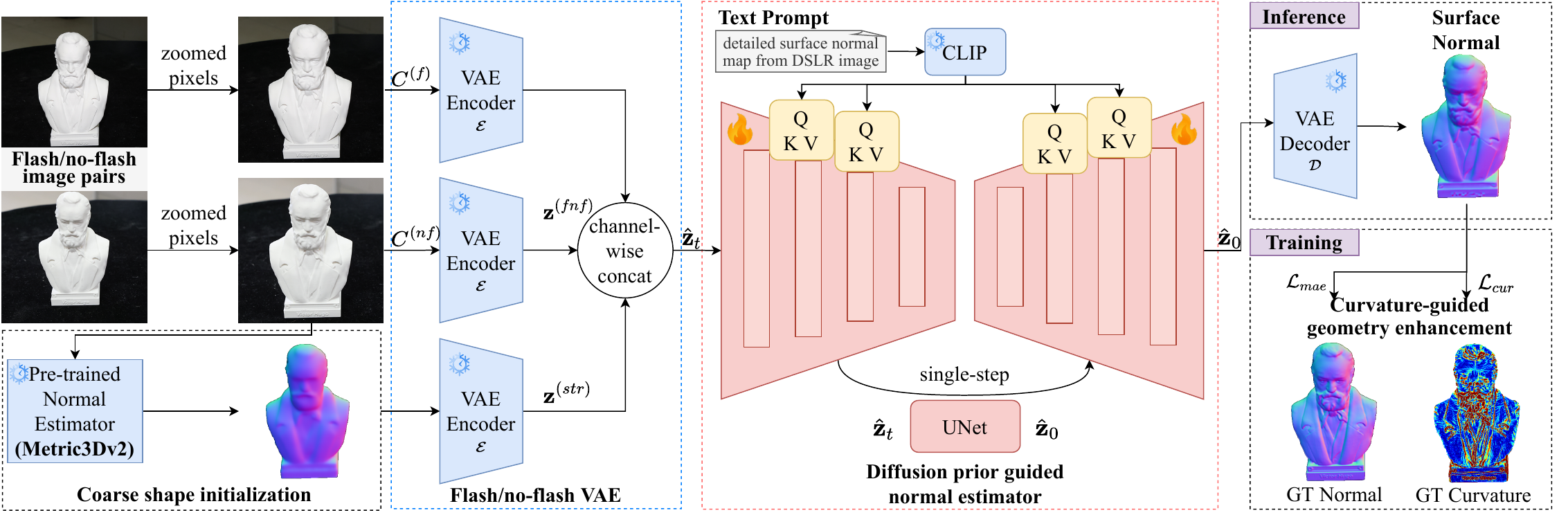}
	\caption{Pipeline of \ours: Flash/no-flash image pairs and a coarse shape initialization are individually encoded by the VAE and then concatenated into a latent representation, which serves as guidance for normal estimator. The normal estimator denoises the latent representation to generate the latent surface normal, which is then decoded to produce the final estimated surface normal map.}
	\label{pipeline}
	\vspace{-15pt}
\end{center}
\end{figure*}
\paragraph{Surface normal estimation from \fnf}
The flash/no-flash setup, as widely available on modern smartphones and cameras, can provide images under varying lighting conditions for reducing shape-reflectance ambiguity while retaining practicality. Cao~\etal~\cite{cao2020stereoscopic} estimate surface normals of Lambertian surfaces based on the image ratio between flash/no-flash pairs. However, their approach requires a depth map as input to resolve shape ambiguity. In follow-up work, Xia~\etal~\cite{darkflash} extend normal estimation to non-Lambertian surfaces via introducing a near-infrared (NIR) camera and NIR flash to capture additional cues. To avoid the need for depth or NIR inputs, Boss~\etal~\cite{2020Two} introduce a cascaded network and inverse-rendering loss to estimate shape, illumination, and SVBRDFs. Li~\etal~\cite{2023li} further demonstrate that an inverse rendering pipeline combined with the flash/no-flash setup can be used for translucent surface reconstruction. Despite these advances, surface normal estimation quality on real images remains limited due to the domain gap created by synthetic datasets used in training. In contrast, we show that the diffusion prior, capturing real-world data distribution, can achieve accurate surface normal recovery.

\section{\ours}
As illustrated in \fref{pipeline}, \ours comprises two key components: the flash/no-flash VAE and the diffusion prior-based surface normal estimator. The former takes flash/no-flash image pairs as input to extract shape variation information to address the shape-reflectance ambiguity. The latter denoises the latent representation obtained from the flash/no-flash VAE in a single-step, generating detailed surface normal maps enhanced by a curvature-guided detail enhancement strategy. Also, following the practice of \stablenormal, we takes a coarse initial surface normal map from \me as input, shown to be effective in accelerating convergence speed of the training process.

\subsection{Flash/no-flash VAE}
Flash/no-flash image pairs are inherently complementary, helping to enhance surface normal estimation accuracy. A straightforward approach to input these image pairs into network is to concatenate them at the original resolution into $6$-channel before applying VAE for compression. However, directly concatenating flash/no-flash image pairs as input could be problematic as pre-trained VAEs are generally trained on large-scale datasets of $3$-channel images, and fine-tuning them with $6$-channel inputs would lead to a mismatch with the pre-trained prior.

To address this problem, we employ a pre-trained VAE to encode the flash and no-flash images separately, yielding two $4$-channel latent space representations. Subsequently, we concatenate the flash and no-flash representations along the channel dimension within the latent space, resulting in an $8$-channel flash/no-flash guidance feature map:
\begin{equation}
    \mathbf{z}^{(fnf)} = \concat{\mathcal{E}(C^{(f)}), \mathcal{E}(C^{(nf)})},
    \label{eq:feature_concat}
\end{equation}
where $\mathbf{z}^{(fnf)}$ represents the latent flash/no-flash features, and $C^{(f)}$ and $C^{(nf)}$ denote the flash/no-flash image pairs, respectively. 

In addition to flash/no-flash image pairs, we incorporate an initial coarse surface normal map as input, obtained from an off-the-shelf efficient surface normal estimation method \me. Following the practice of \stablenormal, this coarse normal initialization could help accelerating the convergence of training process. We choose the no-flash image because it better matches the training distribution of Metric3Dv2~\cite{hu2024metric3dv2}, whereas using the flash image or averaging two predictions would introduce a stronger domain shift. To encode the coarse surface normal map, we utilize a pre-trained VAE to obtain a latent representation $\mathbf{z}^{(str)}$. We then combine it with $\mathbf{z}^{(fnf)}$ to form the final feature map $\hat{\mathbf{z}}_t$ that is input into the subsequently network:
\begin{equation}
    \hat{\mathbf{z}}_t = \concat{\mathbf{z}^{(fnf)}, \mathbf{z}^{(str)}}.
    \label{eq:feature_input}
\end{equation}

\subsection{Diffusion prior guided normal estimator}
We employ a Diffusion U-Net to denoise the feature maps $\hat{\mathbf{z}}_t$ obtained in the previous step. We adopt a deterministic single-step diffusion strategy in our network design. This approach reduces randomness during the denoising process, making it more suitable for the deterministic task of surface normal estimation, and leads to more stable and reliable outputs. Since the single-step generation process is independent of the number of time steps $t$, we set it to $999$. The generation process is guided by the text prompt ``detailed surface normal map from Digital Single-Lens Reflex~(DSLR) image". During training, the input to the Diffusion U-Net $f_\theta$ is $\hat{\mathbf{z}}_t$, and the network outputs the latent surface normal estimation $\hat{\mathbf{z}}_0$:
\begin{equation}
    \hat{\mathbf{z}}_0=-f_\theta(\hat{\mathbf{z}}_t,t=999).
\end{equation}
We adopt an end-to-end training strategy. A VAE decoder $\mathcal{D}$ is utilized to decode the latent representation $\hat{\mathbf{z}}_0$, producing the predicted surface normal in pixel space: $\quad\hat{\mathbf{y}}=\mathcal{D}\left(\hat{\mathbf{z}}_0\right)$. To enhance detail areas surface normal estimation, we employ two methods as follows.

\paragraph{Curvature-guided geometry enhancement} We propose a training strategy that leverages curvature information to enhance fine details in surface normals. Geometric details correspond to strong variations in normals, represented as curvature—higher in detailed regions and lower in smooth areas. By incorporating curvature maps into supervision, we put more attention on detail regions in surface normal estimation. As discussed in Heep~\etal~\cite{adaptive_screen_meshing}, curvatures are derived from the first and second fundamental forms:
\begin{equation}
\begin{gathered}
\label{eq:fun_form}
    \mathit{\boldsymbol{I}} _{ij}=\partial_i{\boldsymbol{x}}\cdot \partial_j{\boldsymbol{x}}, \\
    \mathit{\boldsymbol{I\!I}} _{ij}=-\partial_i{\boldsymbol{x}}\cdot \partial_j{\boldsymbol{n}},
\end{gathered}
\end{equation}
where $i,j$ denotes the uv coordinate in screen space. The first fundamental form measures the 3D distance between two points projected onto a 2D surface. The second fundamental form represents the gradient of the normal map. 
Given a set of surface normals, we can calculate $\partial_i{\boldsymbol{x}}$ since $\mathit{\boldsymbol{n}}\cdot\partial_i{\boldsymbol{x}}=0$. This gives:
\begin{align}
    \partial_u\mathit{\boldsymbol{x}}   &=  \mathit{\boldsymbol{e}}_x-\frac{\mathit{\boldsymbol{n}}_x}{\mathit{\boldsymbol{n}}_z}\cdot\mathit{\boldsymbol{e}}_z, &
    \partial_v\mathit{\boldsymbol{x}}   &=  \mathit{\boldsymbol{e}}_y-\frac{\mathit{\boldsymbol{n}}_y}{\mathit{\boldsymbol{n}}_z}\cdot\mathit{\boldsymbol{e}}_z, &
    \label{eq:pux}
\end{align}
where $\mathit{\boldsymbol{e}}_x, \mathit{\boldsymbol{e}}_y, \mathit{\boldsymbol{e}}_z$ are the unit vectors in the respective coordinate axis directions. Combining \eref{eq:fun_form} with \eref{eq:pux}, we derive the curvature $\kappa_i$ by solving the following generalized eigenvalue problem:
\begin{equation}
\label{eq:eigen}  
    \kappa_i\cdot \mathit{\boldsymbol{I}}\cdot\mathit{\boldsymbol{v}}_i=\mathit{\boldsymbol{I\!I}}\cdot\mathit{\boldsymbol{v}}_i.
\end{equation}

By combining equations~\eqref{eq:fun_form},~\eqref{eq:pux} and~\eqref{eq:eigen}, we generate a curvature map from the surface normal map to supervise geometric surface details. Specifically, we set an empirical threshold of $0.002$, identifying regions with curvature values exceeding this threshold as detail-rich areas. The corresponding detail mask $\mathcal{M}$ is then created to focus supervision on these regions. With this mask, we minimize the difference between curvature maps derived from GT and estimated surface normal maps in detail-rich areas:
\begin{equation}
    \mathcal{L}_{\mathrm{cur}}=\frac{1}{N_d} \sum_{i, j}\left ( \mathcal{M}_{i, j}\cdot(k_{i, j}^{*}-\hat{k}_{i, j})\right )^2,
\end{equation}
where $(i,j)$ denotes the pixel coordinates, and $k^{*}$ and $\hat{k}$ are the GT and predicted curvature map derived from their respective surface normal map. $N_d$ denotes the total number of valid pixels with a value of $1$ in the detail mask $\mathcal{M}$.

In addition to the curvature map loss, we supervise the network by minimizing the angular difference between the GT and predicted surface normals. Considering the detail-rich region dominant the angular error, leading to less attention to smooth and flat regions, we assign more weight to smooth surface normal areas. Specifically, the loss is designed as follows:
\begin{equation}
\begin{gathered}
\mathcal{L}_{\mathrm{mae}}=
\frac{1}{N_d} \sum_{i, j} \mathcal{M}_{i, j} \cdot\arccos \left(\frac{\boldsymbol{n}_{i, j}^{*} \cdot \hat{\boldsymbol{n}}_{i, j}}{\left\|\boldsymbol{n}_{i, j}^{*}\right\|\left\|\hat{\boldsymbol{n}}_{i, j}\right\|}\right)+ \\
10\cdot\frac{1}{N_s} \sum_{i, j} (1-\mathcal{M}_{i, j}) \cdot\arccos \left(\frac{\boldsymbol{n}_{i, j}^{*} \cdot \hat{\boldsymbol{n}}_{i, j}}{\left\|\boldsymbol{n}_{i, j}^{*}\right\|\left\|\hat{\boldsymbol{n}}_{i, j}\right\|}\right),
\end{gathered}
\end{equation}
where $\boldsymbol{n}^{*}$ and $\hat{\boldsymbol{n}}$ are the GT and predicted surface normal maps, respectively. $N_s$ denotes the total number of valid pixels with a value of $0$ in the detail mask $\mathcal{M}$. In this way, the detailed and smooth regions can be well recovered considering both the curvature loss and mean angular error loss. The training process is finalized by minimizing the following total loss function to optimize the parameters of the Diffusion U-Net:
\begin{equation}
\mathcal{L}=\mathcal{L}_{\mathrm{mae}}+\lambda\cdot\mathcal{L}_{\mathrm{cur}},
\end{equation}
where $\lambda$ denotes the weight of total loss. We empirically set it to $100$ in all of our experiments.

\paragraph{Zoomed-pixels for detail enhancement} We observe that the region of interest~(ROI) in the training and testing data could occupy varying image area. Small ROI leads to blurry surface normal estimation due to limited information. To handle this problem, we borrow the idea from object detection techniques, where upscaling small objects improves detection accuracy~\cite{2021AdaZoom,2019autofocus}. Specifically, we propose a zoomed-pixels strategy for detail enhancement. Specifically, we use bounding boxes to zoom in on regions of interest and resize them to $512 \times 512$, ensuring that objects occupy at least 80\% of the final input image. This operation is also applied during inference, enhancing input informativeness and strengthening the guidance from flash/no-flash pairs, leading to improved accuracy on surface normal estimation. At inference time, the ROI can be obtained automatically by first predicting an object mask using SAM~2~\cite{ravi2024sam2} or multimodal large models, and then converting the mask into a bounding box and valid region. This formulation can also be extended to multi-object scenes when the target regions are segmented reliably, while inaccurate masks may lead to inaccurate ROIs and weaken the zoomed-pixels strategy.

\subsection{Analyses regarding flash inputs}
We demonstrate that incorporating flash inputs helps to enhance the detail richness in estimated surface normals and mitigate shape-reflectance ambiguity.

\begin{figure}[!t]
\centering
\begin{adjustbox}{max width=\linewidth}
\addtolength{\tabcolsep}{-0.4em}
\newcommand{\imgsize}{0.3\textwidth}
\begin{NiceTabular}{cccccc}
    & No-flash input & Flash input && No-flash input & Flash input \\
    \raisebox{1.9\height}{\rotatebox{90}{\sc Badge}} &
    \includegraphics[width=\imgsize]{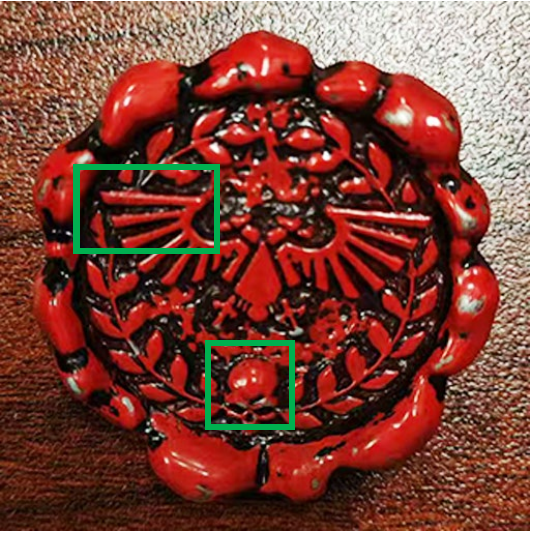} &
    \includegraphics[width=\imgsize]{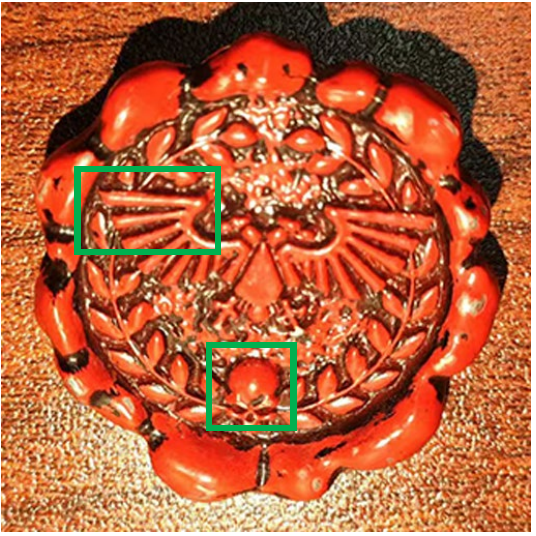} &
    \raisebox{0.4\height}{\rotatebox{90}{\sc Earphone poster}} &
    \includegraphics[width=\imgsize]{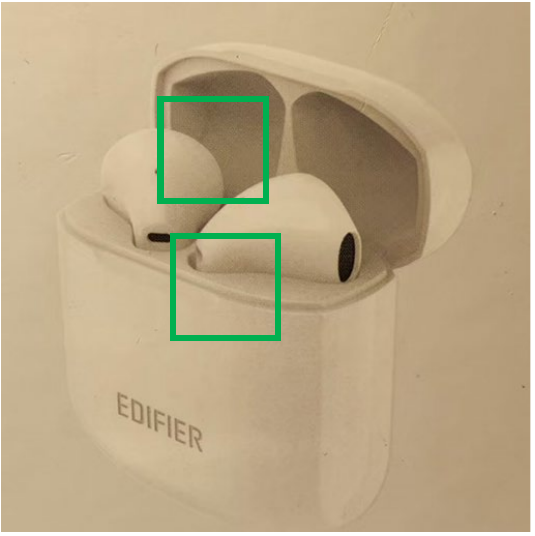} &
    \includegraphics[width=\imgsize]{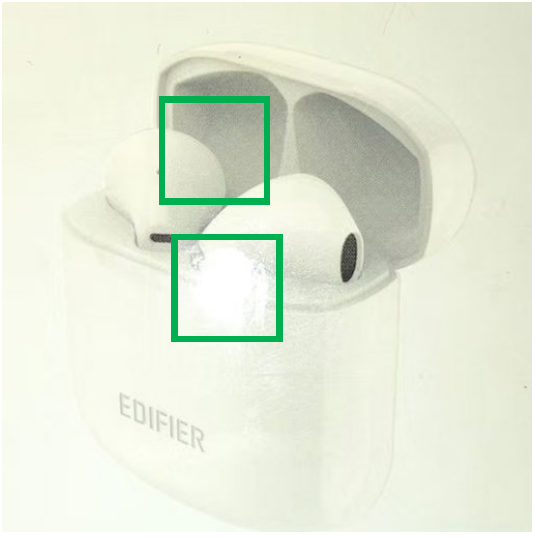} 
\end{NiceTabular}
\end{adjustbox}
\caption{Flash/no-flash image pairs of {\sc Badge} and {\sc Earphone poster}. We show the shading variation between flash and no-flash images through the green boxes. }
\label{fig:flash_input}
\vspace{-10pt}
\end{figure}

\paragraph{Flash inputs help to estimate detail-rich surface normals}
Incorporating flash aids in illuminating fine details of object surfaces, making the geometric structure more distinguishable. For instance, as shown on the left of \fref{fig:flash_input}, the no-flash image is affected by the randomness of ambient lighting, resulting in blurry fine details and obscured concave-convex surfaces. In contrast, the flash image enhances the visibility of these details by providing brighter and more uniform illumination. Furthermore, the shading variations between the flash and no-flash images facilitate a clearer justification of the underlying geometric structure.

\paragraph{Flash inputs help to reduce shape-reflectance ambiguity} 
In this paper, all ambiguity mentioned refers to shape-reflectance ambiguity. As illustrated on the right of \fref{fig:flash_input}, we analyze the appearance of {\sc Earphone poster} under flash and no-flash conditions. In the no-flash image, it is unclear whether the image depicts a real pair of earphones or merely a poster. However, with flash input, the flash halo is concentrated in one spot, and no shading variation is observed in other parts of the earphone. If it is an actual pair of earphones, there would be more diverse shading variations across their surface. This insight allows us to conclude that the object is a poster of earphones. In this way, the shading variation observed between flash and no-flash images provides strong hints to determine the true structure of objects, effectively reducing shape-reflectance ambiguity.

\section{Dataset}
We here introduce our rendered training dataset \traindata and a large-scale synthetic testset EvalFlash-synth for comprehensively evaluating the robustness of surface normal estimation methods. We also present the first-ever real-captured dataset \ourdata for benchmarking flash/no-flash surface normal estimation methods.

\newcommand{\largefigsize}{0.065}     %
\newcommand{\smallboxsize}{0.030}     %
\newcommand{\smallfigsize}{0.030}     %
\newcommand{\rheightsize}{0.55}       %
\newcommand{\vspacesize}{-3.5}        %
\newcommand{\hspacesize}{-3.5}        %

\begin{figure}[!bt] \centering
\addtolength{\tabcolsep}{-0.92em}
\newcommand{\imgsize}{0.3\textwidth}
\begin{NiceTabular}{cccccccccc}
    {\fontsize{7}{7}\selectfont \sc Fortune bag} &
    {\fontsize{7}{7}\selectfont \sc Pineapple}&
    {\fontsize{7}{7}\selectfont \sc Lady} &
    {\fontsize{7}{7}\selectfont \sc Pinkcat} &
    {\fontsize{7}{7}\selectfont \sc Guanyu}
    \\
    
    \includegraphics[width=\largefigsize\textwidth]{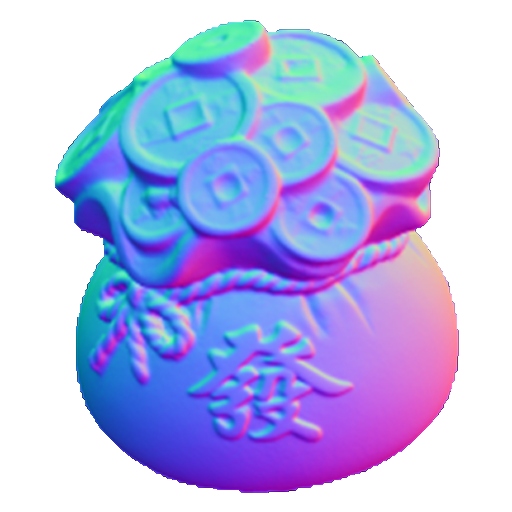}\hspace{\hspacesize pt}
    \raisebox{\rheightsize\height}{
        \makebox[\smallboxsize\textwidth]{
            \makecell{
                \includegraphics[width=\smallfigsize\textwidth]{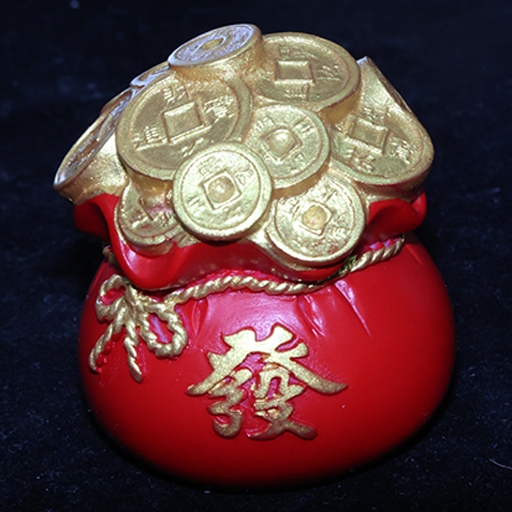}\\[\vspacesize pt]
                \includegraphics[width=\smallfigsize\textwidth]{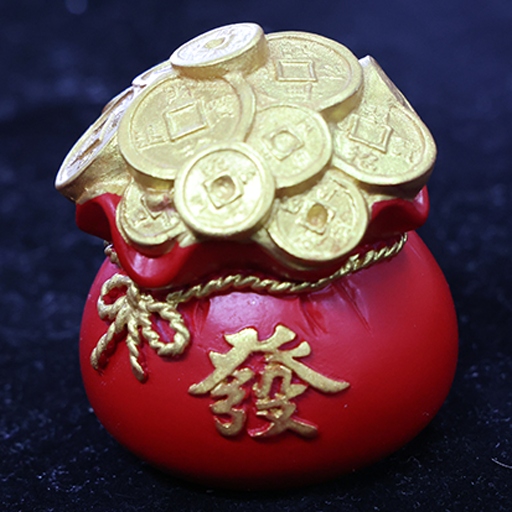}
            }
        }
    }&
    \includegraphics[width=\largefigsize\textwidth]{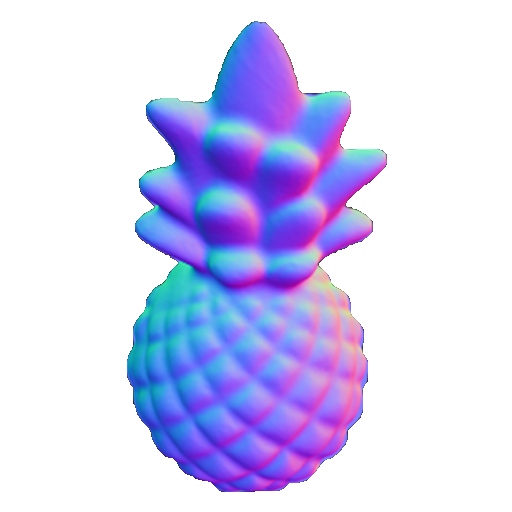}\hspace{\hspacesize pt}
    \raisebox{\rheightsize\height}{
        \makebox[\smallboxsize\textwidth]{
            \makecell{
                \includegraphics[width=\smallfigsize\textwidth]{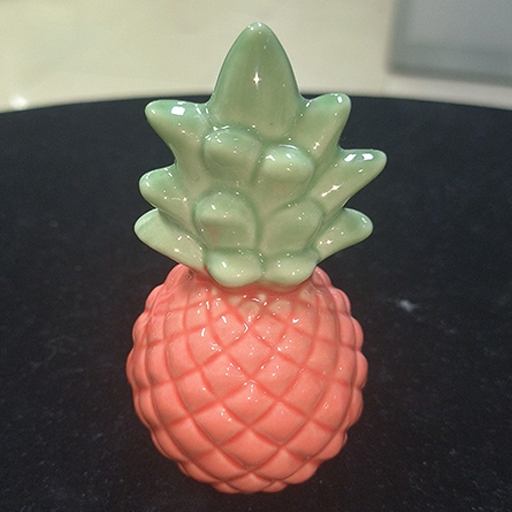}\\[\vspacesize pt]
                \includegraphics[width=\smallfigsize\textwidth]{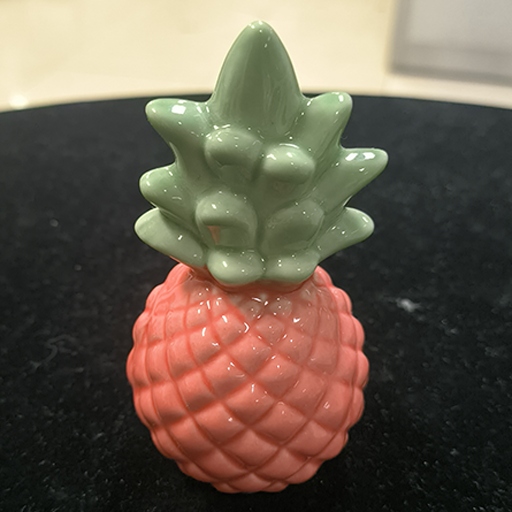}
            }
        }
    }&
    \includegraphics[width=\largefigsize\textwidth]{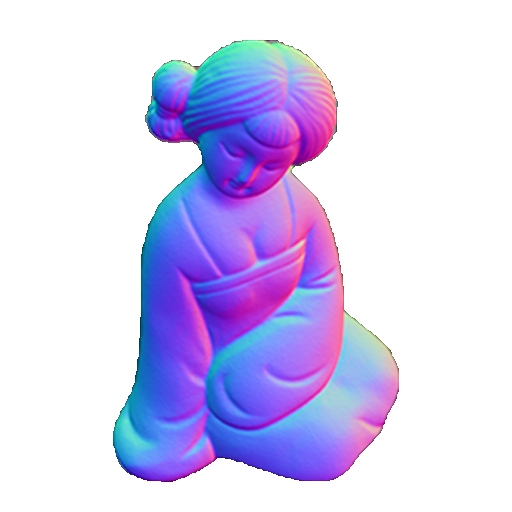}\hspace{\hspacesize pt}
    \raisebox{\rheightsize\height}{
        \makebox[\smallboxsize\textwidth]{
            \makecell{
                \includegraphics[width=\smallfigsize\textwidth]{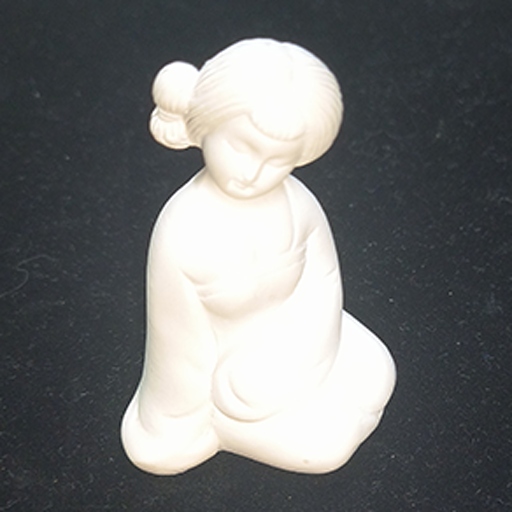}\\[\vspacesize pt]
                \includegraphics[width=\smallfigsize\textwidth]{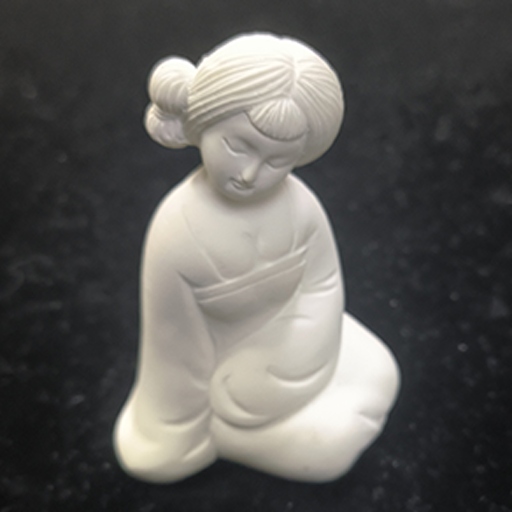}
            }
        }
    }&
    \includegraphics[width=\largefigsize\textwidth]{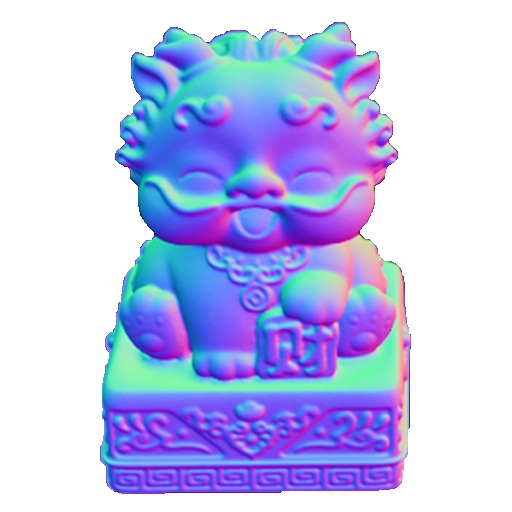}\hspace{\hspacesize pt}
    \raisebox{\rheightsize\height}{
        \makebox[\smallboxsize\textwidth]{
            \makecell{
                \includegraphics[width=\smallfigsize\textwidth]{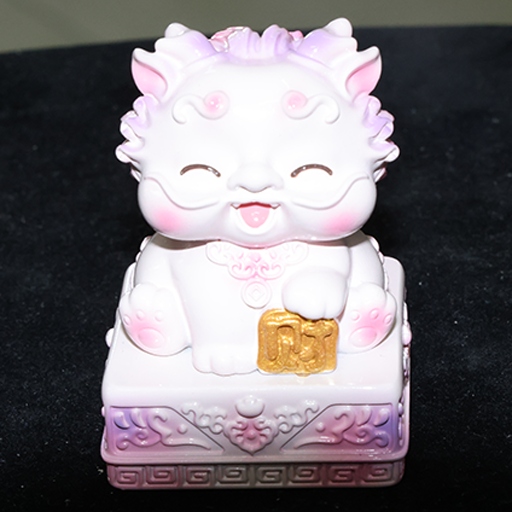}\\[\vspacesize pt]
                \includegraphics[width=\smallfigsize\textwidth]{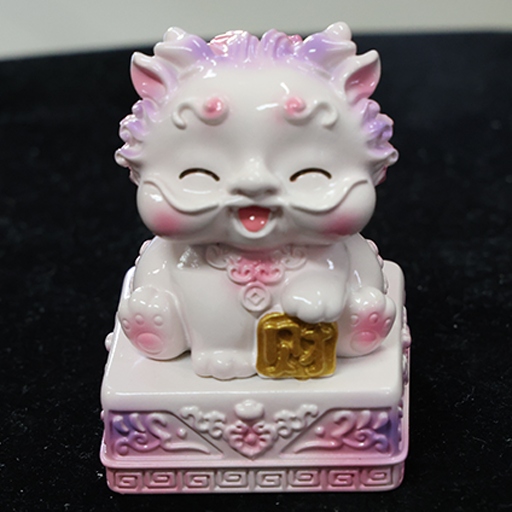}
            }
        }
    }&
    \includegraphics[width=\largefigsize\textwidth]{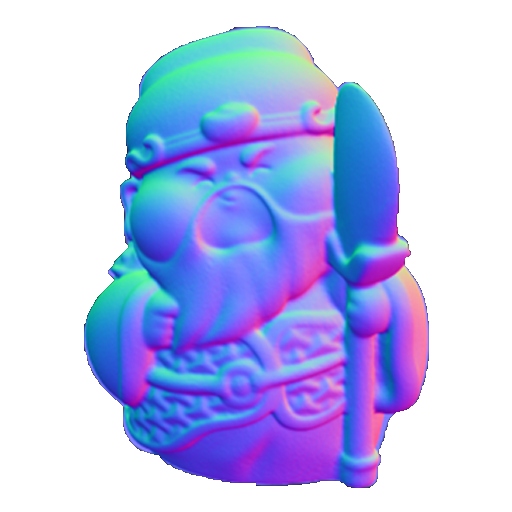}\hspace{\hspacesize pt}
    \raisebox{\rheightsize\height}{
        \makebox[\smallboxsize\textwidth]{
            \makecell{
                \includegraphics[width=\smallfigsize\textwidth]{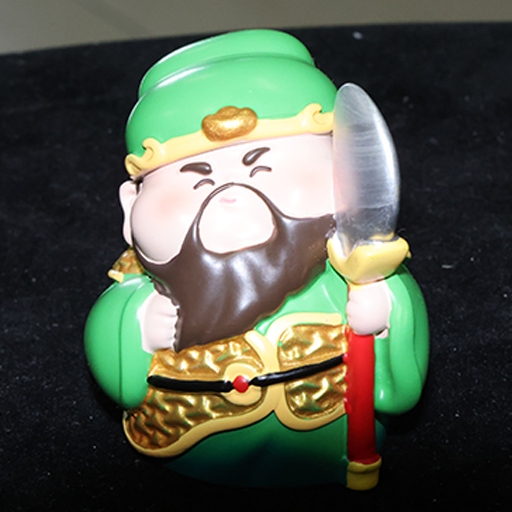}\\[\vspacesize pt]
                \includegraphics[width=\smallfigsize\textwidth]{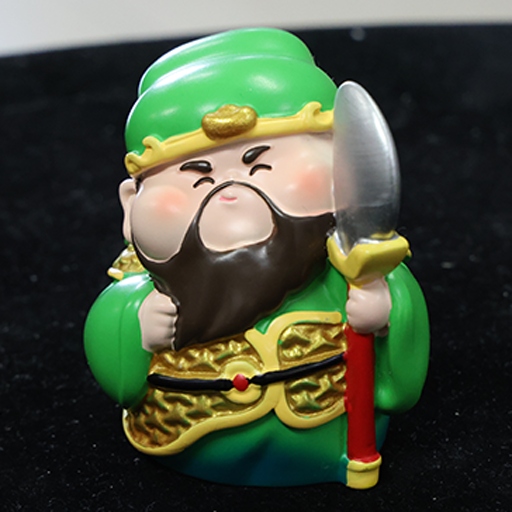}
            }
        }
    }
    \\
    {\fontsize{7}{7}\selectfont\sc Egyptcat} &
    {\fontsize{7}{7}\selectfont\sc Shakespeare}&
    {\fontsize{7}{7}\selectfont\sc Monkey}&
    {\fontsize{7}{7}\selectfont\sc Tree}&
    {\fontsize{7}{7}\selectfont\sc Monster}
    \\
    \includegraphics[width=\largefigsize\textwidth]{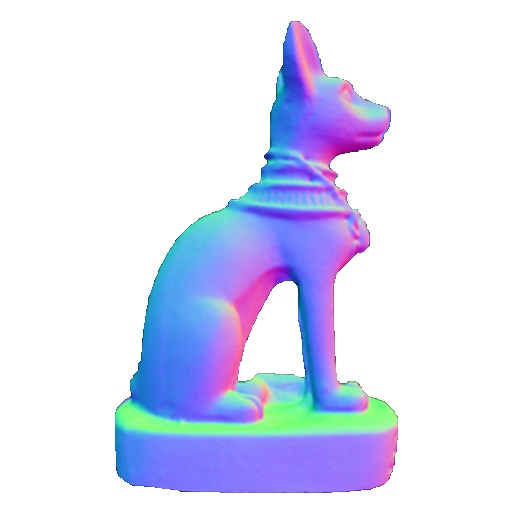}\hspace{\hspacesize pt}
    \raisebox{\rheightsize\height}{
        \makebox[\smallboxsize\textwidth]{
            \makecell{
                \includegraphics[width=\smallfigsize\textwidth]{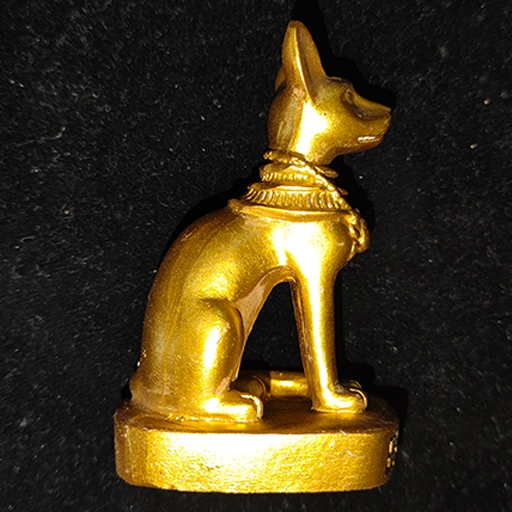}\\[\vspacesize pt]
                \includegraphics[width=\smallfigsize\textwidth]{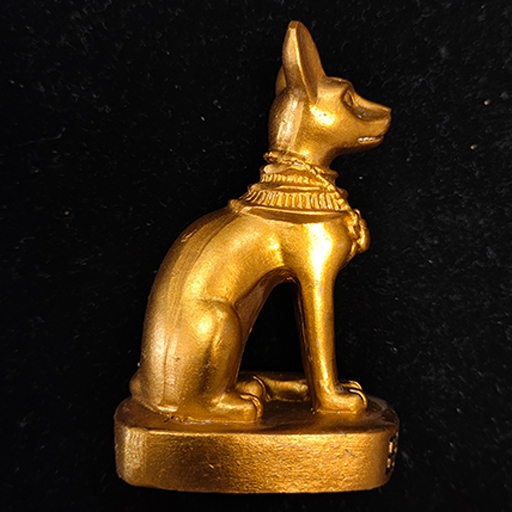}
            }
        }
    }&
    \includegraphics[width=\largefigsize\textwidth]{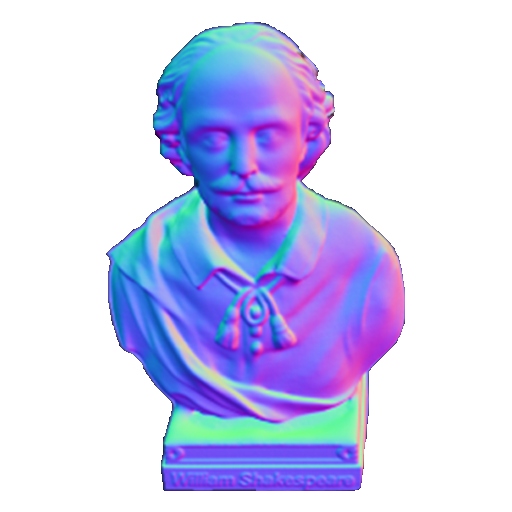}\hspace{\hspacesize pt}
    \raisebox{\rheightsize\height}{
        \makebox[\smallboxsize\textwidth]{
            \makecell{
                \includegraphics[width=\smallfigsize\textwidth]{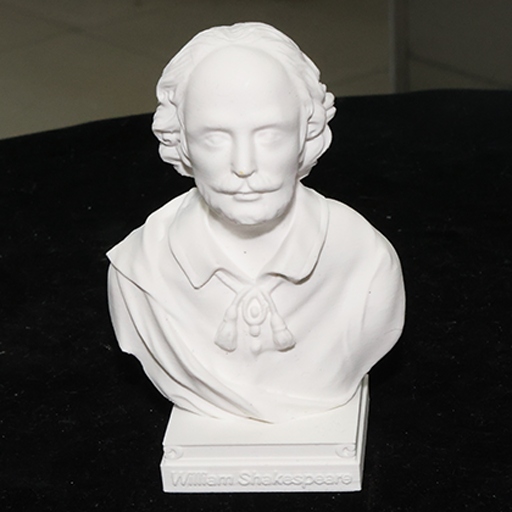}\\[\vspacesize pt]
                \includegraphics[width=\smallfigsize\textwidth]{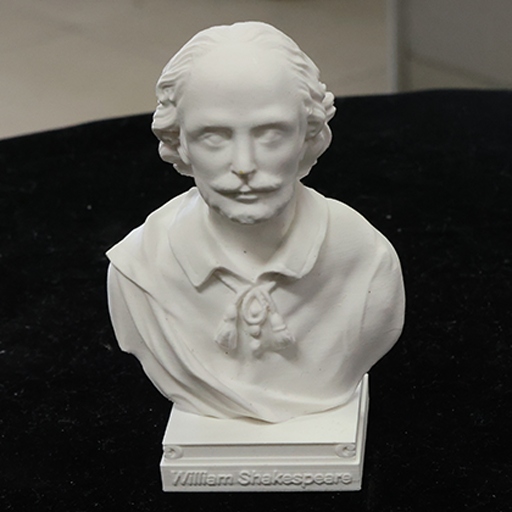}
            }
        }
    }&
    \includegraphics[width=\largefigsize\textwidth]{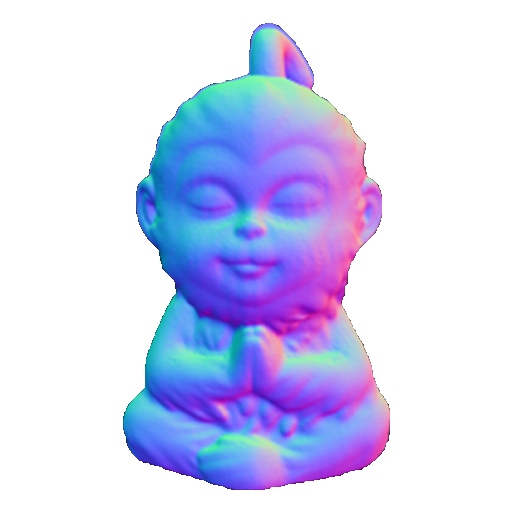}\hspace{\hspacesize pt}
    \raisebox{\rheightsize\height}{
        \makebox[\smallboxsize\textwidth]{
            \makecell{
                \includegraphics[width=\smallfigsize\textwidth]{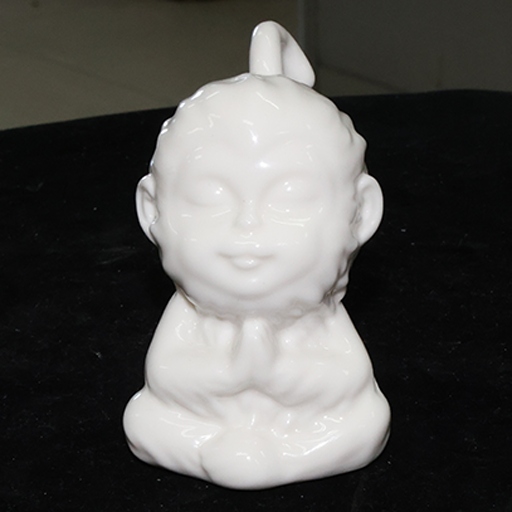}\\[\vspacesize pt]
                \includegraphics[width=\smallfigsize\textwidth]{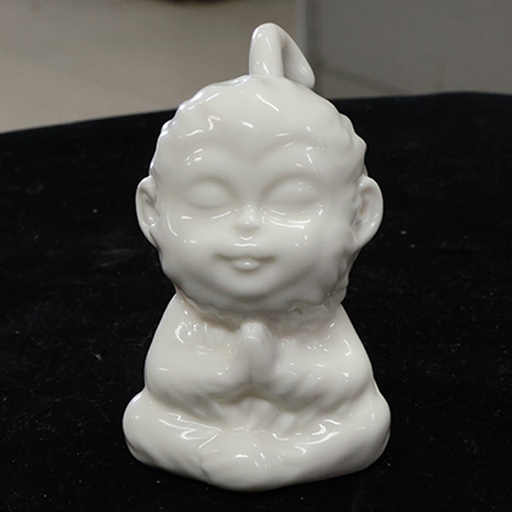}
            }
        }
    }&
    \includegraphics[width=\largefigsize\textwidth]{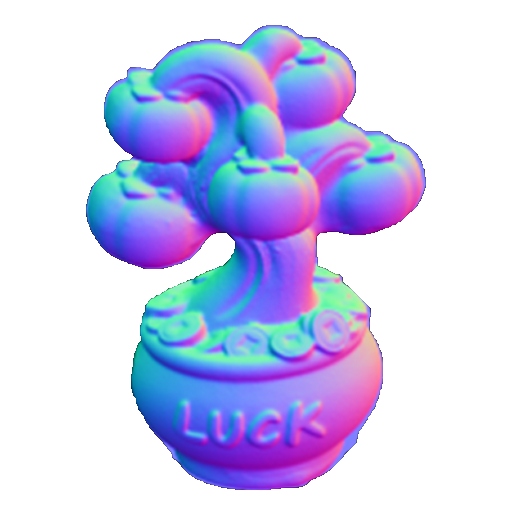}\hspace{\hspacesize pt}
    \raisebox{\rheightsize\height}{
        \makebox[\smallboxsize\textwidth]{
            \makecell{
                \includegraphics[width=\smallfigsize\textwidth]{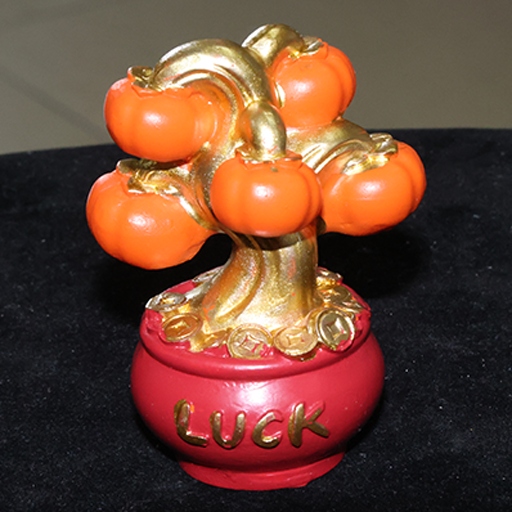}\\[\vspacesize pt]
                \includegraphics[width=\smallfigsize\textwidth]{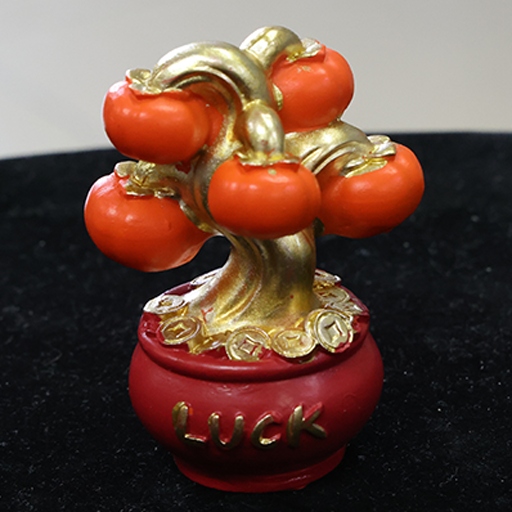}
            }
        }
    }&
    \includegraphics[width=\largefigsize\textwidth]{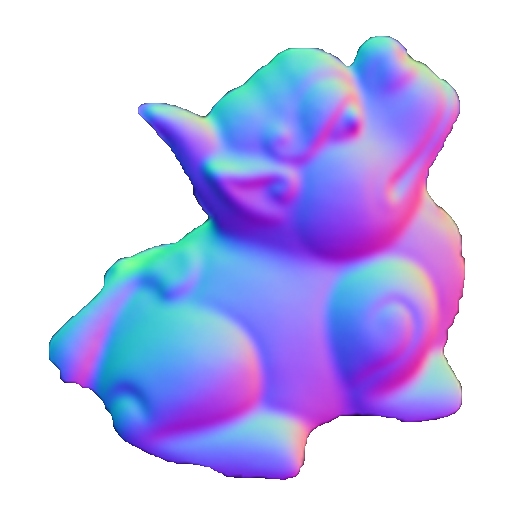}\hspace{\hspacesize pt}
    \raisebox{\rheightsize\height}{
        \makebox[\smallboxsize\textwidth]{
            \makecell{
                \includegraphics[width=\smallfigsize\textwidth]{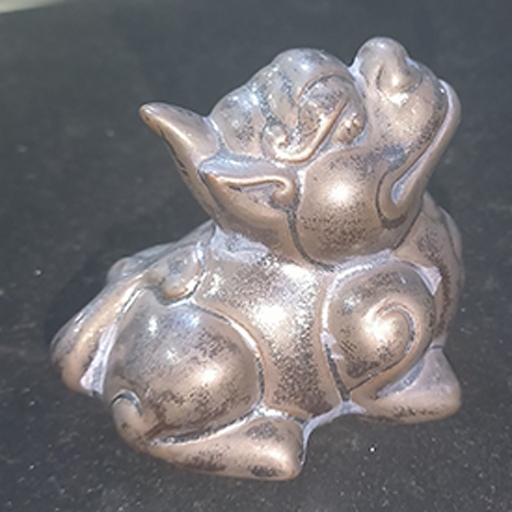}\\[\vspacesize pt]
                \includegraphics[width=\smallfigsize\textwidth]{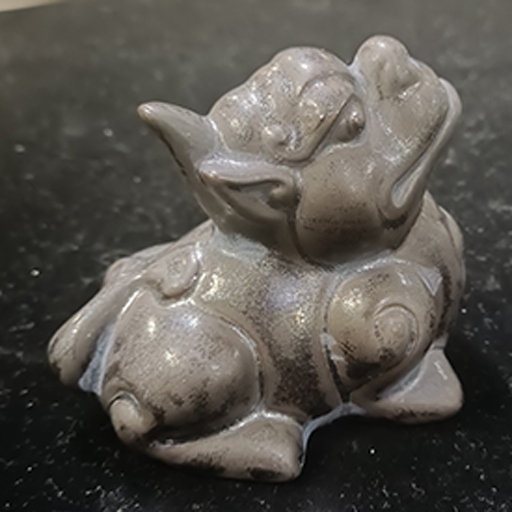}
            }
        }
    }
    \\
    {\fontsize{7}{8}\selectfont \sc Hugo} &
    {\fontsize{7}{8}\selectfont \sc Anderson} &
    {\fontsize{7}{8}\selectfont \sc Happydog} &
    {\fontsize{7}{8}\selectfont \sc Monkeyhead} &
    {\fontsize{7}{8}\selectfont \sc Luxun}
    \\
    \includegraphics[width=\largefigsize\textwidth]{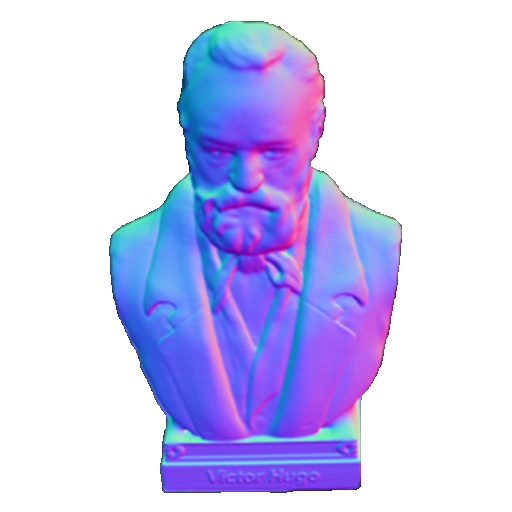}\hspace{\hspacesize pt}
    \raisebox{\rheightsize\height}{
        \makebox[\smallboxsize\textwidth]{
            \makecell{
                \includegraphics[width=\smallfigsize\textwidth]{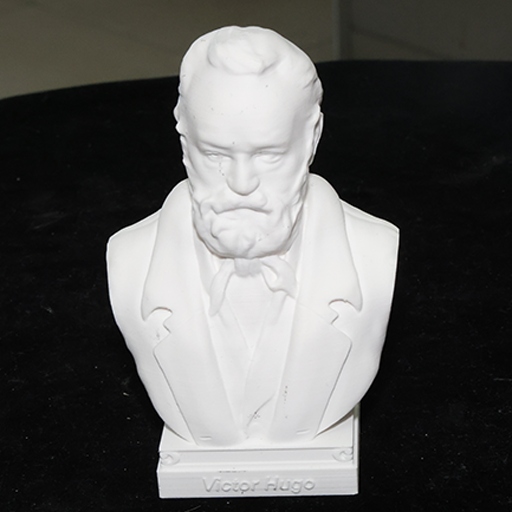}\\[\vspacesize pt]
                \includegraphics[width=\smallfigsize\textwidth]{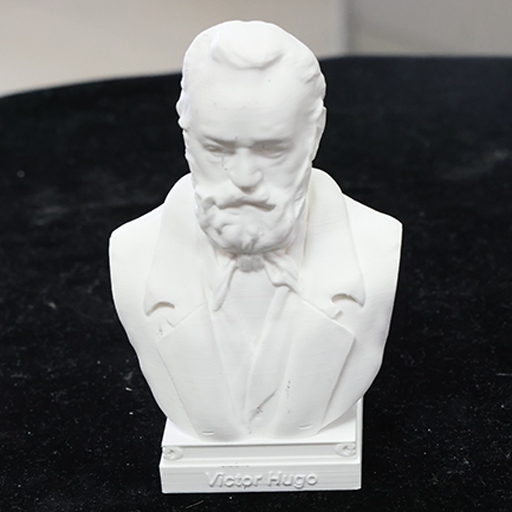}
            }
        }
    } &
    \includegraphics[width=\largefigsize\textwidth]{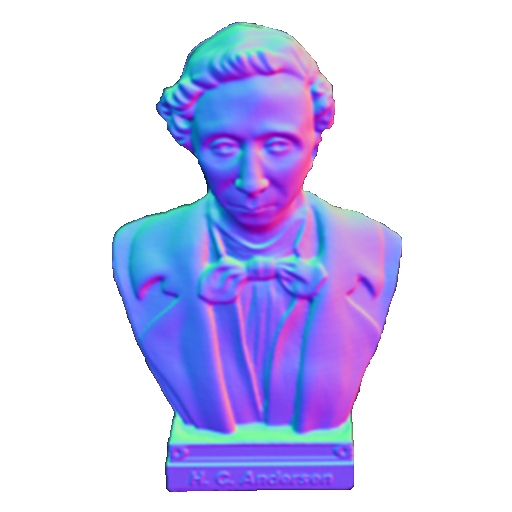}\hspace{\hspacesize pt}
    \raisebox{\rheightsize\height}{
        \makebox[\smallboxsize\textwidth]{
            \makecell{
                \includegraphics[width=\smallfigsize\textwidth]{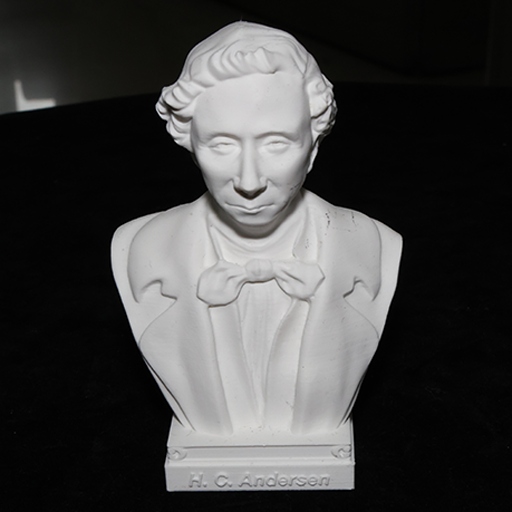}\\[\vspacesize pt]
                \includegraphics[width=\smallfigsize\textwidth]{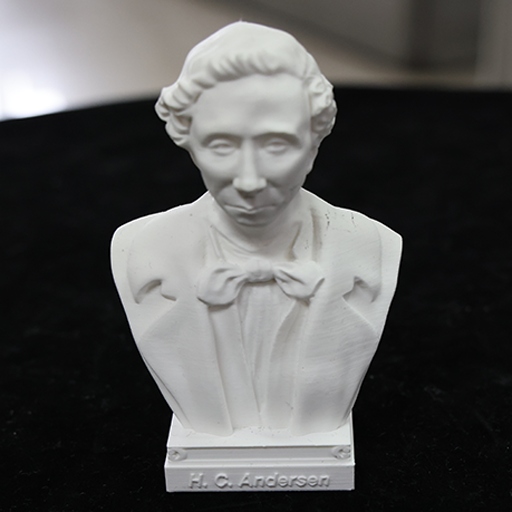}
            }
        }
    } &
    \includegraphics[width=\largefigsize\textwidth]{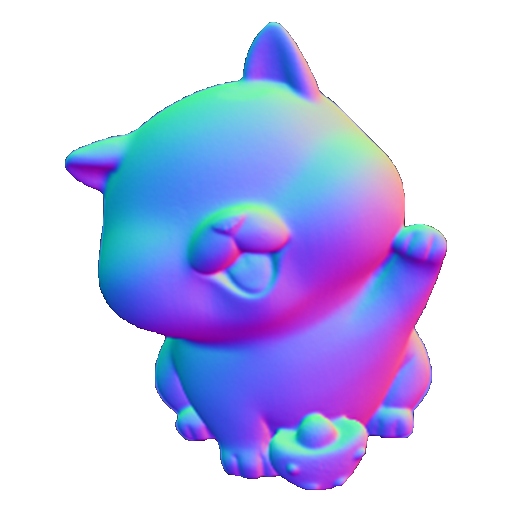}\hspace{\hspacesize pt}
    \raisebox{\rheightsize\height}{
        \makebox[\smallboxsize\textwidth]{
            \makecell{
                \includegraphics[width=\smallfigsize\textwidth]{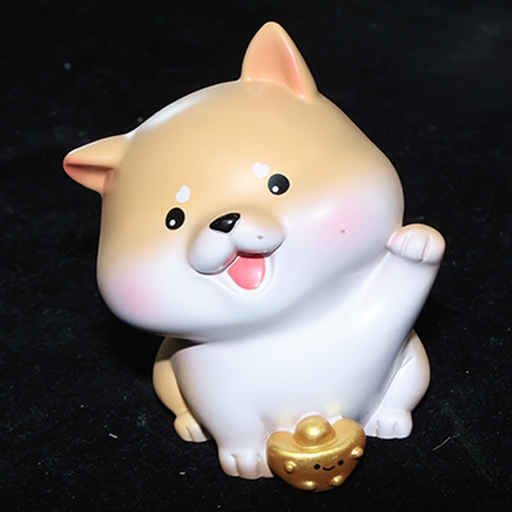}\\[\vspacesize pt]
                \includegraphics[width=\smallfigsize\textwidth]{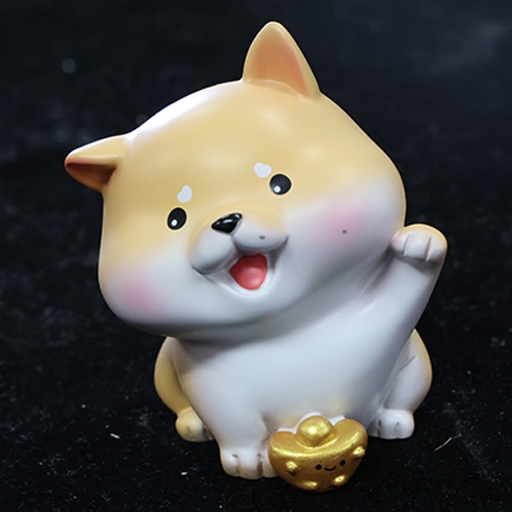}
            }
        }
    } &
    \includegraphics[width=\largefigsize\textwidth]{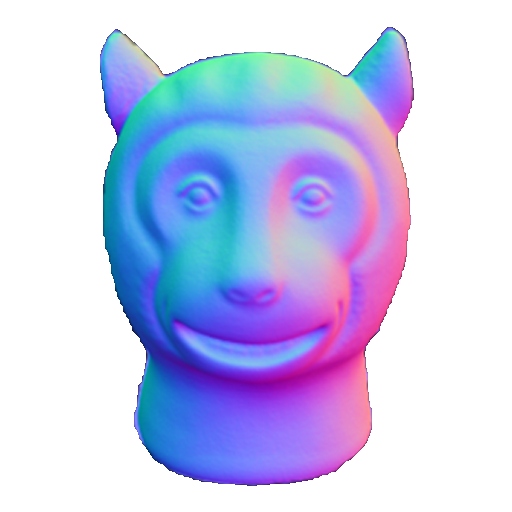}\hspace{\hspacesize pt}
    \raisebox{\rheightsize\height}{
        \makebox[\smallboxsize\textwidth]{
            \makecell{
                \includegraphics[width=\smallfigsize\textwidth]{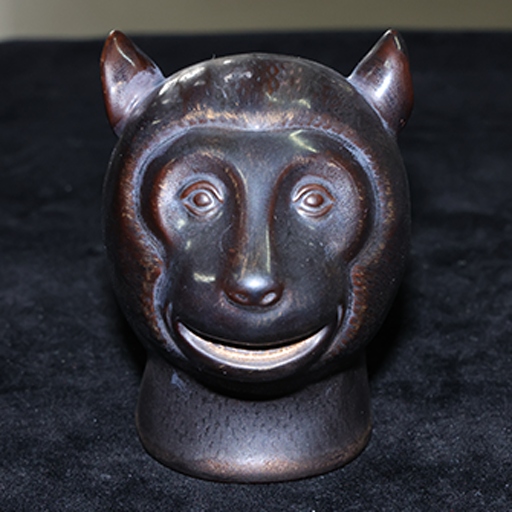}\\[\vspacesize pt]
                \includegraphics[width=\smallfigsize\textwidth]{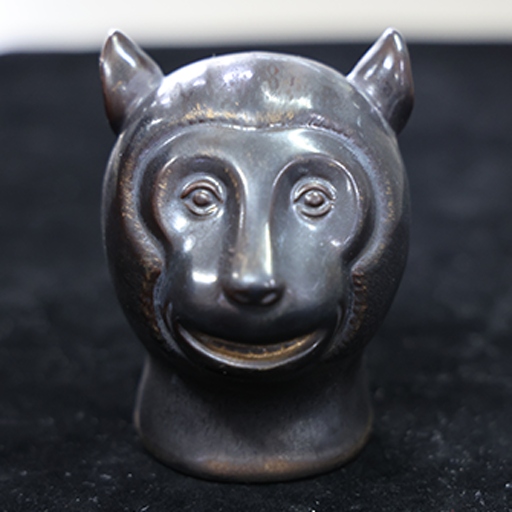}
            }
        }
    } &
    \includegraphics[width=\largefigsize\textwidth]{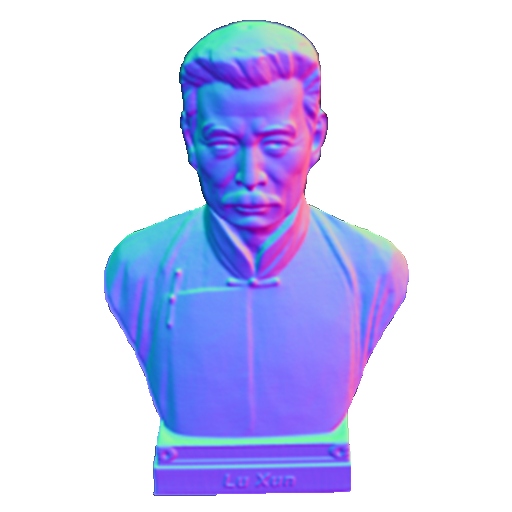}\hspace{\hspacesize pt}
    \raisebox{\rheightsize\height}{
        \makebox[\smallboxsize\textwidth]{
            \makecell{
                \includegraphics[width=\smallfigsize\textwidth]{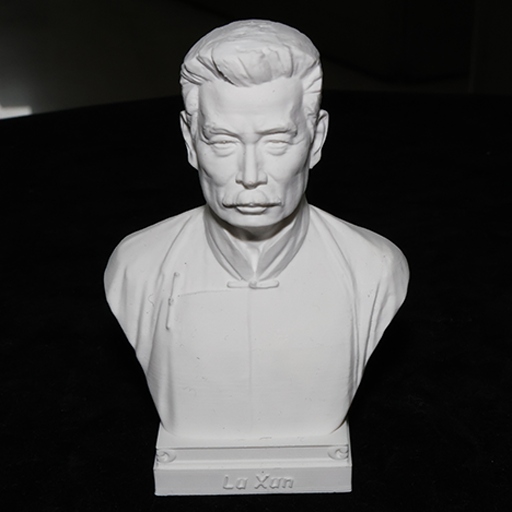}\\[\vspacesize pt]
                \includegraphics[width=\smallfigsize\textwidth]{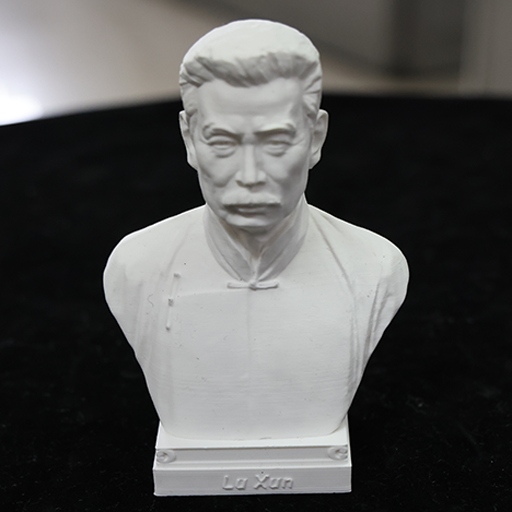}
            }
        }
    }
    \\
    {\fontsize{7}{8}\selectfont \sc Sittingbuddha} &
    {\fontsize{7}{8}\selectfont \sc Goethe} &
    {\fontsize{7}{8}\selectfont \sc Lion} &
    {\fontsize{7}{8}\selectfont \sc Persimmon} &
    {\fontsize{7}{8}\selectfont \sc Toy} &
    \\
    \includegraphics[width=\largefigsize\textwidth]{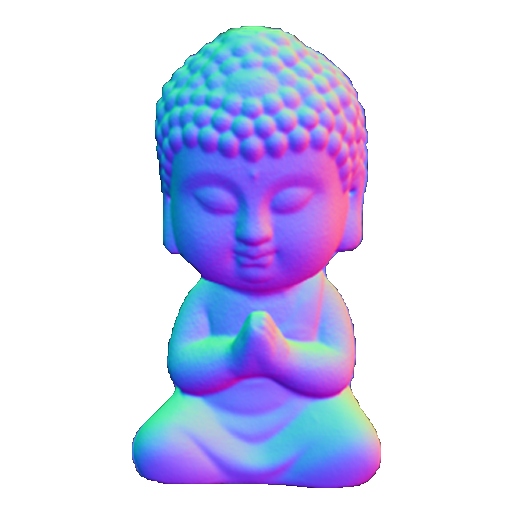}\hspace{\hspacesize pt}
    \raisebox{\rheightsize\height}{
        \makebox[\smallboxsize\textwidth]{
            \makecell{
                \includegraphics[width=\smallfigsize\textwidth]{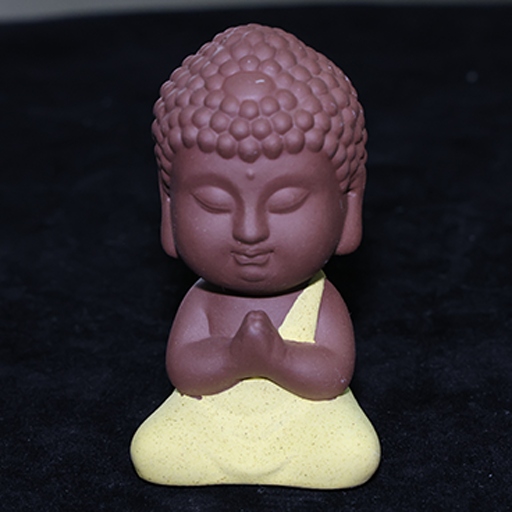}\\[\vspacesize pt]
                \includegraphics[width=\smallfigsize\textwidth]{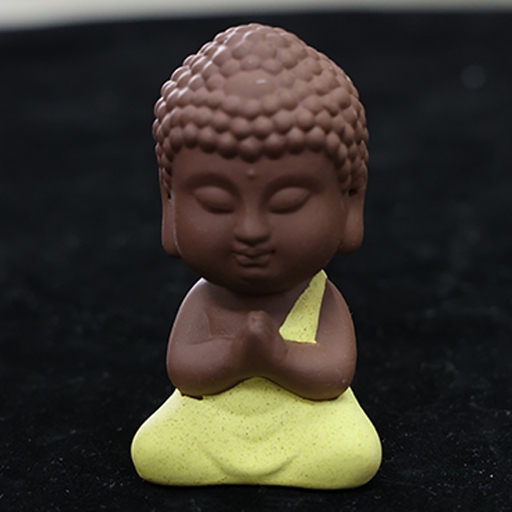}
            }
        }
    } &
    \includegraphics[width=\largefigsize\textwidth]{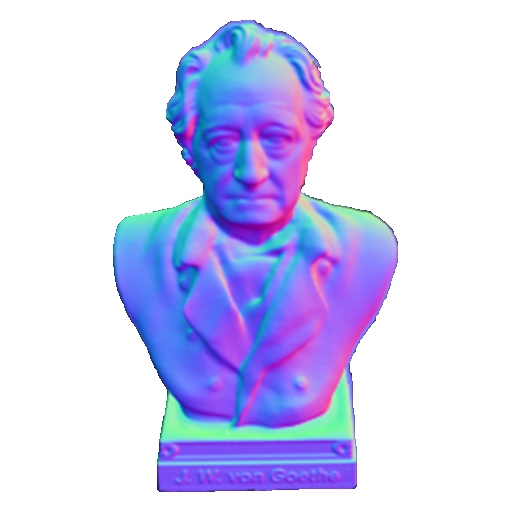}\hspace{\hspacesize pt}
    \raisebox{\rheightsize\height}{
        \makebox[\smallboxsize\textwidth]{
            \makecell{
                \includegraphics[width=\smallfigsize\textwidth]{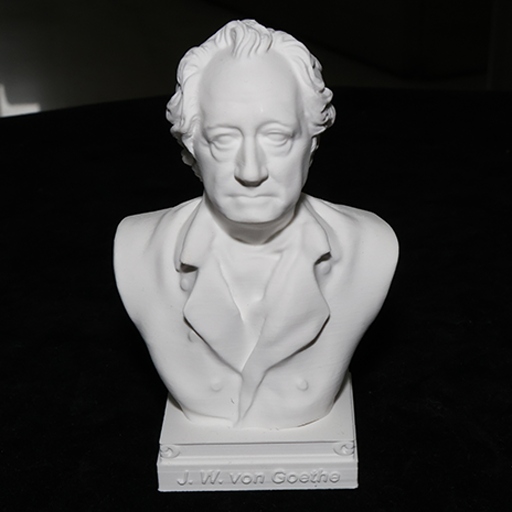}\\[\vspacesize pt]
                \includegraphics[width=\smallfigsize\textwidth]{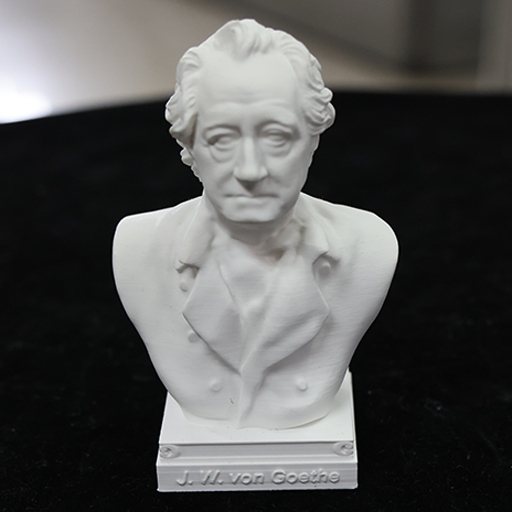}
            }
        }
    } &
    \includegraphics[width=\largefigsize\textwidth]{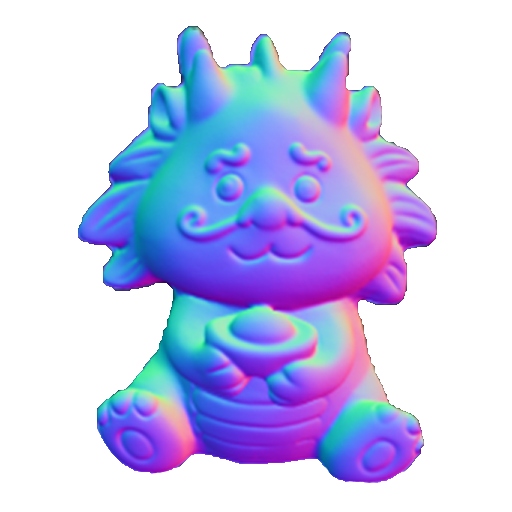}\hspace{\hspacesize pt}
    \raisebox{\rheightsize\height}{
        \makebox[\smallboxsize\textwidth]{
            \makecell{
                \includegraphics[width=\smallfigsize\textwidth]{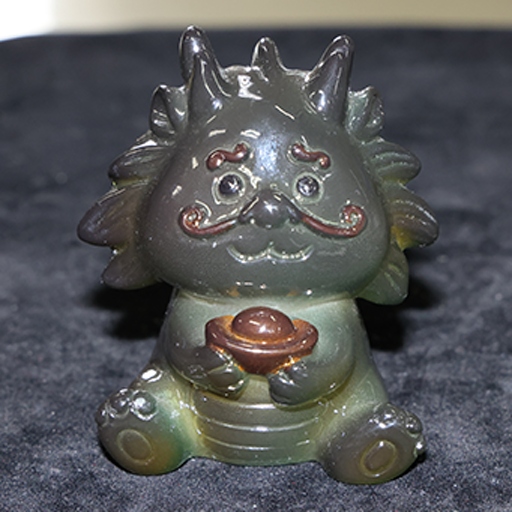}\\[\vspacesize pt]
                \includegraphics[width=\smallfigsize\textwidth]{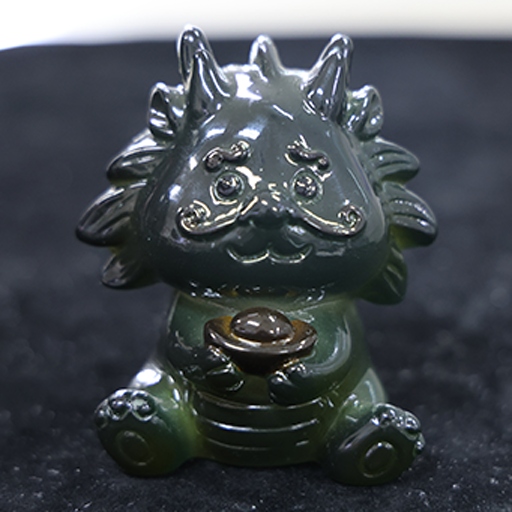}
            }
        }
    } &
    \includegraphics[width=\largefigsize\textwidth]{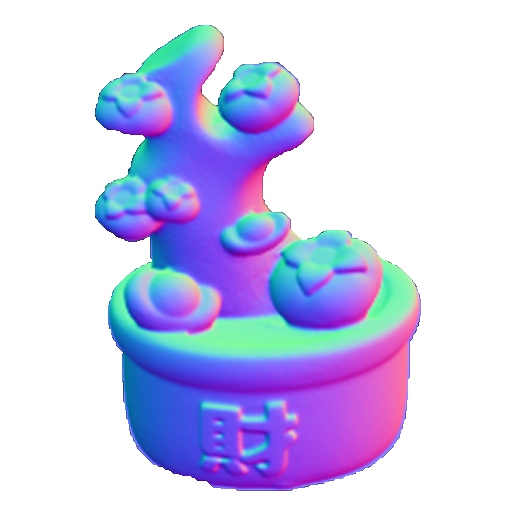}\hspace{\hspacesize pt}
    \raisebox{\rheightsize\height}{
        \makebox[\smallboxsize\textwidth]{
            \makecell{
                \includegraphics[width=\smallfigsize\textwidth]{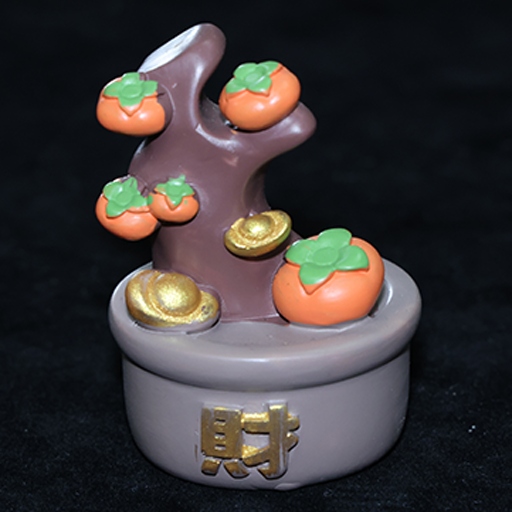}\\[\vspacesize pt]
                \includegraphics[width=\smallfigsize\textwidth]{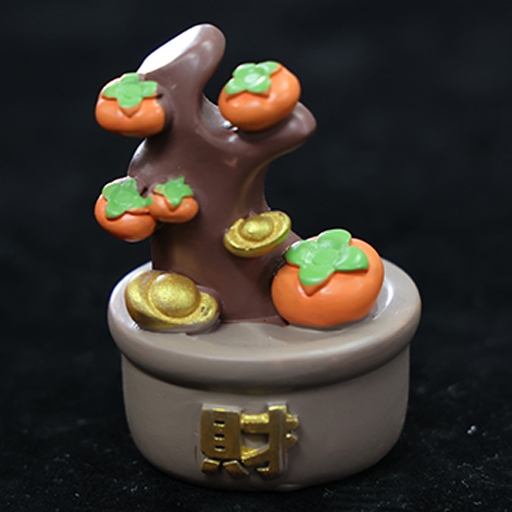}
            }
        }
    } &
    \includegraphics[width=\largefigsize\textwidth]{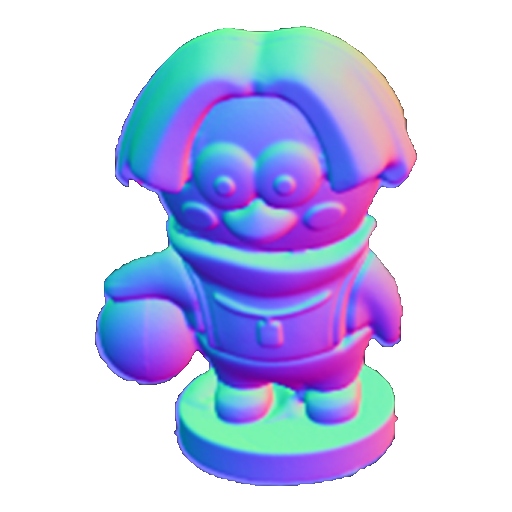}\hspace{\hspacesize pt}
    \raisebox{\rheightsize\height}{
        \makebox[\smallboxsize\textwidth]{
            \makecell{
                \includegraphics[width=\smallfigsize\textwidth]{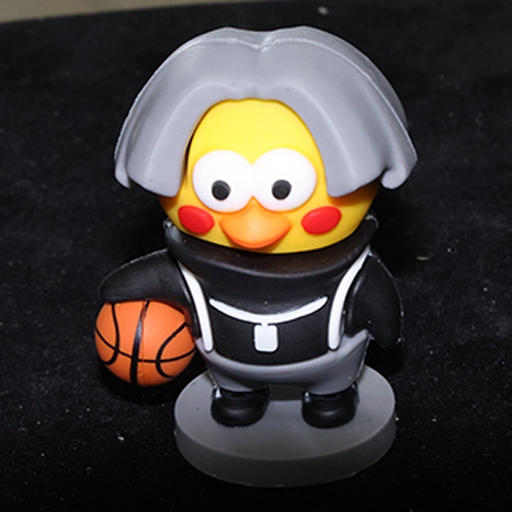}\\[\vspacesize pt]
                \includegraphics[width=\smallfigsize\textwidth]{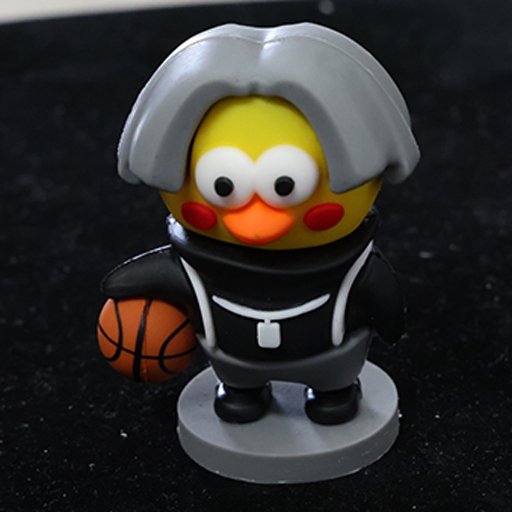}
            }
        }
    }
    \\
\end{NiceTabular}
    \caption{Overview of our \ourdata benchmark. We illustrate the GT surface normals, flash (top), and no-flash (bottom) images of \ourdata.}
    \label{fig:benchmark}

\end{figure}

\begin{figure}[t]
    \centering
    \includegraphics[width=\linewidth]{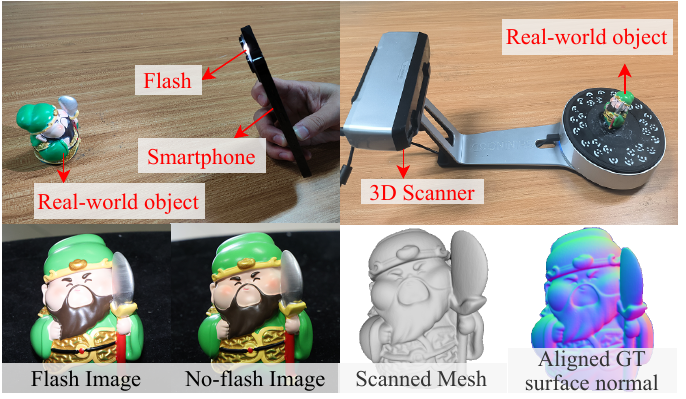}
    \caption{Capture setup of \ourdata. (Top) Flash/no-flash capture setup and 3D scanner used for GT shape acquisition. (Bottom) Flash/no-flash image observations and the aligned surface normal map rendered from the mesh.}
 
\label{fig:capture_setup}
\end{figure}
\subsection{A large-scale training dataset: \traindata}
\begin{figure}[!hbtp]
\includegraphics[width=\linewidth]{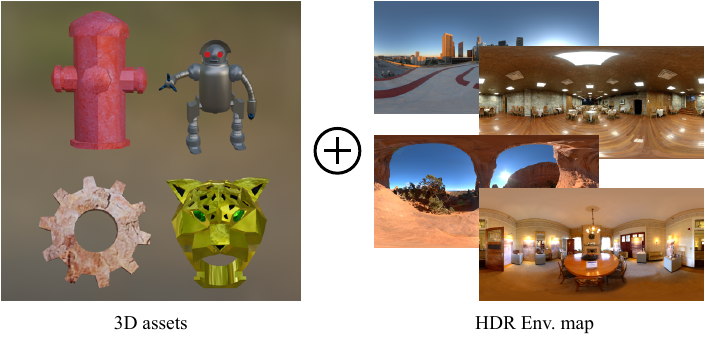}\\
\begin{adjustbox}{max width=\linewidth}
\addtolength{\tabcolsep}{-0.4em}
\begin{NiceTabular}{cccc}
    
    \includegraphics[width=0.4\linewidth]{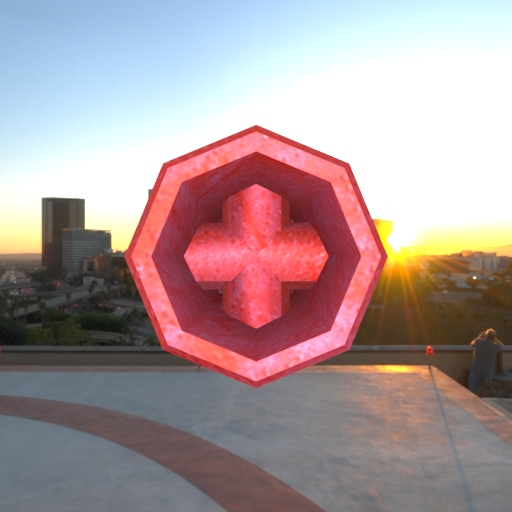} &
    \includegraphics[width=0.4\linewidth]{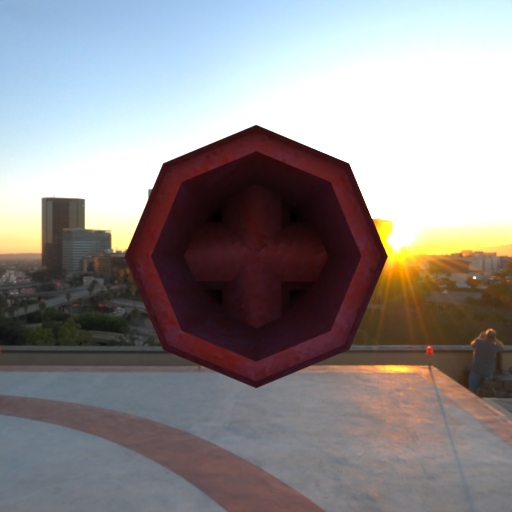}	&
    \includegraphics[width=0.4\linewidth]{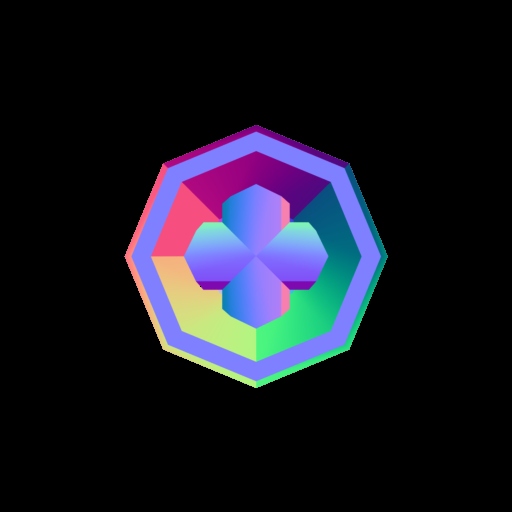} &
    \includegraphics[width=0.4\linewidth]{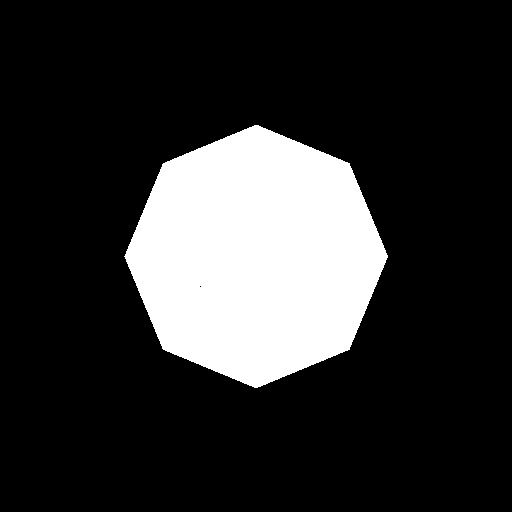} \\

    \includegraphics[width=0.4\linewidth]{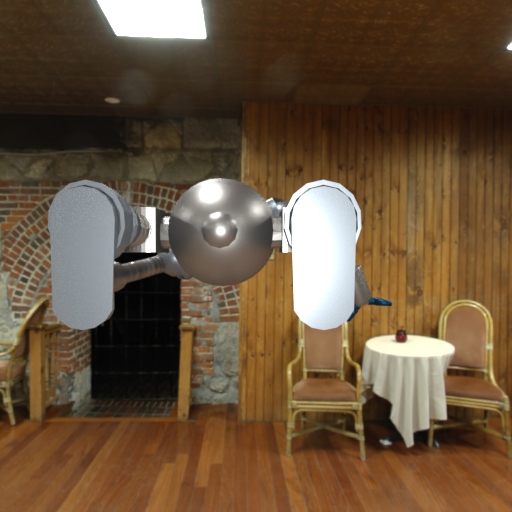} &
    \includegraphics[width=0.4\linewidth]{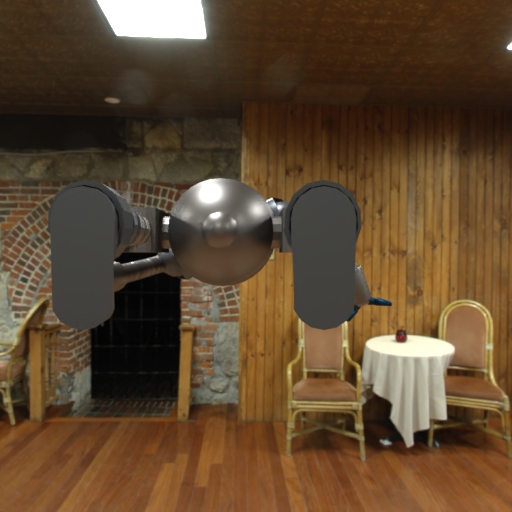}	&
    \includegraphics[width=0.4\linewidth]{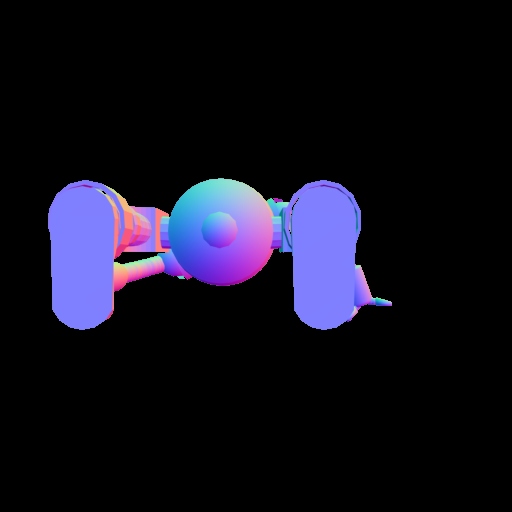} &
    \includegraphics[width=0.4\linewidth]{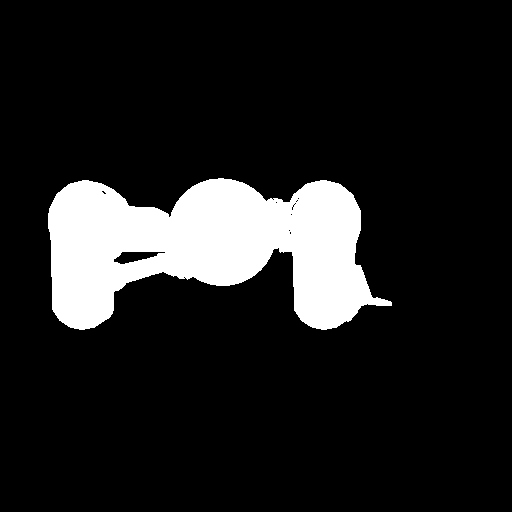} \\

    \includegraphics[width=0.4\linewidth]{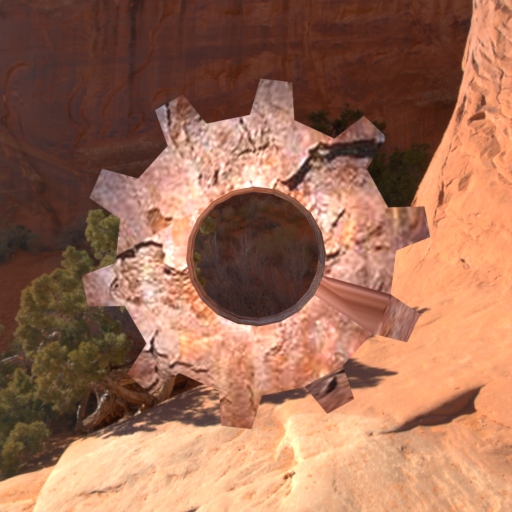} &
    \includegraphics[width=0.4\linewidth]{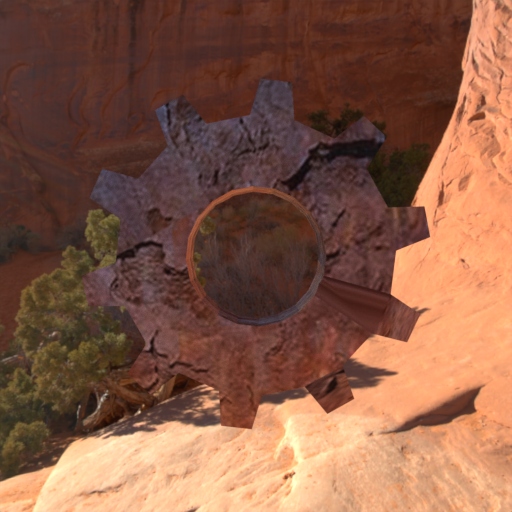}	&
    \includegraphics[width=0.4\linewidth]{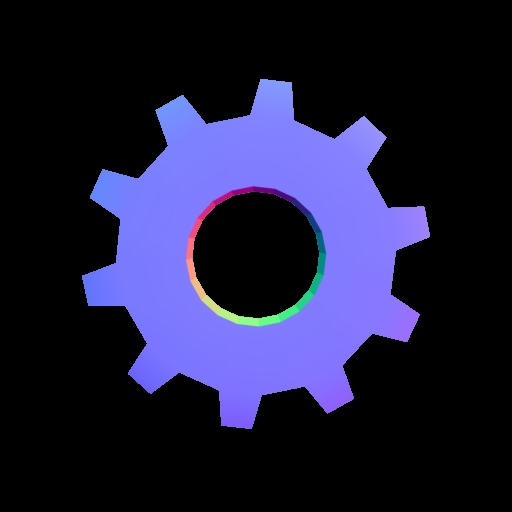} &
    \includegraphics[width=0.4\linewidth]{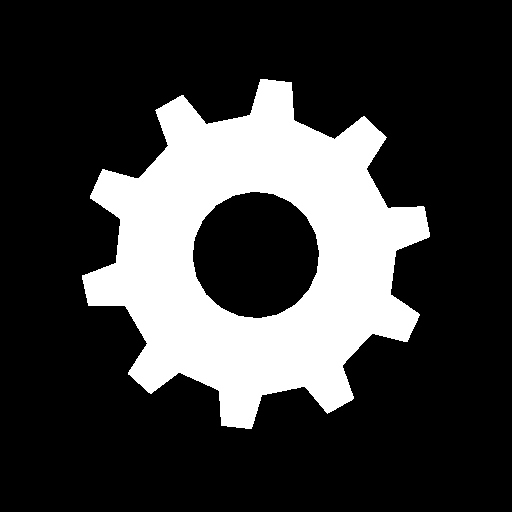} \\

    \includegraphics[width=0.4\linewidth]{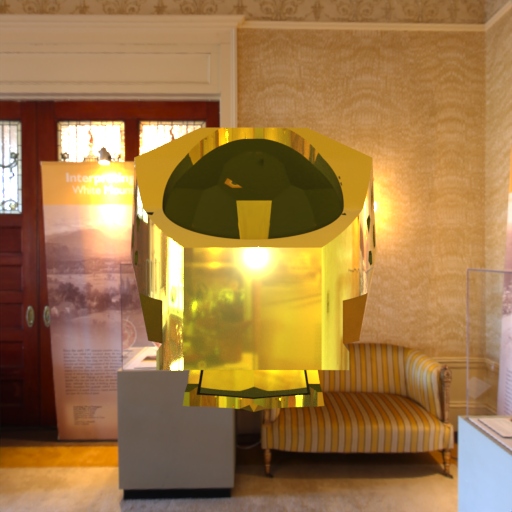} &
    \includegraphics[width=0.4\linewidth]{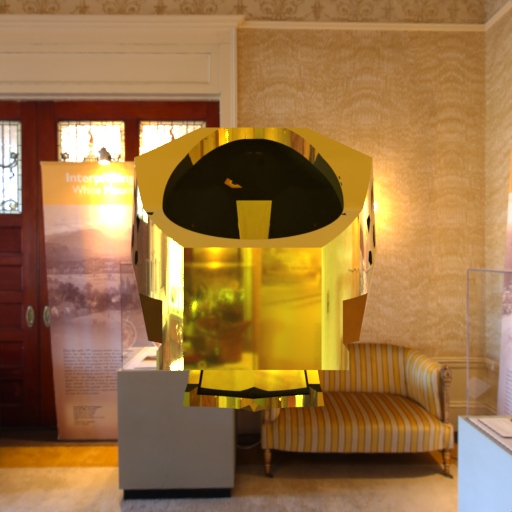}	&
    \includegraphics[width=0.4\linewidth]{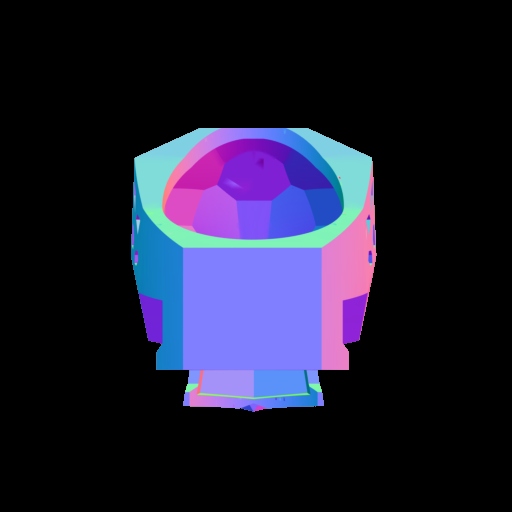} &
    \includegraphics[width=0.4\linewidth]{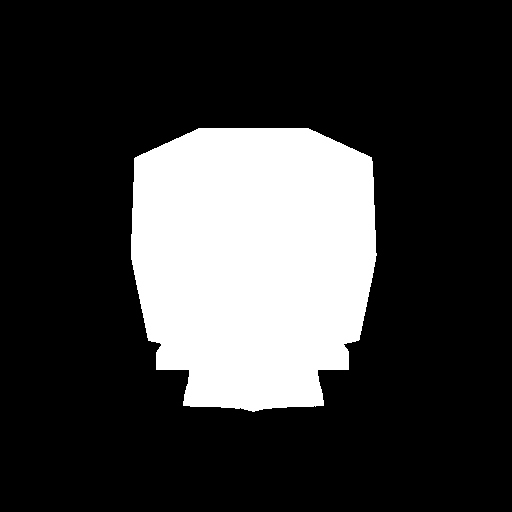} \\
    \makecell{Flash input} &
    \makecell{No-flash input} & 
    \makecell{GT Normal} & 
    \makecell{Mask}  \\
\end{NiceTabular}
\end{adjustbox}
\caption{Four examples from our rendered photorealistic training dataset \traindata. We show four 3D assets and the corresponding HDR environment maps in the first row and the bottom $4$ rows show the rendering data.}
\label{fig:traindata}
\end{figure}

Our \ours requires large-scale flash/no-flash photorealistic image pairs with the GT surface normals to fine-tune SD~\cite{Rombach_2022_CVPR}. While there already exists an off-the-shelf dataset for this purpose~\cite{2020Two}, it fails to reflect real-world object appearances due to the random combination of materials and shapes. Therefore, we rendered a more comprehensive and photorealistic flash/no-flash dataset with carefully composed shapes and materials, which contains $100,000$ flash/no-flash image pairs and corresponding GT surface normals, wishing to narrow the domain gaps between rendered data and the real-world counterpart as much as possible. 
The 3D assets for our training dataset \traindata are derived from the G-buffer Objaverse (gObjaverse) dataset~\cite{zuo2024sparse3d}. During rendering, after initially choosing $20,000$ samples from gObjaverse~\cite{zuo2024sparse3d}, we manually removed non-object assets to finalize selecting $17,000$ 3D assets with high-quality mesh and physics-based rendering material. Additionally, we collect $15$ HDR environmental maps to provide diverse scene illumination. We use the Cycles renderer in Blender~\cite{blender} for physically-based rendering. Each 3D object is normalized to fit within a unit sphere. We position six virtual cameras with a focal length of $30\mathrm{mm}$ at the top, bottom, left, right, front, and back of the object with a consistent camera-object distance of $500\mathrm{mm}$. To keep in line with the real-world layout of most smartphones and consumer cameras, we co-locate the point light source with the camera to capture flash and no-flash image pairs. For each capture, one of the $15$ environmental maps is selected to provide ambient illumination, and two images are shot with and without the point light activated. In this way, a large-scale dataset containing $100,000$ flash/no-flash image pairs is created, which are shown to be essential for \ours to produce detailed surface normal maps. \Fref{fig:traindata} illustrates four example training samples. We show four 3D assets about these examples and their corresponding HDR environment maps during rendering on the top. The rendering training data of \traindata is shown at the bottom.

\subsection{A comprehensive large-scale synthetic evaluation dataset: EvalFlash-synth}
Since our method leverages diffusion prior for surface normal estimation, a comprehensive evaluation of its robustness and stable normal estimation capabilities requires a large-scale synthetic test set. To this end, we introduce EvalFlash-synth, the first synthetic test set specifically designed to assess normal estimation performance under flash/no-flash conditions. Comprising $839$ high-quality test samples, this dataset is constructed by rendering 3D assets from gObjaverse~\cite{zuo2024sparse3d} that are disjoint from the training set and exhibit rich geometric details and diverse material properties. The rendering methodology used for the test set is consistent with that used for the training set, ensuring a fair and reliable evaluation.

\subsection{A real-world evaluation dataset: \ourdata}
A real-captured dataset with GT surface normal maps is essential for quantitatively evaluating the real-world performance of our proposed \ours. Since there does not exist such a dataset\footnote{FAID~\cite{aksoy2018flashambient} is the most relevant dataset, but its public links are currently unavailable. Photometric-stereo-style datasets typically use distant directional lighting rather than the near-light flash setup considered in our work.} in our setting, we decided to capture one by ourselves. Specifically, as illustrated in \fref{fig:benchmark}, we have collected data from $20$ objects characterized by intricate shapes and diverse materials. Objects were specifically chosen for their rich geometric details and diverse reflectance properties, ensuring thorough evaluation on highly detailed surface reconstruction and robustness across varying material types. The images were captured using a Canon EOS R5 camera at a resolution of $8192 \times 5464$ with the flash both on and off, followed by manual segmentation of the silhouette of objects. To collect the GT, as demonstrated in \fref{fig:capture_setup}, we utilize the EinScan-SP scanner to obtain the 3D mesh of each object. A typical mesh contains about 102k vertices and 206k faces, which is sufficient to support accurate rendering of ground-truth normal maps for quantitative evaluation. To align the scanned meshes with the object masks, we follow the methodology established by the DiLiGenT dataset~\cite{shi2016benchmark} and render the GT surface normal maps. Similar to the dataset collection of DiLiGenT dataset~\cite{shi2016benchmark}, the alignment and rendering process is time-consuming, taking us about $10$ days for building the scale of $20$ objects. To verify the alignment quality quantitatively, we measured the overlap between the RGB object masks and the rendered masks from the aligned meshes, and obtained an average IoU above 99.0\% across the 20 objects. This confirms that the mesh-to-image alignment is sufficiently accurate for quantitative evaluation, and small residual alignment errors are unlikely to affect the overall MAE trends substantially.

\section{Experiment}

We compare \ours with state-of-the-art methods on surface normal estimation. We also employ 3D reconstruction as an application to demonstrate the superiority of ours.
\subsection{Experimental settings}
\paragraph{Details about implementation}
Our model is fine-tuned from Stable Diffusion v2~\cite{Rombach_2022_CVPR}, leveraging its pretrained image diffusion weights rather than training from scratch. We train \ours on \traindata using AdamW optimizer~\cite{loshchilov2019decoupled} with a fixed learning rate of 3e-5. The batch size is set to $256$ with the data distributed across $8$ NVIDIA A6000 GPUs. The training spans $10$ epochs, which requires approximately $2$ days to complete. As our employed NVIDIA A6000 GPU can only process 2 batches at a time, we employ a gradient accumulation strategy by setting the batch sizes by conducting $16$ accumulations per card, resulting in a total batch size of $2 \times 16 \times 8 = 256$ before back-propagation. During pre-processing, we scale all the flash/no-flash image pairs to the range of $[-1,1]$ to match the input requirements of our Flash/no-flash VAE. We employ the accelerate library~\cite{accelerate} for efficient multi-GPU training and the xFormers library~\cite{xFormers2022} for optimized attention implementation. FlashNormal is efficient. It takes about 1s to predict the normal map of a $512 \times 512$ image on an NVIDIA RTX A6000 GPU, including preprocessing steps such as ROI extraction and image resizing.

\paragraph{Baselines and metrics} 
We select the state-of-the-art single image-based surface normal estimation methods including \ee, \me, \stablenormal and \lxm; photometric stereo method \sdmunips; and flash/no-flash based surface normal estimation method \mvm for comparison. We input the no-flash images to the single image-based methods, and let the others take in the flash/no-flash pairs. We use the released code and checkpoint of the baselines in the following experiments. 

For evaluation, we follow \stablenormal and employ the mean angular error~(MAE) in degree, the root mean square error~(RMSE), and the mean square error~(MSE) to measure the difference between the estimated and GT surface normal maps.

\subsection{Benchmark evaluation on \ourdata}
\newcommand\imgsize{0.3\linewidth}
\begin{figure*}[!t]
    \resizebox{\linewidth}{!}{
    \large
    \begin{NiceTabular}{cccccccc:cccccccc}
        \fontsize{25}{16}\selectfont \makecell{GT} 
        &\fontsize{25}{16}\selectfont \makecell{\ours\\\fontsize{15}{11}\selectfont(Ours)} &\fontsize{25}{16}\selectfont \makecell{E2E-FT\\\fontsize{15}{11}\selectfont\cite{martingarcia2024diffusione2eft}} &\fontsize{25}{16}\selectfont  \makecell{Metric3Dv2\\\fontsize{15}{11}\selectfont\cite{hu2024metric3dv2}}
        &\fontsize{25}{16}\selectfont  \makecell{StableNormal\\\fontsize{15}{11}\selectfont\cite{Ye2024StableNormalRD}}  &\fontsize{25}{16}\selectfont \makecell{LX18\\\fontsize{15}{11}\selectfont\cite{Li2018LearningTR} }
        &\fontsize{25}{16}\selectfont  \makecell{SDM-UniPS\\\fontsize{15}{11}\selectfont\cite{ikehata2023sdmunips}} &\fontsize{25}{16}\selectfont \makecell{MV20\\\fontsize{15}{11}\selectfont\cite{2020Two}} &\fontsize{25}{16}\selectfont \makecell{GT} 
        &\fontsize{25}{16}\selectfont  \makecell{\ours\\\fontsize{15}{11}\selectfont(Ours)} &\fontsize{25}{16}\selectfont  \makecell{E2E-FT\\\fontsize{15}{11}\selectfont\cite{martingarcia2024diffusione2eft}} &\fontsize{25}{16}\selectfont  \makecell{Metric3Dv2\\\fontsize{15}{11}\selectfont\cite{hu2024metric3dv2}}
        &\fontsize{25}{16}\selectfont  \makecell{StableNormal\\\fontsize{15}{11}\selectfont\cite{Ye2024StableNormalRD}}  &\fontsize{25}{16}\selectfont \makecell{LX18\\\fontsize{15}{11}\selectfont\cite{Li2018LearningTR} }&\fontsize{25}{16}\selectfont  \makecell{SDM-UniPS\\\fontsize{15}{11}\selectfont\cite{ikehata2023sdmunips}} &\fontsize{25}{16}\selectfont \makecell{MV20\\\fontsize{15}{11}\selectfont\cite{2020Two}} \\
        
        {\includegraphics[width=0.3\linewidth, trim=100px 2px 100px 4px, clip=true]{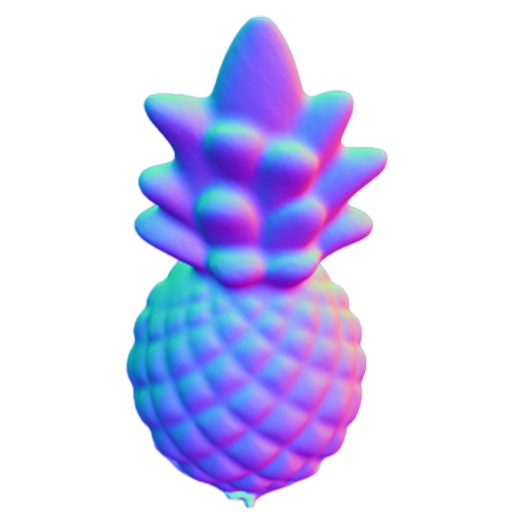}}
        &{\includegraphics[width=0.3\linewidth, trim=100px 2px 100px 4px, clip=true]{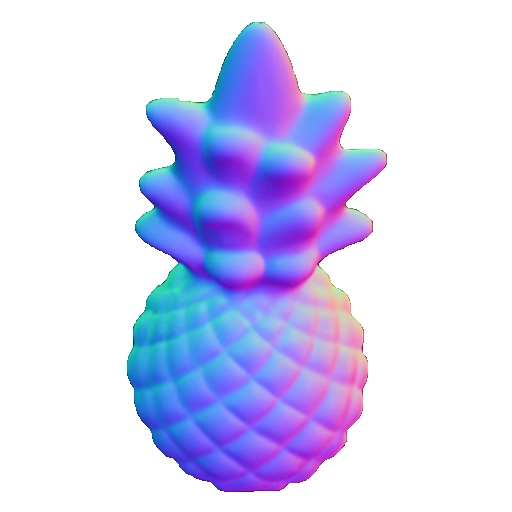}}
        &{\includegraphics[width=0.3\linewidth, trim=100px 2px 100px 4px, clip=true]{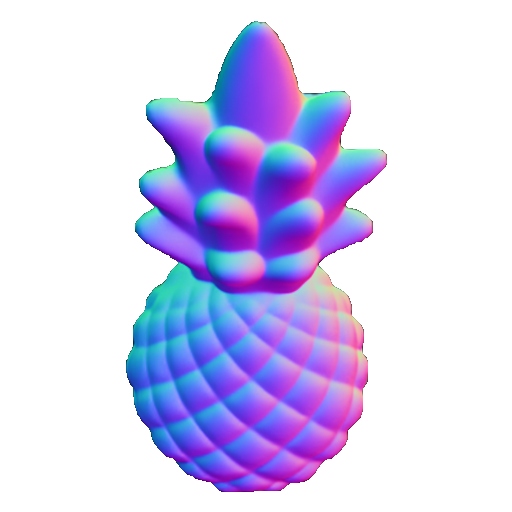}}
        &{\includegraphics[width=0.3\linewidth, trim=100px 2px 100px 4px, clip=true]{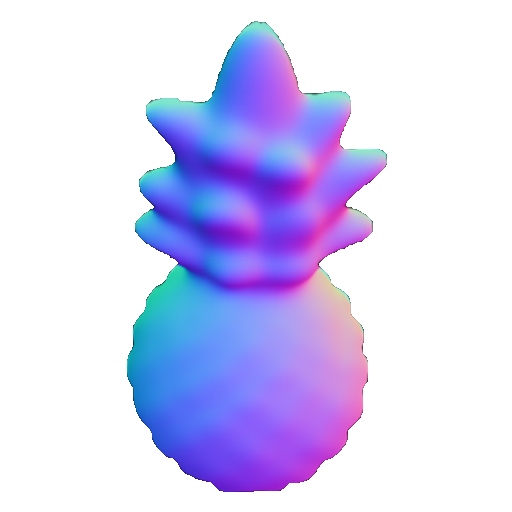}}
        &{\includegraphics[width=0.3\linewidth, trim=100px 2px 100px 4px, clip=true]{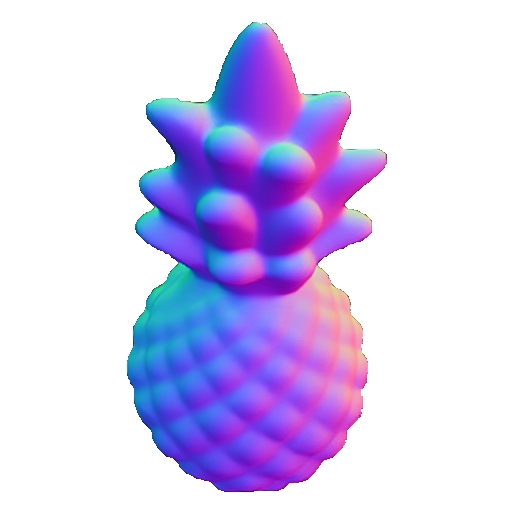}}
        &{\includegraphics[width=0.3\linewidth, trim=100px 2px 100px 4px, clip=true]{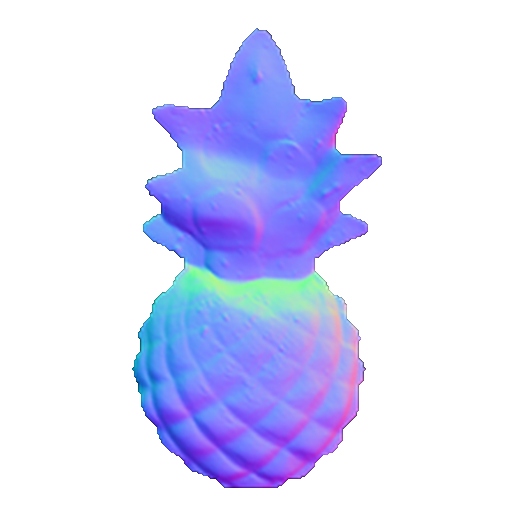}}
        &{\includegraphics[width=0.3\linewidth, trim=100px 2px 100px 4px, clip=true]{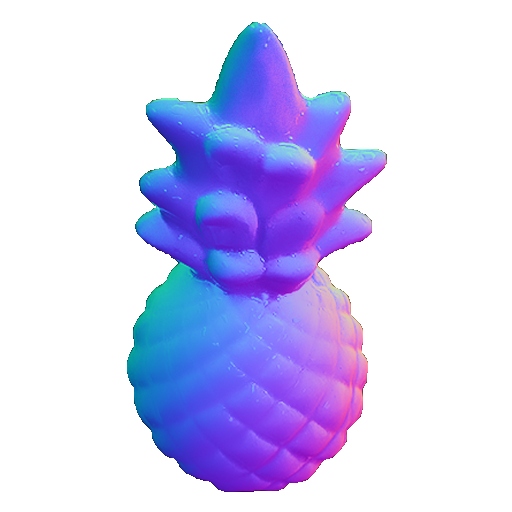}}
        &{\includegraphics[width=0.3\linewidth, trim=100px 2px 100px 4px, clip=true]{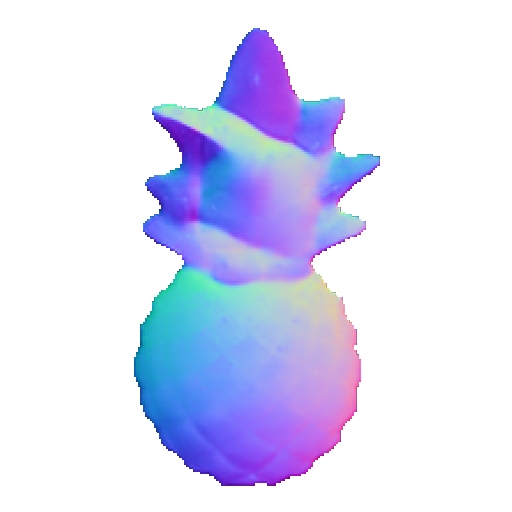}} 
        &{\includegraphics[ width=\imgsize, trim=100px 2px 100px 4px, clip=true]{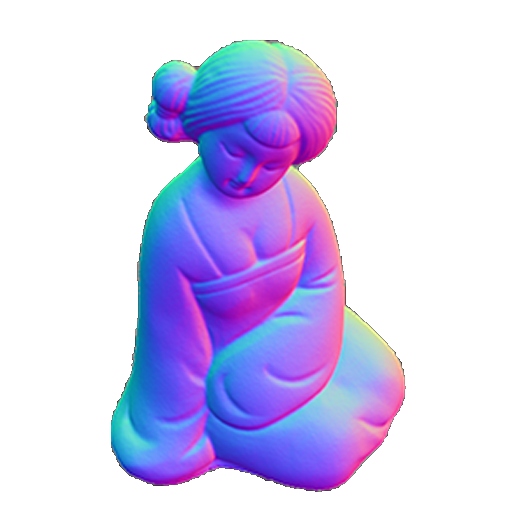}}
        &{\includegraphics[ width=\imgsize, trim=100px 2px 100px 4px, clip=true]{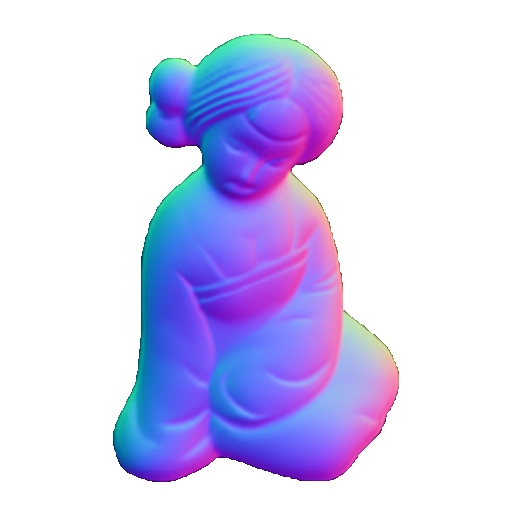}}
        &{\includegraphics[ width=\imgsize, trim=100px 2px 100px 4px, clip=true]{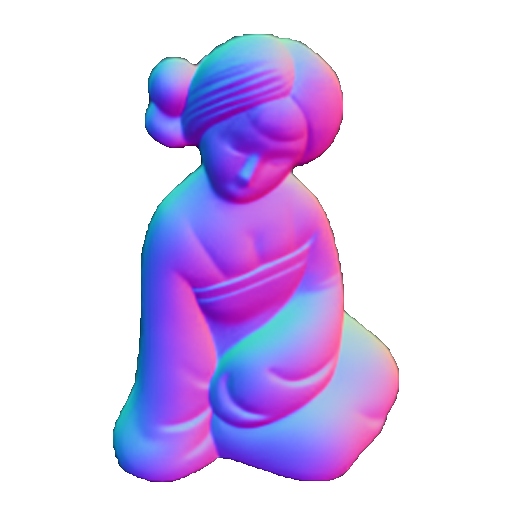}}
        &{\includegraphics[ width=\imgsize, trim=100px 2px 100px 4px, clip=true]{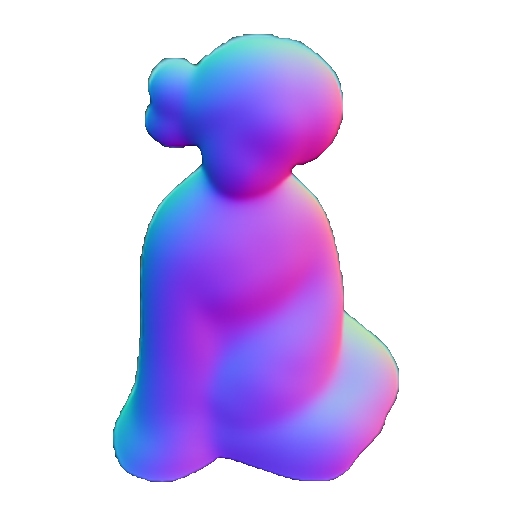}}
        &{\includegraphics[ width=\imgsize, trim=100px 2px 100px 4px, clip=true]{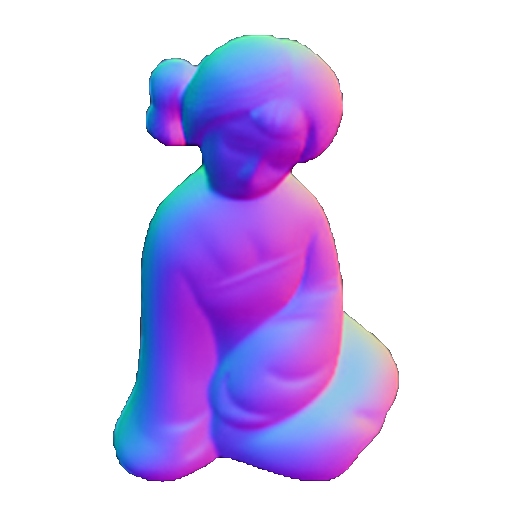}}
        &{\includegraphics[ width=\imgsize, trim=100px 2px 100px 4px, clip=true]{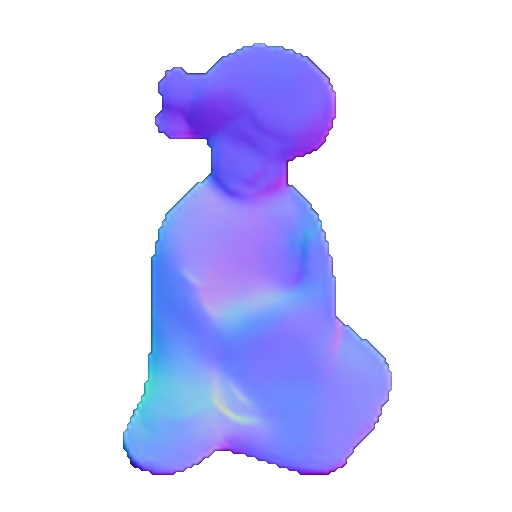}}
        &{\includegraphics[ width=\imgsize, trim=100px 2px 100px 4px, clip=true]{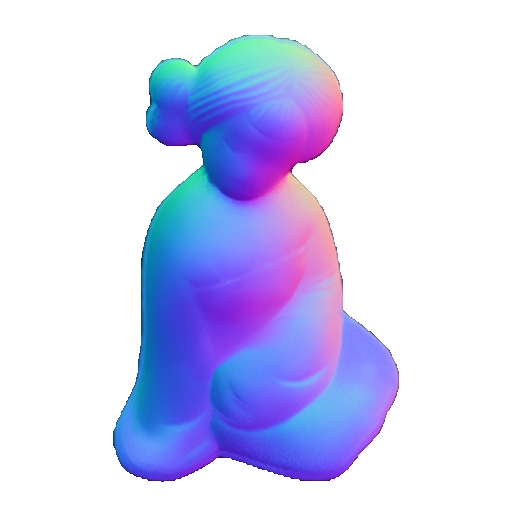}}
        &{\includegraphics[ width=\imgsize, trim=100px 2px 100px 4px, clip=true]{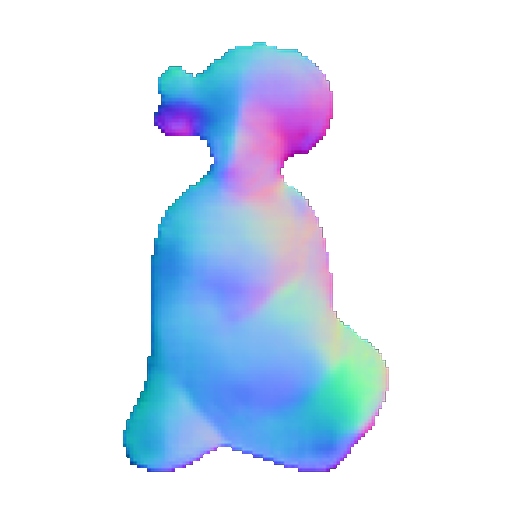}}\\
        
        & \fontsize{25}{16}\selectfont MAE=\textbf{11.13} 
        &\fontsize{25}{16}\selectfont MAE=14.45 
        &\fontsize{25}{16}\selectfont MAE=16.03 
        &\fontsize{25}{16}\selectfont MAE=18.27  
        &\fontsize{25}{16}\selectfont MAE=32.10 
        &\fontsize{25}{16}\selectfont MAE=18.51 
        &\fontsize{25}{16}\selectfont MAE=33.19 
        && \fontsize{25}{16}\selectfont MAE=\textbf{17.79} 
        & \fontsize{25}{16}\selectfont MAE=19.84 
        & \fontsize{25}{16}\selectfont MAE=18.75 
        & \fontsize{25}{16}\selectfont MAE=18.15  
        & \fontsize{25}{16}\selectfont MAE=40.98 
        & \fontsize{25}{16}\selectfont MAE=24.01 
        &\fontsize{25}{16}\selectfont MAE=45.92\\
        
        {\includegraphics[ width=\imgsize]{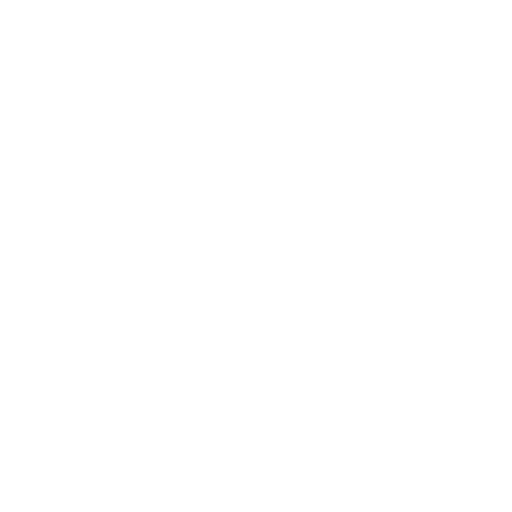}}
        &{\includegraphics[ width=\imgsize, trim=100px 2px 100px 4px, clip=true]{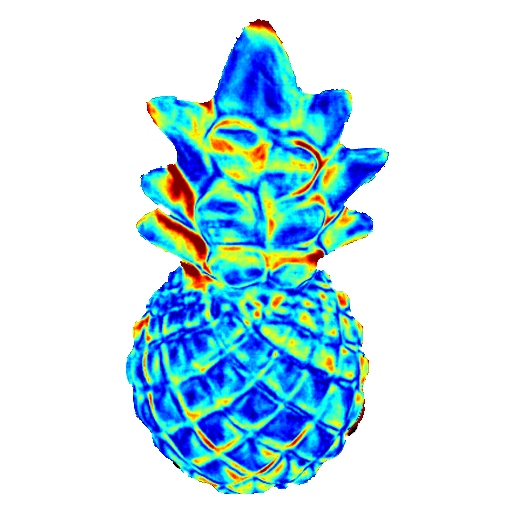}}
        &{\includegraphics[ width=\imgsize, trim=100px 2px 100px 4px, clip=true]{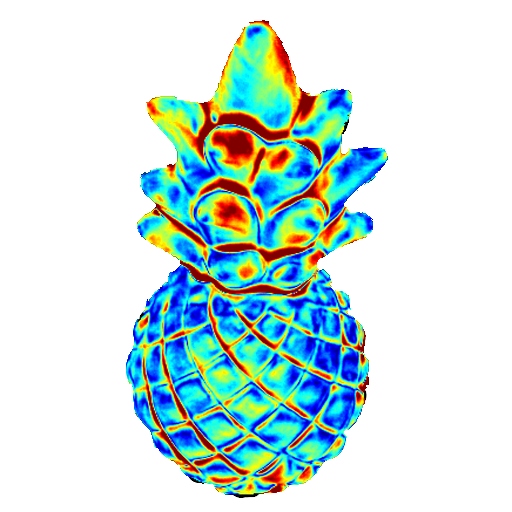}}
        &{\includegraphics[ width=\imgsize, trim=100px 2px 100px 4px, clip=true]{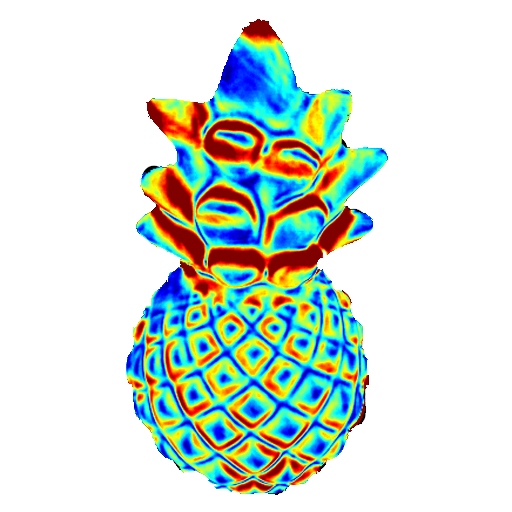}}
        &{\includegraphics[ width=\imgsize, trim=100px 2px 100px 4px, clip=true]{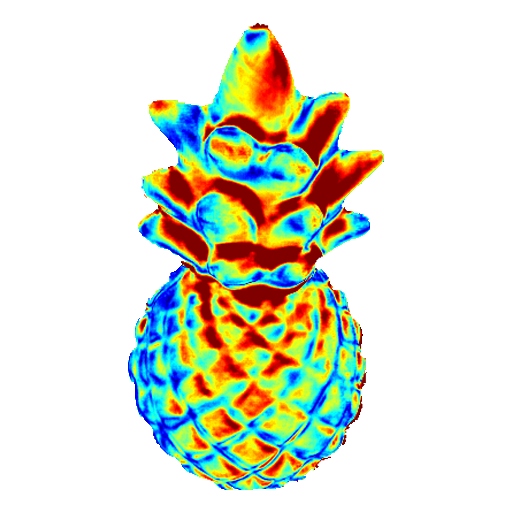}}
        
        &{\includegraphics[ width=\imgsize, trim=100px 2px 100px 4px, clip=true]{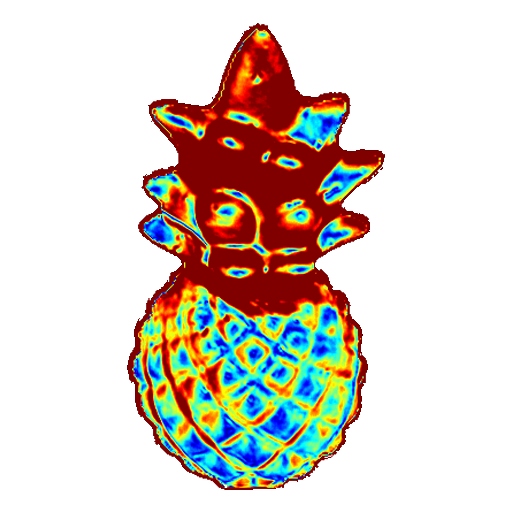}}
        &{\includegraphics[ width=\imgsize, trim=100px 2px 100px 4px, clip=true]{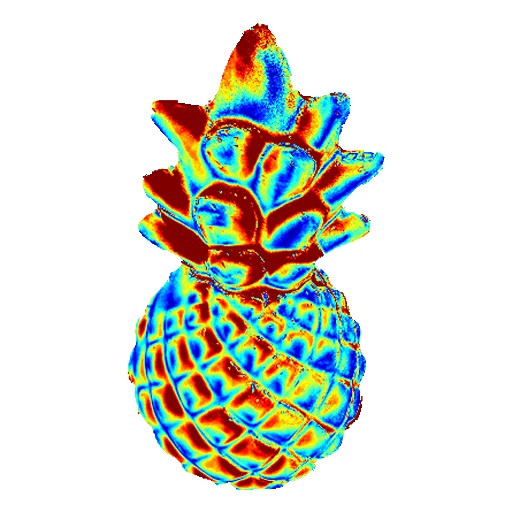}} 
        &{\includegraphics[ width=\imgsize, trim=100px 2px 100px 4px, clip=true]{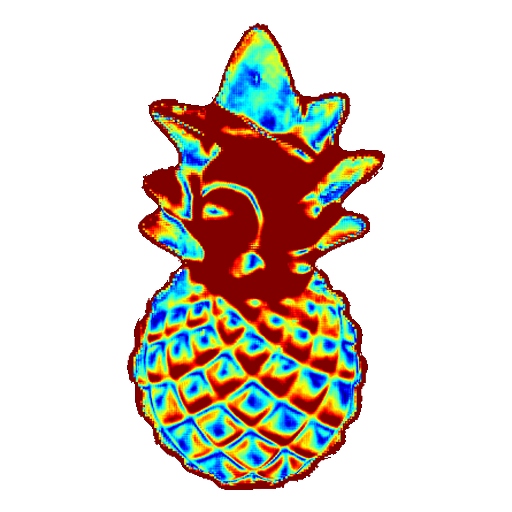}}
        \begin{minipage}{0.030\textwidth} \centering
        	\vspace{-20.5em} \makebox[\textwidth]{ \fontsize{20}{16}\selectfont $\,30^{\circ}$}\\ 
        	\makebox[0.12\textwidth]{}\includegraphics[width=0.9\textwidth]{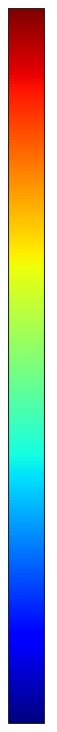} \\ 
        	\makebox[\textwidth]{ \fontsize{20}{16}\selectfont $\,0^{\circ}$}\\
        \end{minipage}
		&{\includegraphics[ width=\imgsize]{fig/quan_com/white.jpg}}
		&{\includegraphics[ width=\imgsize, trim=100px 2px 100px 4px, clip=true]{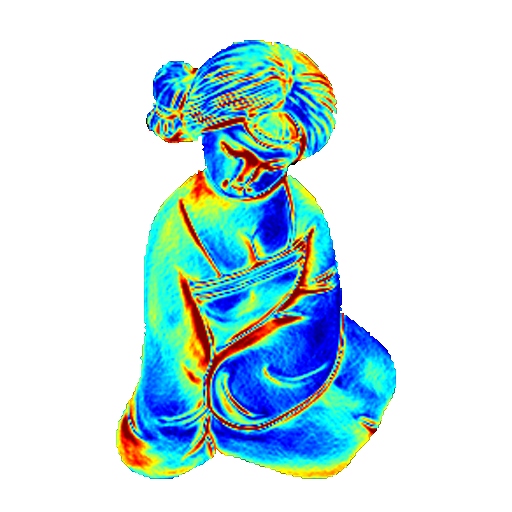}}
		&{\includegraphics[ width=\imgsize, trim=100px 2px 100px 4px, clip=true]{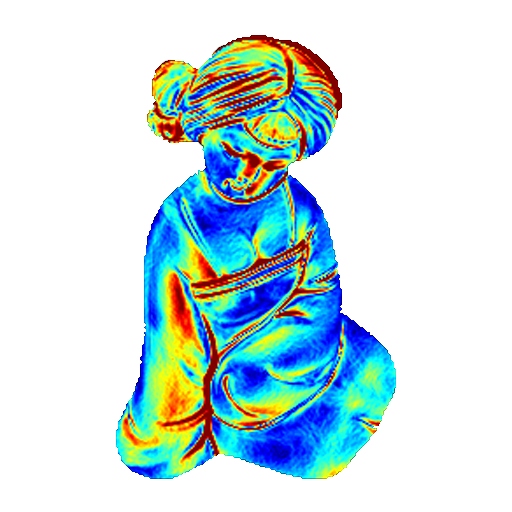}}
		&{\includegraphics[ width=\imgsize, trim=100px 2px 100px 4px, clip=true]{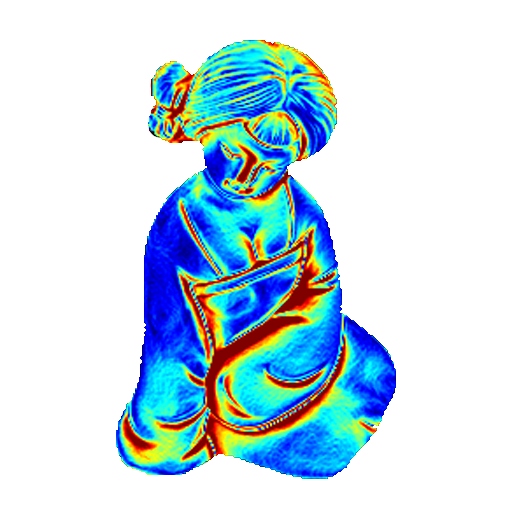}}
		&{\includegraphics[ width=\imgsize, trim=100px 2px 100px 4px, clip=true]{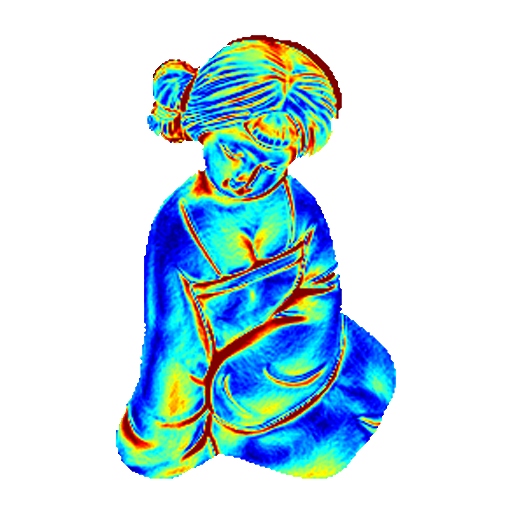}}
		&{\includegraphics[ width=\imgsize, trim=100px 2px 100px 4px, clip=true]{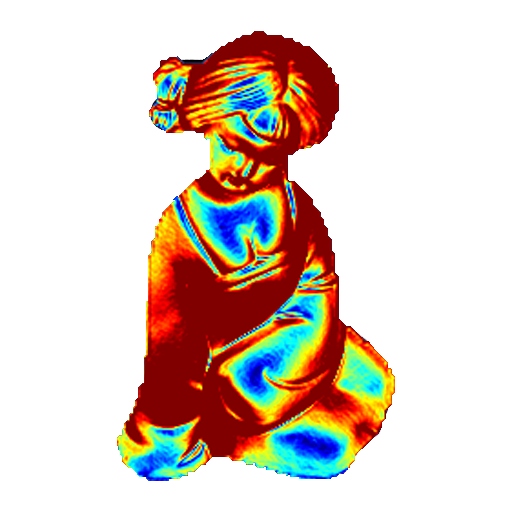}}
		&{\includegraphics[ width=\imgsize, trim=100px 2px 100px 4px, clip=true]{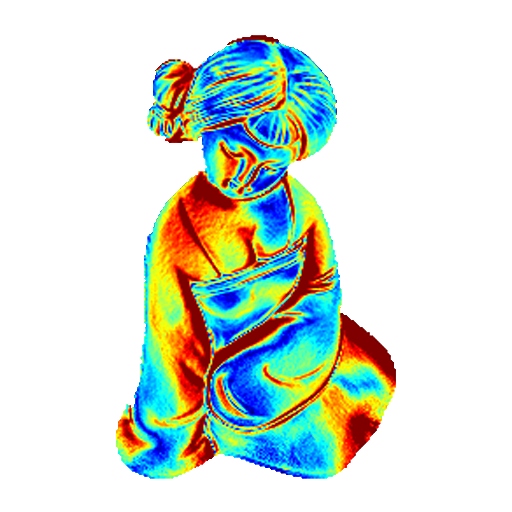}} 
		&{\includegraphics[ width=\imgsize, trim=100px 2px 100px 4px, clip=true]{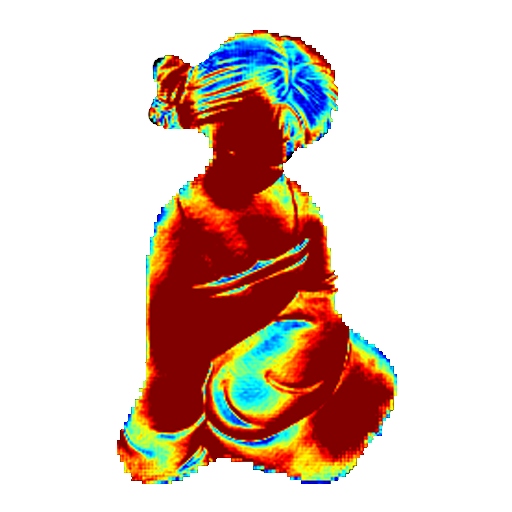}}
		\begin{minipage}{0.030\textwidth} \centering
			\vspace{-20.5em} \makebox[\textwidth]{ \fontsize{20}{16}\selectfont $\,40^{\circ}$}\\ 
			\makebox[0.12\textwidth]{}\includegraphics[width=0.9\textwidth]{fig/quan_com/colorbar.jpg} \\ 
			\makebox[\textwidth]{ \fontsize{20}{16}\selectfont $\,0^{\circ}$}\\
		\end{minipage}
	\end{NiceTabular}	
	}
    \caption{Surface normal estimation of {\sc Pineapple} and {\sc Lady} from baseline methods and ours. We show the estimated surface normals and the corresponding angular error distributions with MAE in degrees shown on the top. } 
	\label{fig:quan_com}
	\vspace{-10pt}
\end{figure*}

\begin{table}[t]
	\begin{center}
    \caption{Comparison On \emph{EvalFlash} and \emph{EvalFlash-synth}. We Calculate the Average Values Over the Two Test Sets.}
    \label{table:overall_quan_com}
    \resizebox{0.48\textwidth}{!}{
    \begin{tabular}{lcccccc}
        \toprule
        &  \multicolumn{3}{c}{\emph{EvalFlash}} & \multicolumn{3}{c}{\emph{EvalFlash-synth}} \\
        Method &  MAE &  RMSE &  MSE &  MAE &  RMSE &  MSE \\
        \midrule
        \ee &\underline{16.02} &\underline{0.33} &\underline{0.04} &17.50 & 0.38& 0.06 \\
        \me &16.37 &0.34 &\underline{0.04} &\underline{16.14} &\underline{0.35} &\underline{0.05} \\
        \stablenormal &18.53 &0.36 &0.05 & 18.48 & 0.38 & 0.06 \\
        \lxm          &38.83 &0.88 &0.27 & 45.39 & 1.02 & 0.38 \\
        \sdmunips     &21.89 &0.44 &0.07 & 26.32 & 0.52 & 0.11 \\
        \mvm          &36.01 &0.72 &0.21 & 55.38 & 1.21 & 0.53 \\ 
        FlashNormal~(Ours) &\textbf{14.96} &\textbf{0.31} & \textbf{0.03} & \textbf{15.47} & \textbf{0.33} & \textbf{0.04} \\
        \bottomrule
    \end{tabular}}
    \end{center}
\end{table}

\begin{table*}[!htbp]
	     \centering
	     \caption{Quantitative Comparison (MAE/RMSE) on Surface Normal Estimation Between Baseline Methods and Ours. The Best and Second-best Results Are Emboldened and Underlined, Respectively. These Are the $10$ Objects Results from \ourdata.}
	     \label{table:quan_com}
	     \resizebox{\linewidth}{!}{
		         \begin{tabular}{lccccccccccc}
			         \toprule Method
			         & {\sc Fortune bag} 
			         & {\sc Pineapple}
			         & {\sc Lady} 
			         & {\sc Pinkcat} 
			         & {\sc Guanyu}
			         & {\sc Eygptcat}
			         & {\sc Shakespeare}
			         & {\sc Monkey}
			         & {\sc Tree}
			         & {\sc Monster}
			         \\
			         \midrule
			         \ee &15.36~/~0.32 &\underline{14.45}~/~\underline{0.29} &19.84~/~0.45 &\textbf{17.22}~/~\textbf{0.35} &20.02~/~0.41 &15.09~/~0.32 &\underline{16.42}~/~\underline{0.34} &\underline{14.52}~/~\underline{0.29} &18.71~/~0.40 &\textbf{12.91}~/~\textbf{0.26} \\ 
			
			         \me &\underline{14.74}~/~\underline{0.31} &16.03~/~0.32 &18.75~/~0.44 &20.31~/~0.41 &\textbf{14.88}~/~\textbf{0.31} &\underline{12.00}~/~\underline{0.27} &18.70~/~0.39 &15.54~/~0.31 &\textbf{15.48}~/~\textbf{0.33} &16.44~/~0.33  \\
			         \stablenormal&15.50~/~0.32&18.27~/~0.35&\underline{18.15}~/~\underline{0.42}&20.45~/~0.39&19.67~/~0.38&15.19~/~0.31&20.82~/~0.42&17.33~/~0.33&18.10~/~0.35&15.99~/~0.31\\
			
			         \lxm &31.70~/~0.70&32.10~/~0.74&40.98~/~0.86&41.92~/~0.86&36.90~/~0.77&40.26~/~0.88&33.49~/~0.75&29.53~/~0.70&36.43~/~0.81&35.61~/~0.83\\ 
			         \sdmunips &21.95~/~0.46&18.51~/~0.36&24.01~/~0.51&36.66~/~0.68&29.79~/~0.58&17.73~/~0.38&19.51~/~0.41&16.54~/~0.33&23.00~/~0.46&16.67~/~0.33\\ 
			         \mvm &25.01~/~0.55&33.19~/~0.69&45.92~/~0.86&44.00~/~0.83&33.42~/~0.68&42.05~/~0.82&33.85~/~0.68&31.76~/~0.67&34.96~/~0.71&38.30~/~0.75\\
			
			         FlashNormal~(Ours) & \textbf{14.68}~/~\textbf{0.30}&\textbf{11.13}~/~\textbf{0.23}&\textbf{17.79}~/~\textbf{0.40}&\underline{18.95}~/~\underline{0.38}&\underline{16.45}~/~\underline{0.33}&\textbf{11.96}~/~\textbf{0.25}&\textbf{14.98}~/~\textbf{0.31}&\textbf{11.52}~/~\textbf{0.24}&\underline{17.01}~/~\underline{0.34}&\underline{14.62}~/~\underline{0.28}\\
			
			         \bottomrule
			         \end{tabular}
		     }
		     \vspace{-5pt}
\end{table*}

\begin{table*}[htbp]
    \centering
    \caption{Quantitative Comparison (MAE/RMSE) on Surface Normal Estimation Between Baseline Methods and Ours. The Best and Second-best Results Are Emboldened and Underlined, Respectively. These are the other $10$ objects results from \ourdata.}
    \label{table:quan_com_suppl}
    \resizebox{\linewidth}{!}{
        \begin{tabular}{lcccccccccc}
        \toprule Method
        & {\sc Hugo} 
        & {\sc Anderson}
        & {\sc Happydog} 
        & {\sc Monkeyhead} 
        & {\sc Luxun}
        & {\sc Sittingbuddha}
        & {\sc Goethe}
        & {\sc Lion}
        & {\sc Persimmon}
        & {\sc Toy}
        \\
        \midrule
        \ee &\underline{14.87}~/~\underline{0.30} &\underline{17.69}~/~\underline{0.36} &14.00~/~0.29 &14.15~/~0.29 &\underline{15.69}~/~\underline{0.33} &\textbf{13.15}~/~\textbf{0.27} &\underline{17.65}~/~\underline{0.37} &18.20~/~\underline{0.37} &\underline{14.58}~/~\underline{0.32} &\textbf{15.51}~/~\textbf{0.34} \\ 
        
        \me  &18.77~/~0.38 &19.00~/~0.38 &\underline{13.28}~/~\underline{0.28} &\underline{13.74}~/~\underline{0.27} &18.89~/~0.38 &15.69~/~0.31 &19.59~/~0.39 &\textbf{15.64}~/~\textbf{0.32} &\textbf{13.12}~/~\textbf{0.29} &16.91~/~0.36  \\
        \stablenormal&20.41~/~0.39&21.33~/~0.42&19.48~/~0.38&17.45~/~0.34&19.84~/~0.38&17.89~/~0.36&20.89~/~0.41&\underline{18.11}~/~\underline{0.37}&17.42~/~0.37&18.44~/~0.39\\
         
        \lxm &30.66~/~0.62&31.61~/~0.64&28.17~/~0.61&49.03~/~0.89&28.82~/~0.63&38.09~/~0.73&30.10~/~0.63&43.89~/~1.04&48.35~/~1.12&50.77~/~1.13\\ 
        \sdmunips &23.62~/~0.46&21.79~/~0.44&21.02~/~0.43&14.55~/~0.29&17.87~/~0.36&22.61~/~0.43&21.70~/~0.43&22.86~/~0.47&17.02~/~0.36&31.34~/~0.61\\ 
        \mvm &33.59~/~0.85&41.31~/~0.96&36.74~/~0.90&43.29~/~0.98&32.41~/~0.93&52.38~/~1.11&34.93~/~0.89&33.09~/~0.69&40.67~/~0.81&47.46~/~0.90\\
        FlashNormal(ours) &\textbf{13.79}~/~\textbf{0.29}&\textbf{14.55}~/~\textbf{0.31}&\textbf{13.13}~/~\textbf{0.27}&\textbf{13.60}~/~\textbf{0.26}&\textbf{14.61}~/~\textbf{0.31}&\underline{14.08}~/~\underline{0.28}&\textbf{15.66}~/~\textbf{0.33}&18.21~/~\underline{0.37}&15.68~/~\underline{0.32}&\underline{16.83}~/~\underline{0.35}\\

        \bottomrule
        \end{tabular}
    }
    \vspace{-5pt}
\end{table*}

\begin{table*}[!htbp]
    \centering
    \caption{Quantitative Comparison (MSE) on Surface Normal Estimation Between Baseline Methods and Ours. The Best and Second-best Results Are Emboldened and Underlined, Respectively.These Are the $10$ Objects Results from \ourdata.}
    \label{table:quan_com_mse1}
    \resizebox{\linewidth}{!}{
        \begin{tabular}{lcccccccccc}
        \toprule Method
        & {\sc Fortune bag} 
        & {\sc Pineapple}
        & {\sc Lady} 
        & {\sc Pinkcat} 
        & {\sc Guanyu}
        & {\sc Eygptcat}
        & {\sc Shakespeare}
        & {\sc Monkey}
        & {\sc Tree}
        & {\sc Monster}
        \\
        \midrule
        \ee &\textbf{0.03} &\underline{0.03} &0.07 &0.08 &0.05 &\underline{0.03} &\underline{0.04} &\underline{0.03} &\underline{0.05} &\textbf{0.02} \\ 
        
        \me  &\underline{0.04} &\underline{0.03} &\underline{0.06} &\underline{0.06} &\textbf{0.03} &\textbf{0.02} &0.05 &\underline{0.03} &\textbf{0.04} &0.04 \\
        \stablenormal &\underline{0.04} &0.05 &0.10 &0.07 &0.06 &\underline{0.03} &0.05 &\underline{0.03} &0.06 &0.06\\
         
        \lxm &0.18 &0.27 &0.34 &0.24 &0.29 &0.35 &0.27 &0.22 &0.33 &0.28\\ 
        \sdmunips &0.05 &0.06 &0.09 &0.15 &0.11 &0.06 &0.06 &0.04 &0.06 &0.04\\ 
        \mvm &0.10 &0.14 &0.24 &0.19 &0.12 &0.23 &0.16 &0.14 &0.15 &0.19\\
        FlashNormal(ours) &\textbf{0.03} &\textbf{0.01} &\textbf{0.05} &\textbf{0.05} &\underline{0.04} &\textbf{0.02} &\textbf{0.03} &\textbf{0.02} &\textbf{0.04} &\underline{0.03}\\

        \bottomrule
        \end{tabular}
    }
    \vspace{-5pt}
\end{table*}

\begin{table*}[!htbp]
    \centering
    \caption{Quantitative Comparison (MSE) on Surface Normal Estimation Between Baseline Methods and Ours. The Best and Second-best Results Are Emboldened and Underlined, Respectively. These Are the Other $10$ Objects Results from \ourdata.}
    \label{table:quan_com_mse2}
    \resizebox{\linewidth}{!}{
        \begin{tabular}{lcccccccccc}
        \toprule Method
        & {\sc Hugo} 
        & {\sc Anderson}
        & {\sc Happydog} 
        & {\sc Monkeyhead} 
        & {\sc Luxun}
        & {\sc Sittingbuddha}
        & {\sc Goethe}
        & {\sc Lion}
        & {\sc Persimmon}
        & {\sc Toy}
        \\
        \midrule
        \ee &\underline{0.03} &\underline{0.04} &\underline{0.03} &\underline{0.03} &\underline{0.04} &\textbf{0.02} &\underline{0.04} &\underline{0.04} &\textbf{0.03} &\textbf{0.04} \\ 
        
        \me  &0.04 &\underline{0.04} &\underline{0.03} &\textbf{0.02} &0.05 &\underline{0.03} &0.05 &\textbf{0.03} &\textbf{0.03} &\textbf{0.04} \\
        \stablenormal &0.04 &0.05 &\underline{0.03} &\underline{0.03} &\underline{0.04} &0.04 &0.05 &0.07 &\underline{0.05} &\underline{0.07}\\
         
        \lxm &0.21 &0.26 &0.19 &0.24 &0.22 &0.33 &0.24 &0.29 &0.30 &0.33\\ 
        \sdmunips &0.07 &0.08 &0.05 &\textbf{0.02} &0.06 &0.05 &0.06 &0.07 &\underline{0.05} &0.12\\ 
        \mvm &0.24 &0.26 &0.25 &0.26 &0.28 &0.31 &0.29 &0.17 &0.23 &0.27\\
        FlashNormal(ours) &\textbf{0.02} &\textbf{0.03} &\textbf{0.02} &\textbf{0.02} &\textbf{0.03} &\textbf{0.02} &\textbf{0.03} &\underline{0.04} &\textbf{0.03} &\textbf{0.04}\\

        \bottomrule
        \end{tabular}
    }
    \vspace{-5pt}
\end{table*}

As shown in \Tref{table:overall_quan_com}, we present the average metric values on $20$ objects. Compared to the second-best method \ee, \ours improves the surface normal estimation accuracy by $6.6\%$. Compared with the state-of-the-art flash/no-flash surface normal estimation method MV20~\cite{2020Two}, we improve performance by $58.4\%$, demonstrating the effectiveness of \ours. \Tref{table:quan_com} and \Tref{table:quan_com_suppl} report the per-object quantitative results of $20$ objects of \ourdata. For certain objects, \ee demonstrate superior performance compared to \me, whereas \me outperforms them on others. On the other hand, our method consistently achieves the smallest MAE and RMSE across most tested objects, showcasing its stability in estimating surface normals for objects with rich details. We also show the MSE value across $20$ objects in EvalFlash, as shown in \Tref{table:quan_com_mse1} and \Tref{table:quan_com_mse2}.

In addition, we show a qualitative comparison in \fref{fig:quan_com} on {\sc Pineapple} and {\sc Lady}, representing specular and diffuse surfaces, respectively. While existing methods like \ee and \stablenormal perform well on only one of the two, our \ours robustly estimates plausible surface normals for both, demonstrating its adaptability to diverse reflectances. Both \me and \mvm can only produce coarse estimations, while ours can capture fine geometric details. \sdmunips produces blurry predictions, likely due to an insufficient number of multi-light image inputs.

\subsection{Evaluation on EvalFlash-synth}
We use the large-scale synthetic test set EvalFlash-synth to comprehensively compare the robustness and generalizability of different methods. \Tref{table:overall_quan_com} presents the quantitative results. Our proposed \ours achieves the smallest MAE and RMSE on the average results across $839$ test samples. Compared to the second-best method \me, \ours improves the surface normal estimation accuracy by $4.2\%$. Compared with the state-of-the-art flash/no-flash surface normal estimation method MV20~\cite{2020Two}, we achieve a performance gain of $72.1\%$. 
\vspace{-5pt}
\subsection{Evaluation on causally captured images}
\begin{figure*}[!t]
\centering
\begin{adjustbox}{max width=\linewidth}
\addtolength{\tabcolsep}{-0.4em}
\begin{NiceTabular}{ccccccccc}
    & 
    \multicolumn{2}{c}{ Flash/no-flash images} & 
    \multicolumn{2}{c}{ \makecell{FlashNormal~(Ours)}} & 
    \multicolumn{2}{c}{ \ee} &
    \multicolumn{2}{c}{ \me} \\

    \raisebox{0.05\height}{\rotatebox{90}{ \makecell{\sc Textured box}}} &
    \includegraphics[width=0.16\textwidth]{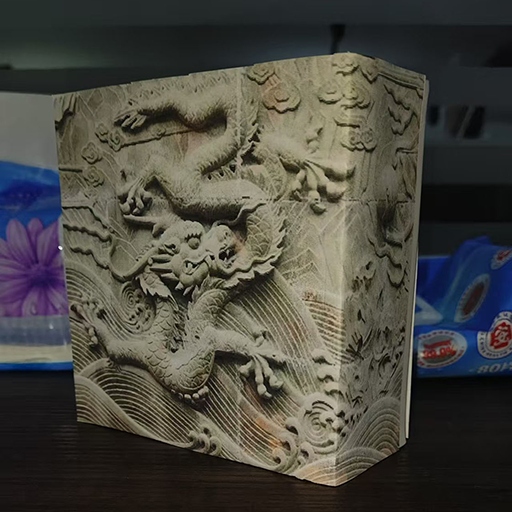} &
    \includegraphics[width=0.16\textwidth]{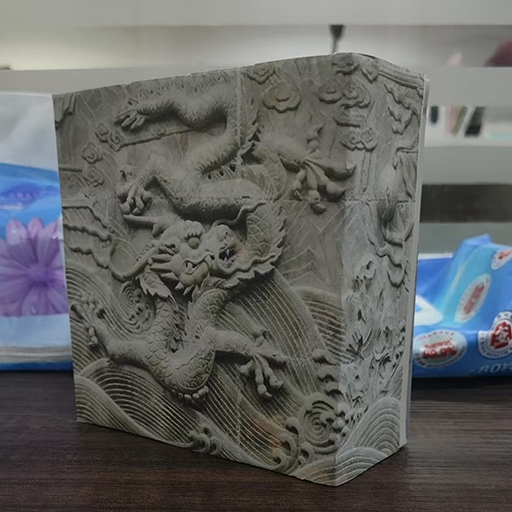} &
    \includegraphics[width=0.16\textwidth]{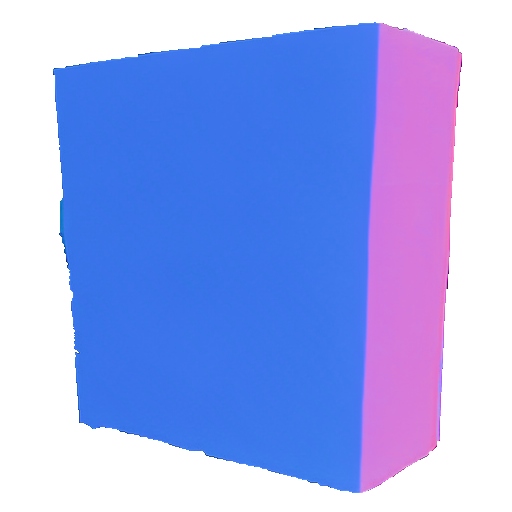} &
    \includegraphics[width=0.16\textwidth]{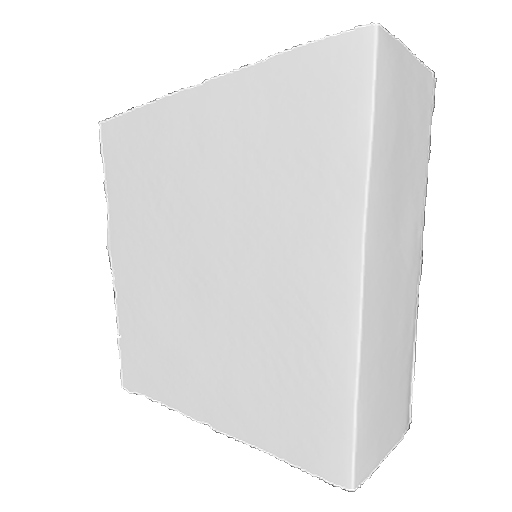} &
    \includegraphics[width=0.16\textwidth]{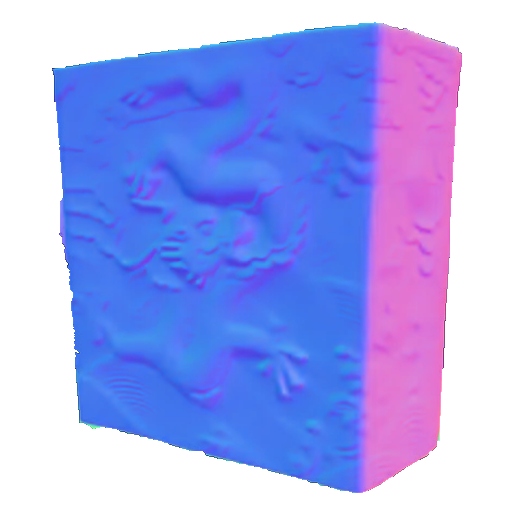} &
    \includegraphics[width=0.16\textwidth]{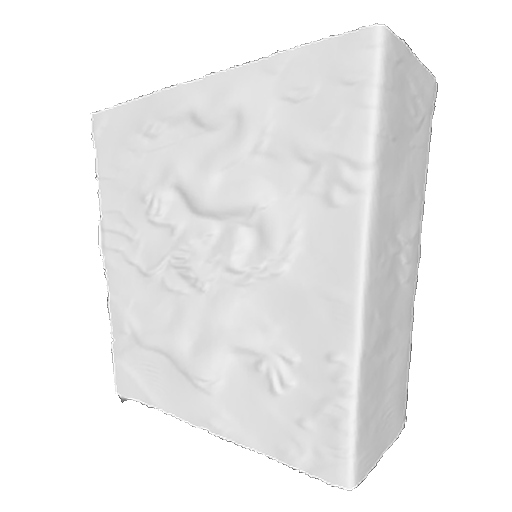} &
    \includegraphics[width=0.16\textwidth]{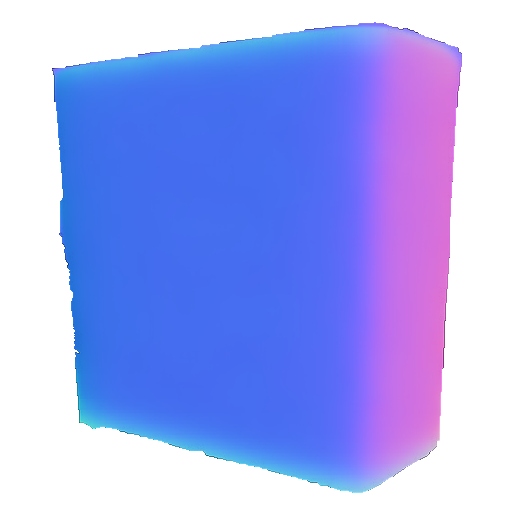} &
    \includegraphics[width=0.16\textwidth]{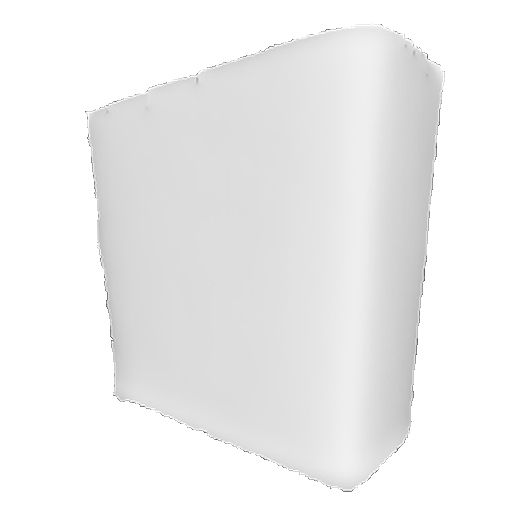} \\
    
    \raisebox{0.4\height}{\rotatebox{90}{ \sc Mirror}} &
    \includegraphics[width=0.16\textwidth]{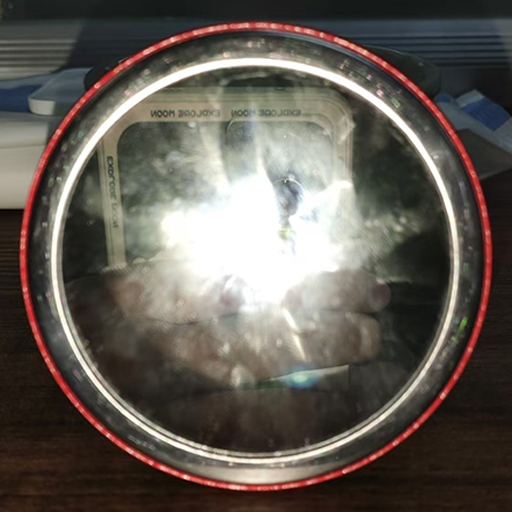} &
    \includegraphics[width=0.16\textwidth]{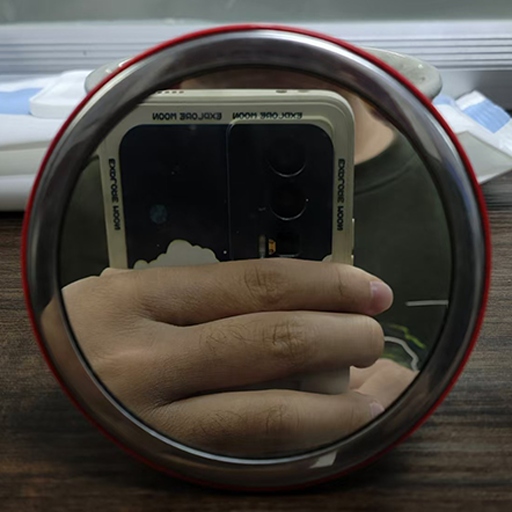} &
    \includegraphics[width=0.16\textwidth]{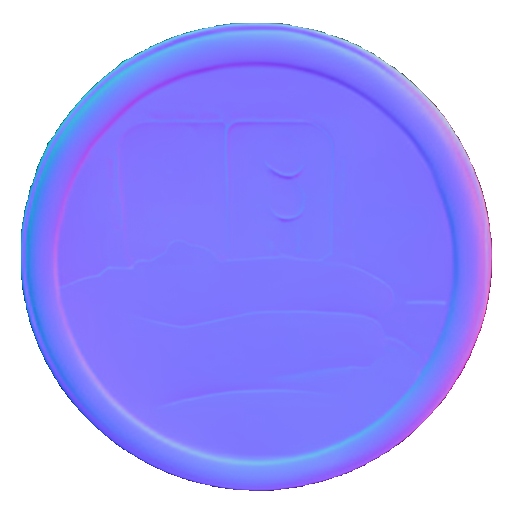} &
    \includegraphics[width=0.16\textwidth]{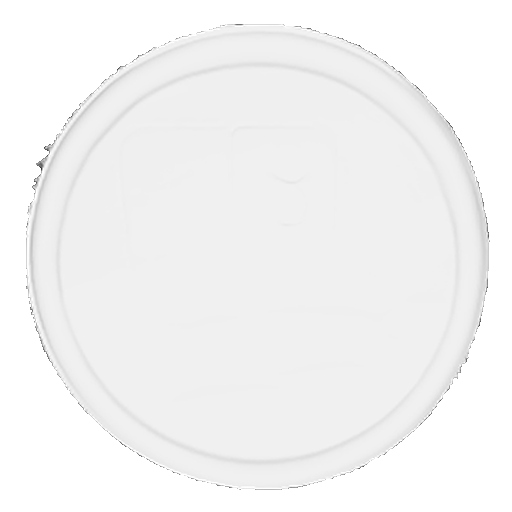} &
    \includegraphics[width=0.16\textwidth]{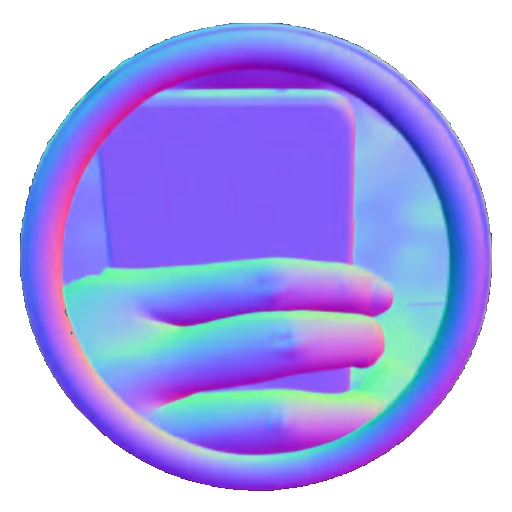} &
    \includegraphics[width=0.16\textwidth]{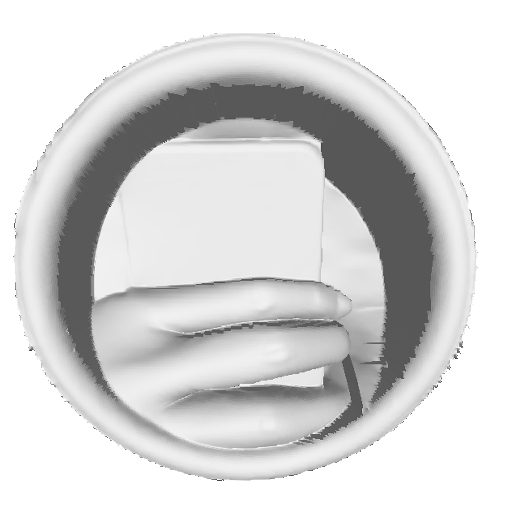} &
    \includegraphics[width=0.16\textwidth]{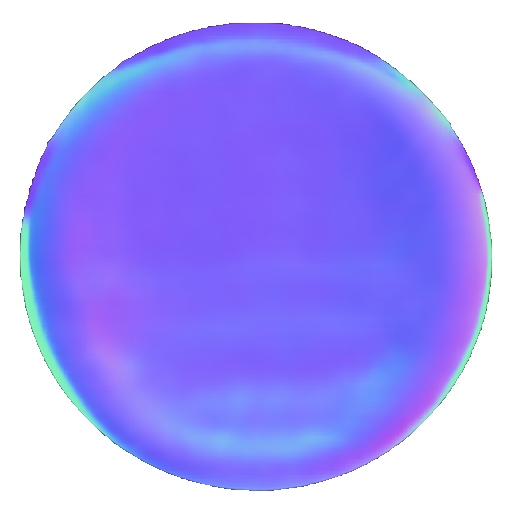} &
    \includegraphics[width=0.16\textwidth]{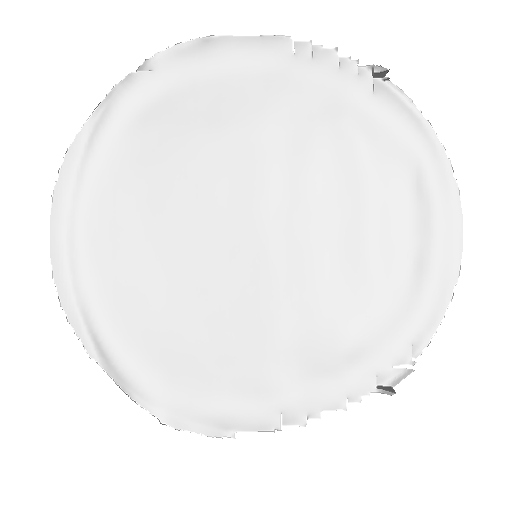} \\

	\raisebox{0.15\height}{\rotatebox{90}{\makecell{\sc Drum poster}}} &
	\includegraphics[width=0.16\textwidth]{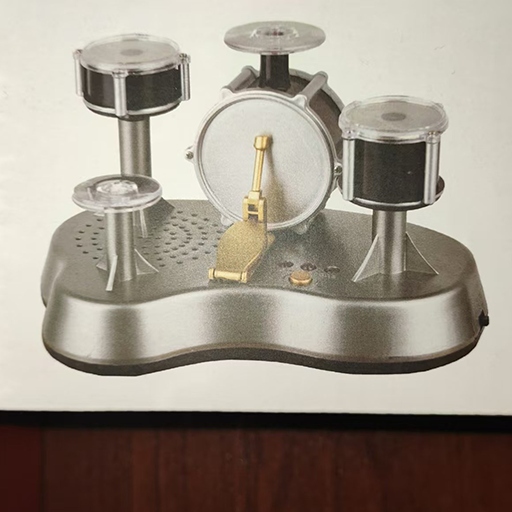} &
	\includegraphics[width=0.16\textwidth]{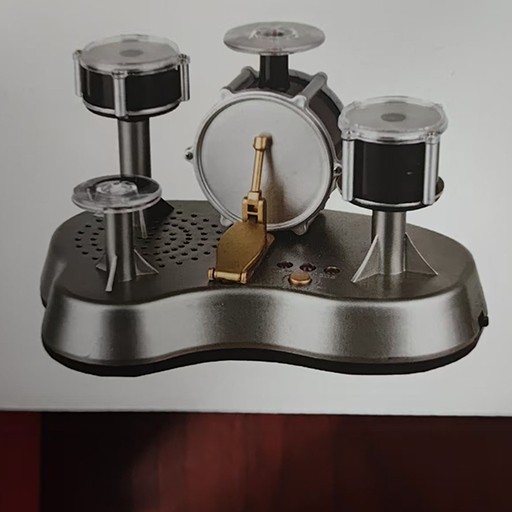} &
	\includegraphics[width=0.16\textwidth]{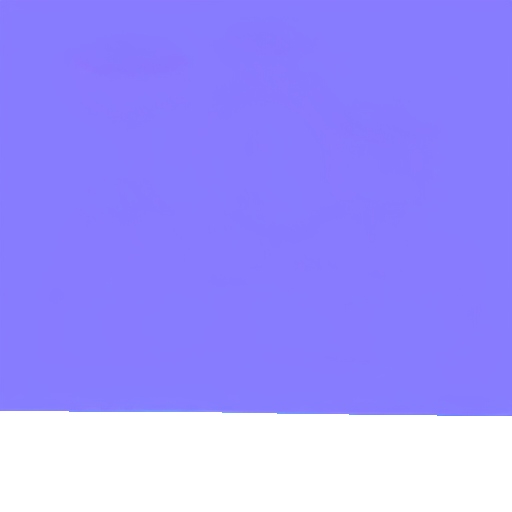} &
	\includegraphics[width=0.16\textwidth]{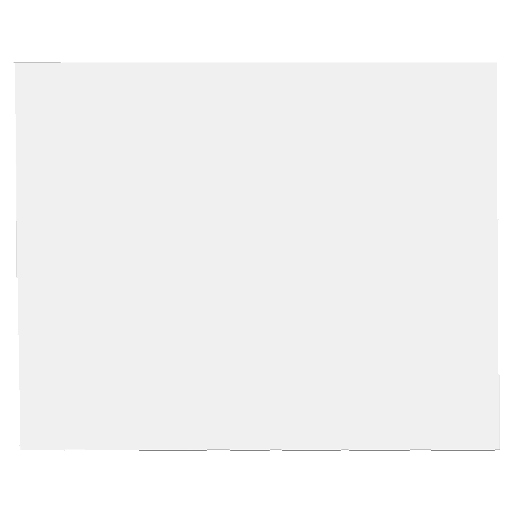} &
	\includegraphics[width=0.16\textwidth]{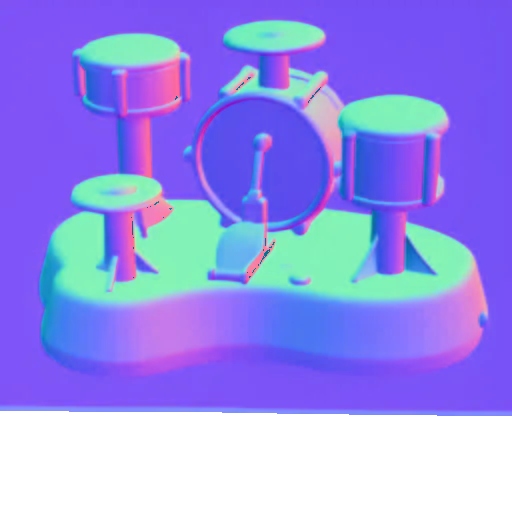} &
	\includegraphics[width=0.16\textwidth]{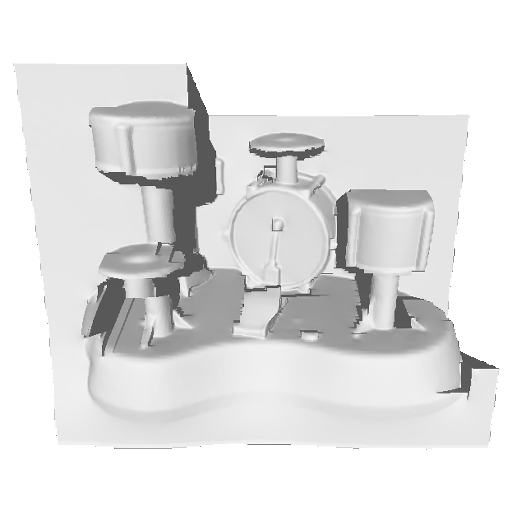} &
	\includegraphics[width=0.16\textwidth]{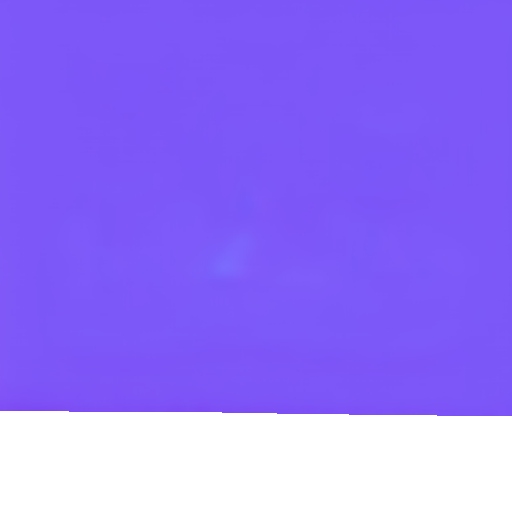} &
	\includegraphics[width=0.16\textwidth]{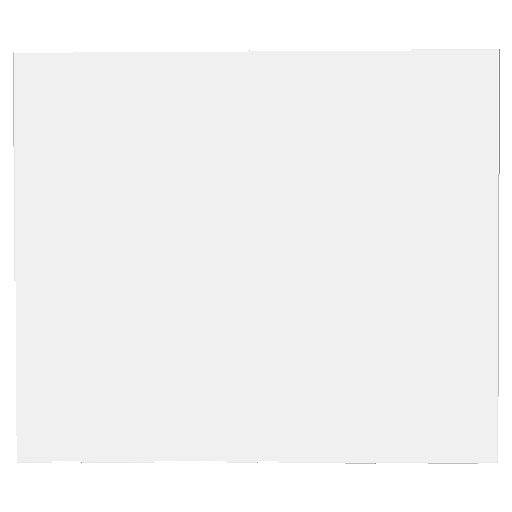} \\
	
	\raisebox{0.15\height}{\rotatebox{90}{\makecell{\sc Fan Poster}}} &
	\includegraphics[width=0.16\textwidth]{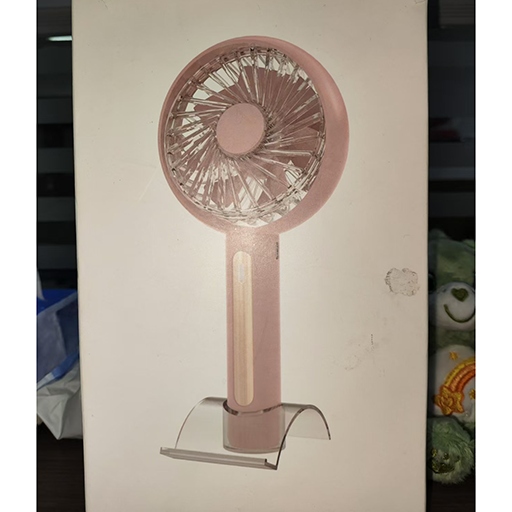} &
	\includegraphics[width=0.16\textwidth]{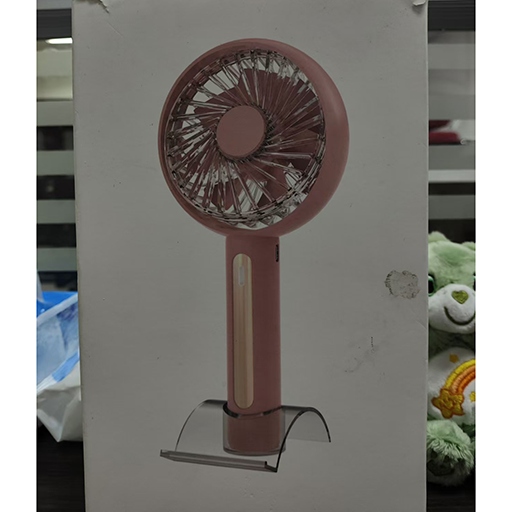} &
	\includegraphics[width=0.16\textwidth]{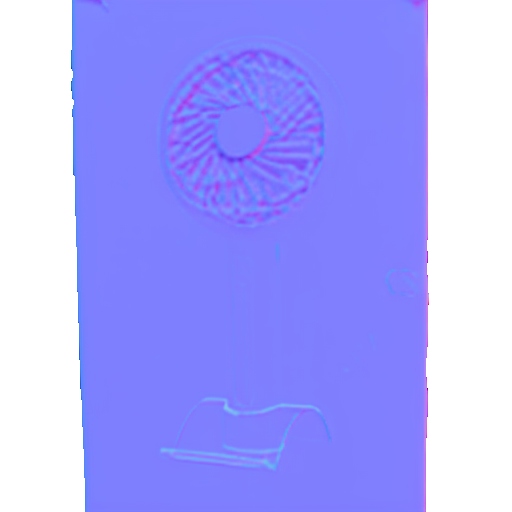} &
	\includegraphics[width=0.16\textwidth]{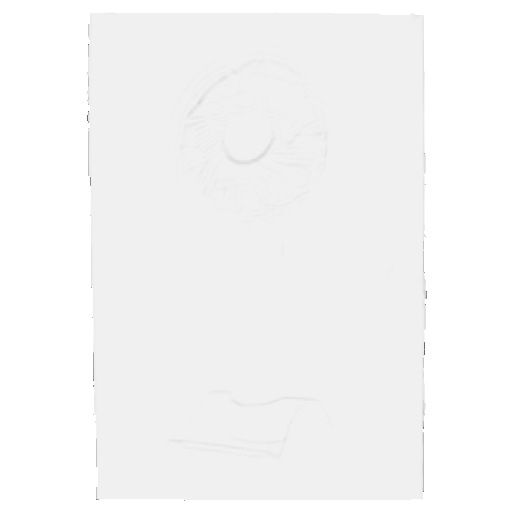} &
	\includegraphics[width=0.16\textwidth]{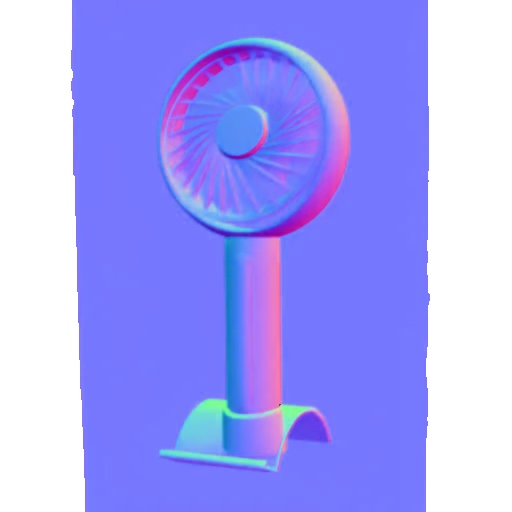} &
	\includegraphics[width=0.16\textwidth]{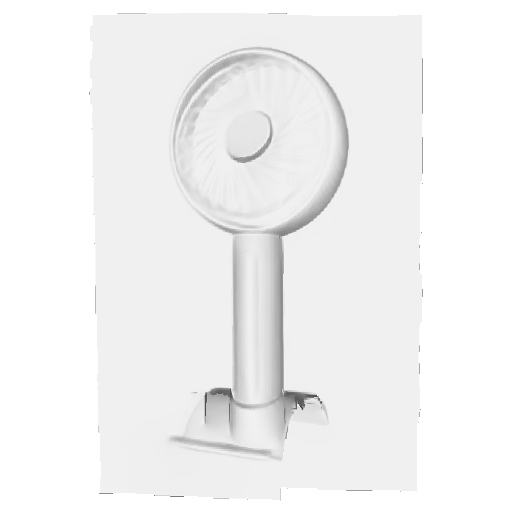} &
	\includegraphics[width=0.16\textwidth]{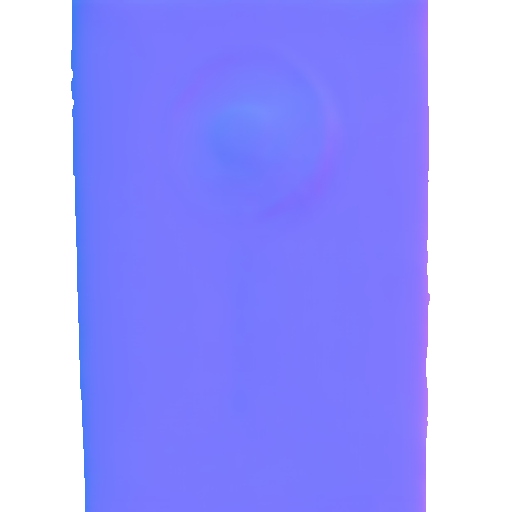} &
	\includegraphics[width=0.16\textwidth]{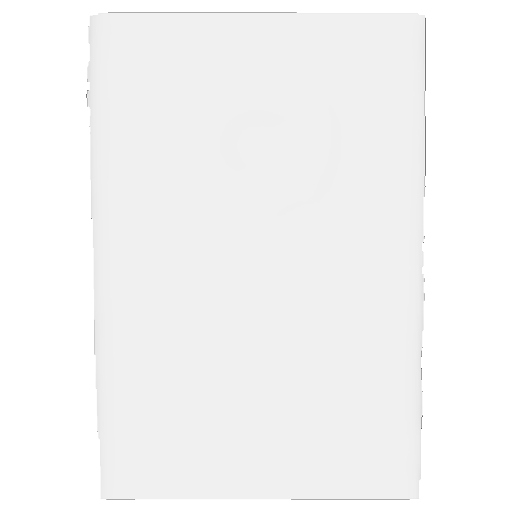} \\
\end{NiceTabular}
\end{adjustbox}
\caption{Qualitative comparison with baseline methods in terms of shape-reflectance ambiguity. Our method can estimate high-quality ambiguous-less surface normal maps, leading to visibly flatter meshes after surface normal integration.} 
\label{fig:qua_com_ambiguity}
\vspace{-8pt}
\end{figure*}

\begin{figure*}[!t]
\centering
\begin{adjustbox}{max width=\linewidth}
\addtolength{\tabcolsep}{-0.4em}
\begin{NiceTabular}{ccccccccc}
    & \multicolumn{2}{c}{Flash/no-flash images} & 
    \multicolumn{2}{c}{\makecell{FlashNormal~(Ours)}} & 
    \multicolumn{2}{c}{\ee} &
    \multicolumn{2}{c}{\me} \\

    \raisebox{0.4\height}{\rotatebox{90}{\sc Ganesha}} &
    \includegraphics[width=0.16\textwidth]{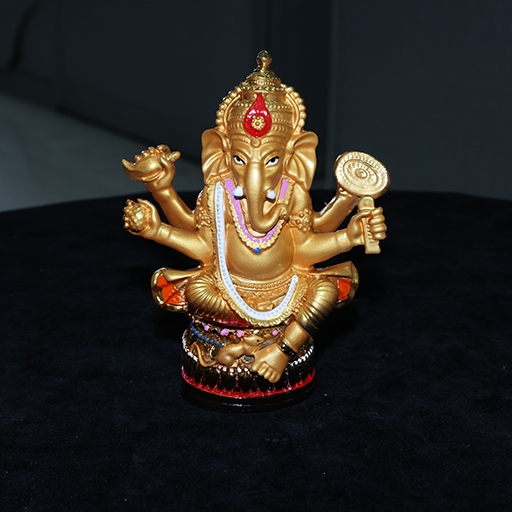} &
    \includegraphics[width=0.16\textwidth]{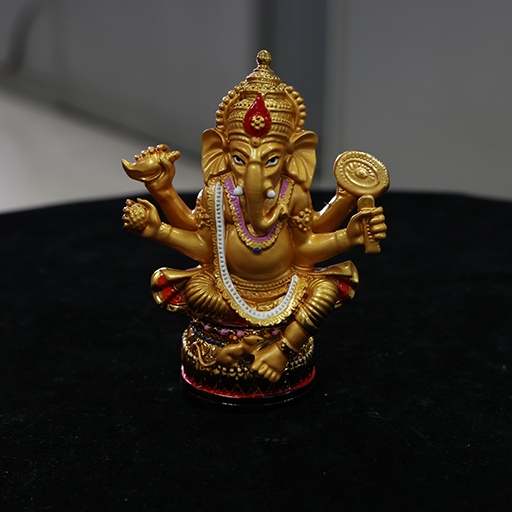} &
    \includegraphics[width=0.16\textwidth]{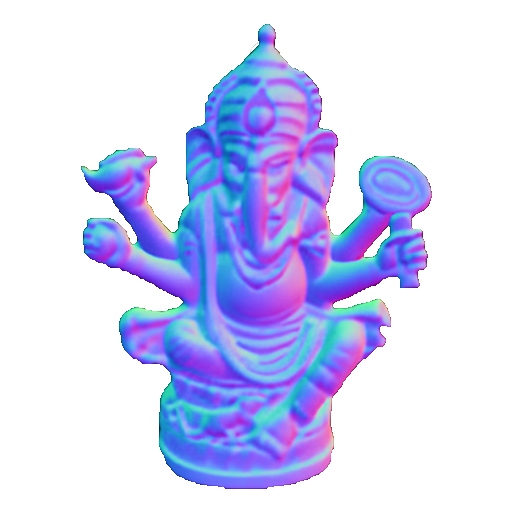} &
    \includegraphics[width=0.16\textwidth]{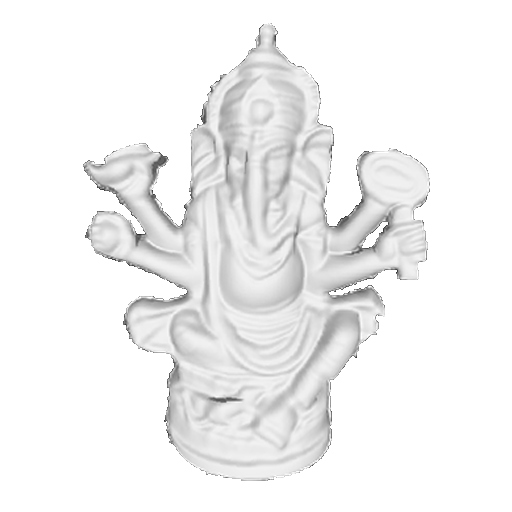} &
    \includegraphics[width=0.16\textwidth]{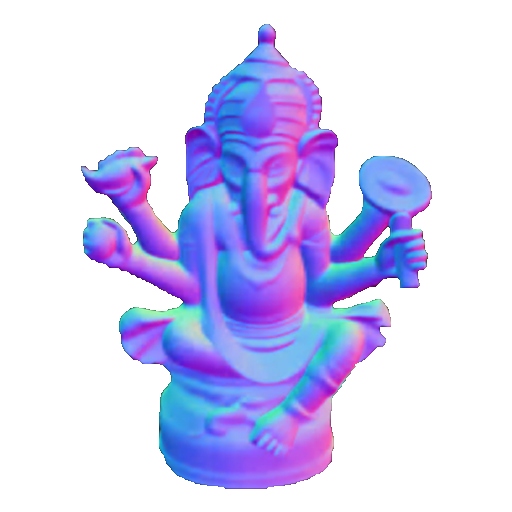} &
    \includegraphics[width=0.16\textwidth]{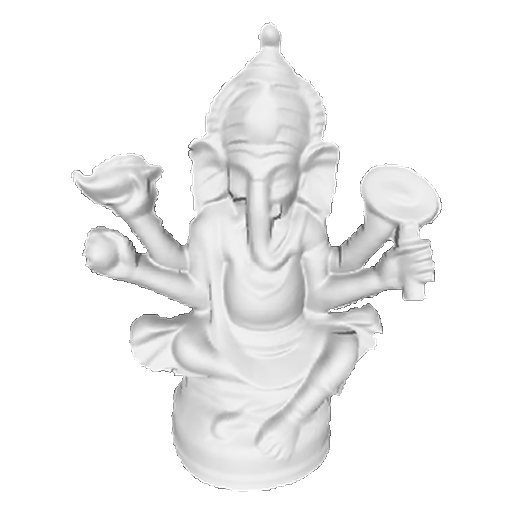} &
    \includegraphics[width=0.16\textwidth]{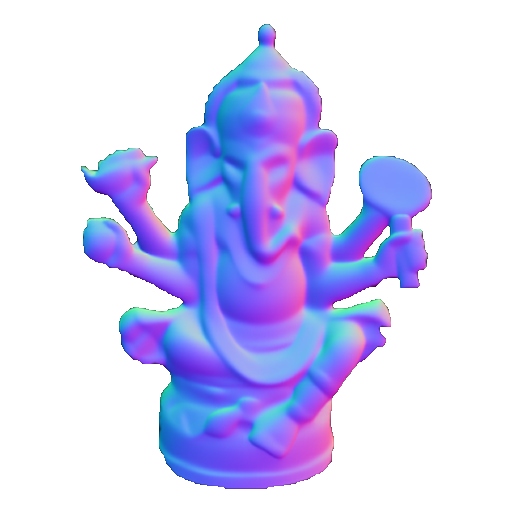} &
    \includegraphics[width=0.16\textwidth]{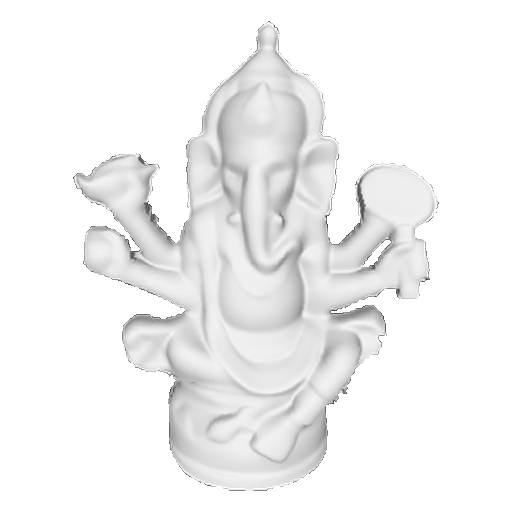} \\
    
    \raisebox{0.8\height}{\rotatebox{90}{\makecell{\sc Santa}}} &
    \includegraphics[width=0.16\textwidth]{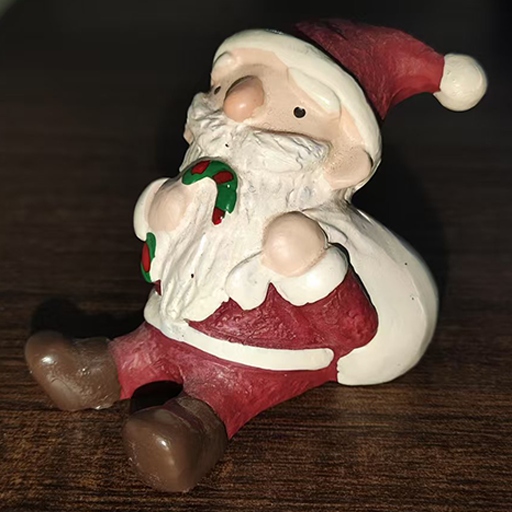} &
    \includegraphics[width=0.16\textwidth]{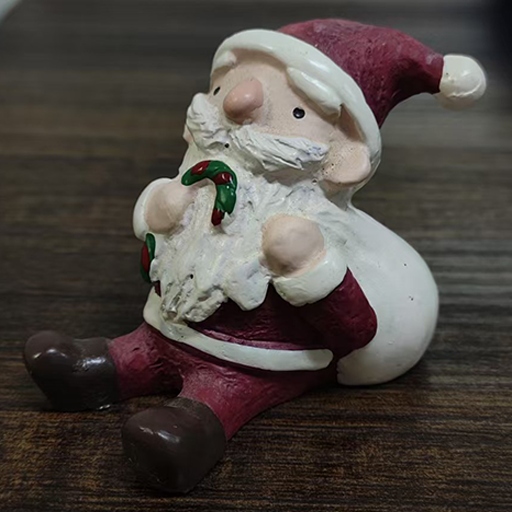} &
    \includegraphics[width=0.16\textwidth]{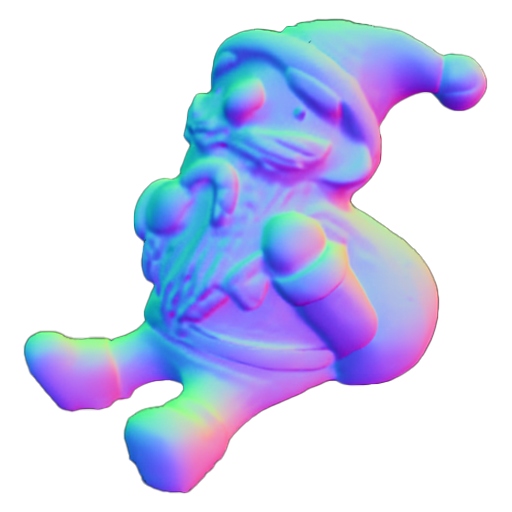} &
    \includegraphics[width=0.16\textwidth]{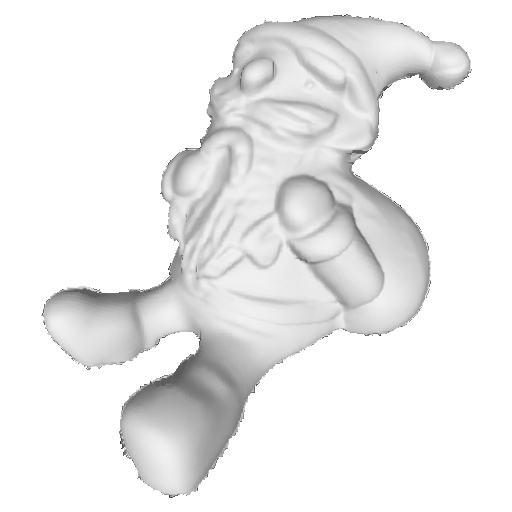} &
    \includegraphics[width=0.16\textwidth]{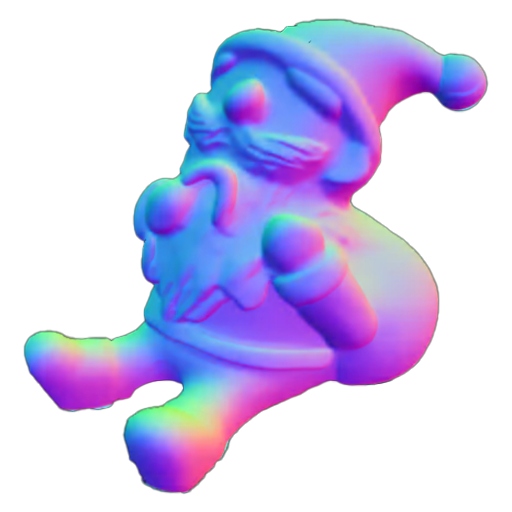} &
    \includegraphics[width=0.16\textwidth]{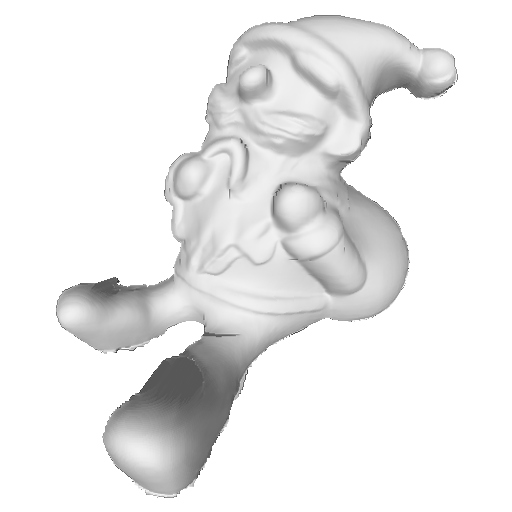} &
    \includegraphics[width=0.16\textwidth]{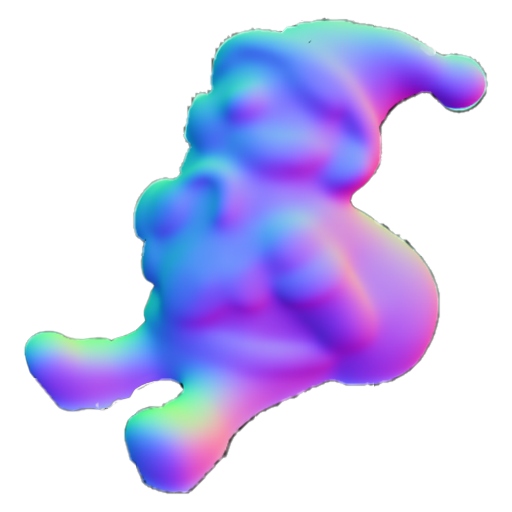} &
    \includegraphics[width=0.16\textwidth]{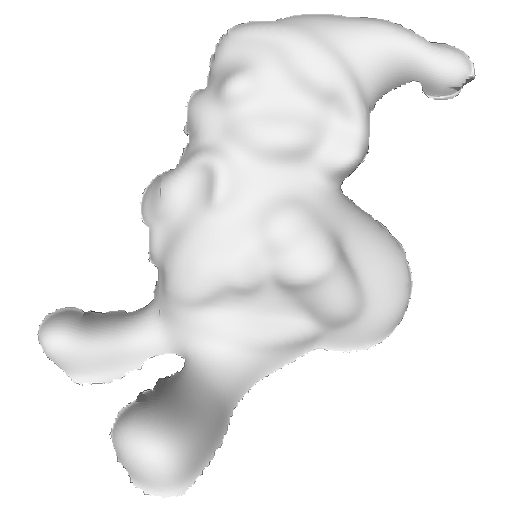} \\
    
    \raisebox{0.5\height}{\rotatebox{90}{\makecell{\sc Squirrel}}} &
    \includegraphics[width=0.16\textwidth]{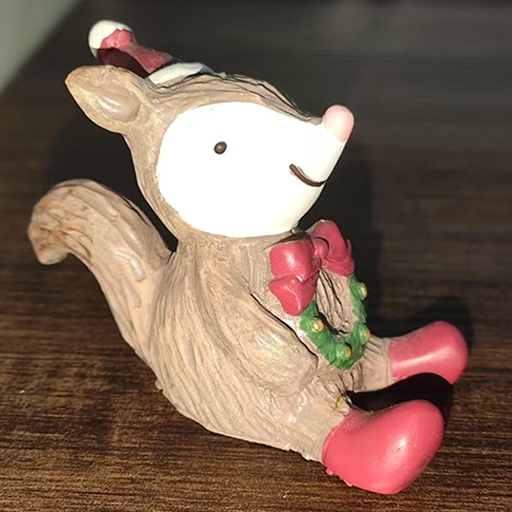} &
    \includegraphics[width=0.16\textwidth]{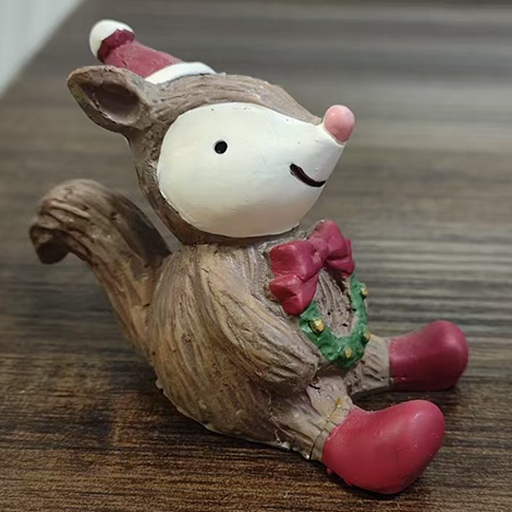} &
    \includegraphics[width=0.16\textwidth]{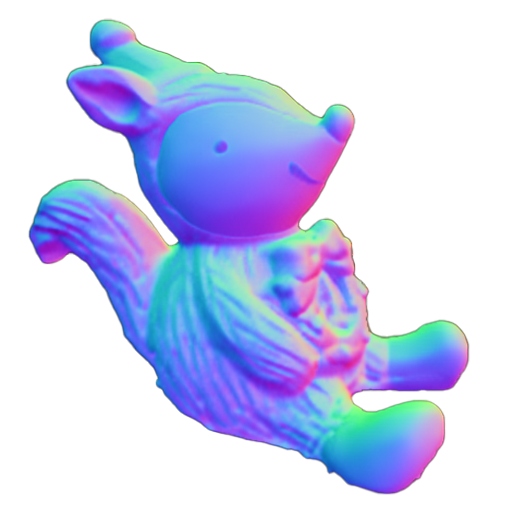} &
    \includegraphics[width=0.16\textwidth]{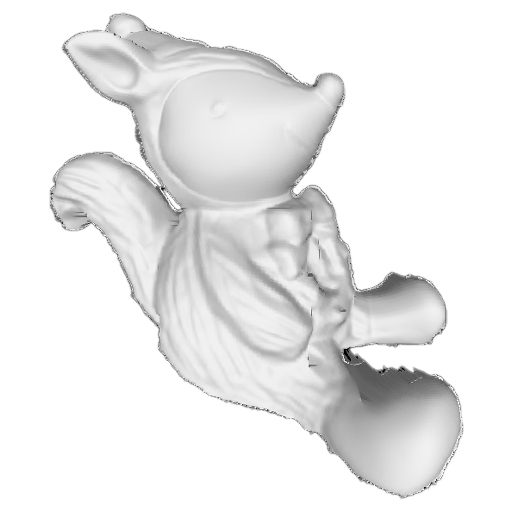} &
    \includegraphics[width=0.16\textwidth]{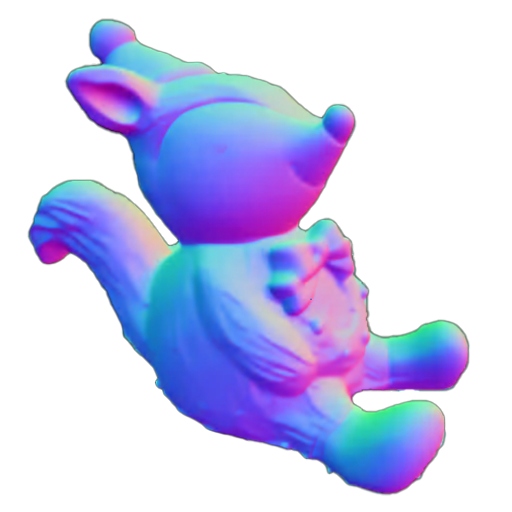} &
    \includegraphics[width=0.16\textwidth]{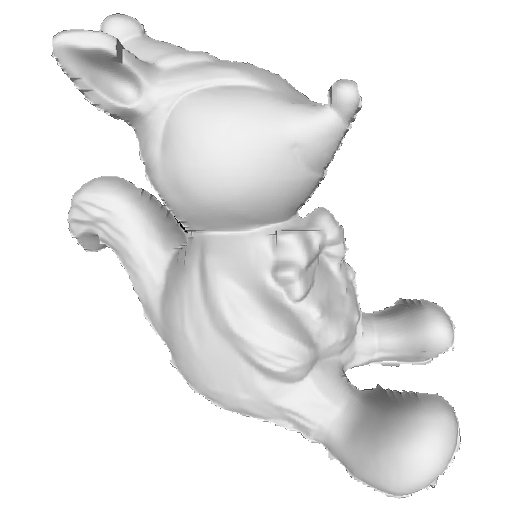} &
    \includegraphics[width=0.16\textwidth]{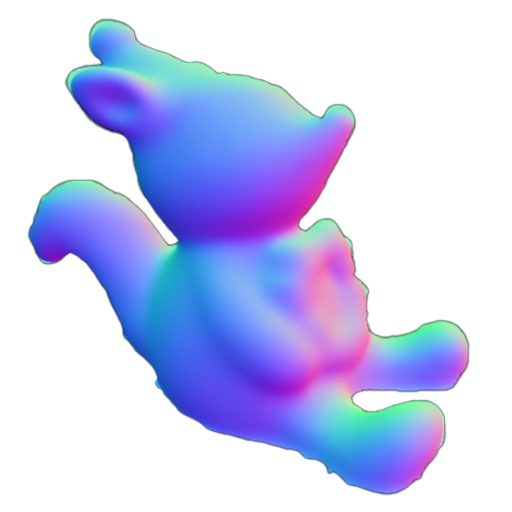} &
    \includegraphics[width=0.16\textwidth]{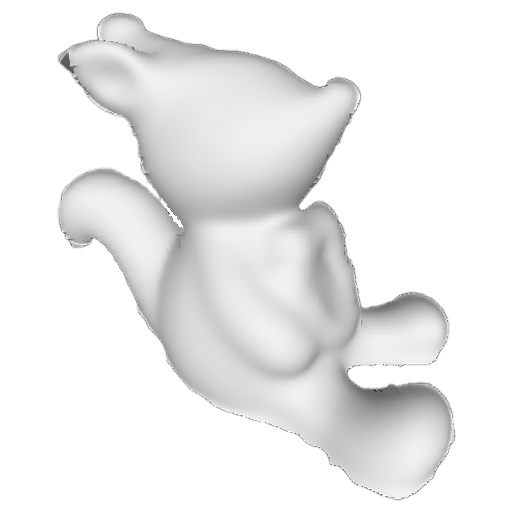} \\
\end{NiceTabular}
\end{adjustbox}
\caption{Estimated surface normal maps and reconstructed meshes of detail-rich objects. Compared to the peer methods, our estimated surface normals present richer details, leading to higher-quality structured meshes.}
\label{fig:qua_com_detail_suppl}
\end{figure*}

Besides the comparison on \ourdata, which is captured by a high-end camera, we also evaluate \ours on flash/no-flash images causally captured by a smartphone~(Redmi K60) to show the robustness of our proposal \wrt capturing devices. 

As shown in \fref{fig:qua_com_ambiguity}, we show the influence of shape-reflectance ambiguity on single-image surface normal estimation. From a single image, \ee cannot disentangle the shape and image texture shown on near-planar boxes or the mirror. As a result, the recoveries are heavily deviated from their actual shapes, as demonstrated by their surface normals and the visualized meshes integrated from the normal maps~\cite{bini2022cao}. On the other hand, \me can estimate more realistic surface normals, despite the results are also relatively blurry. Given the diffusion prior and shading variations from flash image pairs, our \ours appears to be effective in reducing the shape-reflectance ambiguity and achieves plausible surface normal estimation. This demonstrates that by introducing flash inputs, our method effectively reduces shape-reflectance ambiguity, thanks to shading variations.

To assess detailed surface shape recovery, we capture three objects with rich geometric features. As shown in \fref{fig:qua_com_detail_suppl},  both \ee and \me fail to capture fine details in their surface normals, as seen in the normal maps and integrated meshes. This highlights the importance of shadow and specular distribution in flash image pairs for detail recovery. Additionally, curvature map supervision during training helps \ours enhance geometric detail reconstruction. For {\sc Santa} and {\sc Squirrel}, which have intricate surface textures, our method outperforms others by capturing complex details in the estimated surface normals. The meshes generated from these normals reveal pronounced protrusions, closely resembling the real objects. In contrast, competing methods produce smoother normals and meshes, indicating their limitations in handling highly detailed surfaces.

\subsection{Comparison with single-image-based mesh generation methods}
\begin{figure}
\renewcommand{\imgsize}{0.2}
\begin{adjustbox}{max width=\linewidth}
\addtolength{\tabcolsep}{-0.4em}
\begin{NiceTabular}{cccc:ccc}
    & \makecell{Input image} &
    \makecell{Front view} & 
    \makecell{Side view}  &
    \makecell{Input image} &
    \makecell{Front view} & 
    \makecell{Side view}
      \\
    
    \raisebox{0.28\height}{\rotatebox{90}{\makecell{GT mesh}}} &
    \includegraphics[width=\imgsize\linewidth]{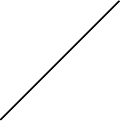}	&
    \includegraphics[width=\imgsize\linewidth, trim=60px 2px 60px 4px, clip=true]{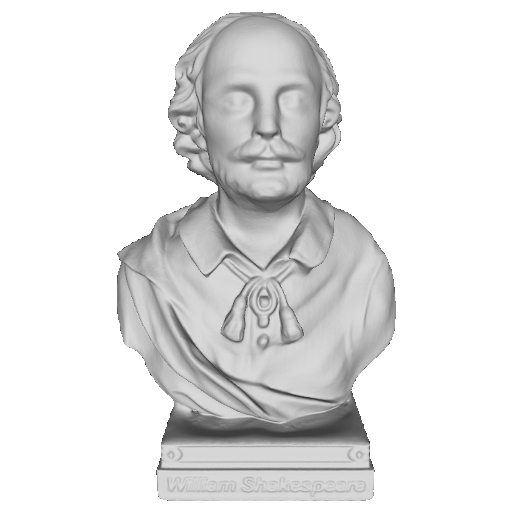} &
    \includegraphics[width=\imgsize\linewidth, trim=60px 2px 60px 4px, clip=true]{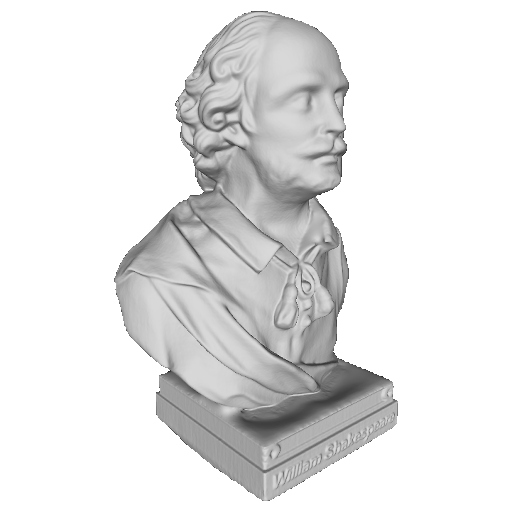} &
    \includegraphics[width=\imgsize\linewidth]{fig/no.drawio.pdf}	&
    \includegraphics[width=\imgsize\linewidth, trim=60px 2px 60px 4px, clip=true]{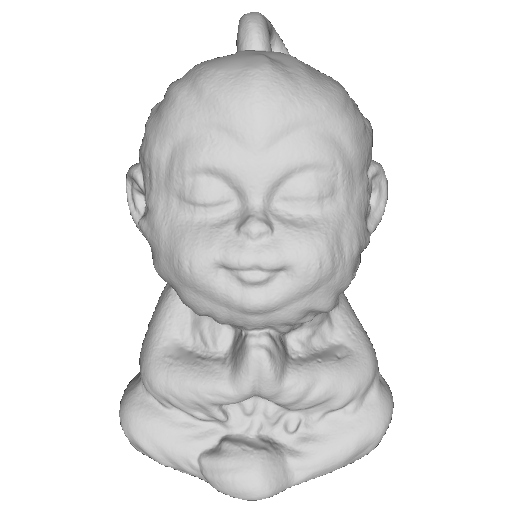} &
    \includegraphics[width=\imgsize\linewidth, trim=60px 2px 60px 4px, clip=true]{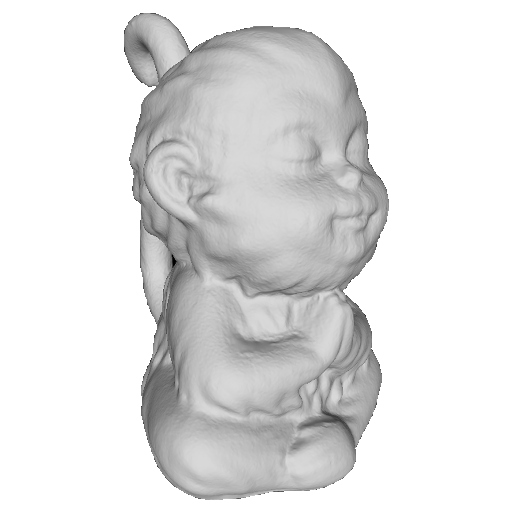} 
    \\

    \raisebox{0.1\height}{\rotatebox{90}{\makecell{FlashNormal\\(Ours)}}} & 
    \includegraphics[width=\imgsize\linewidth]{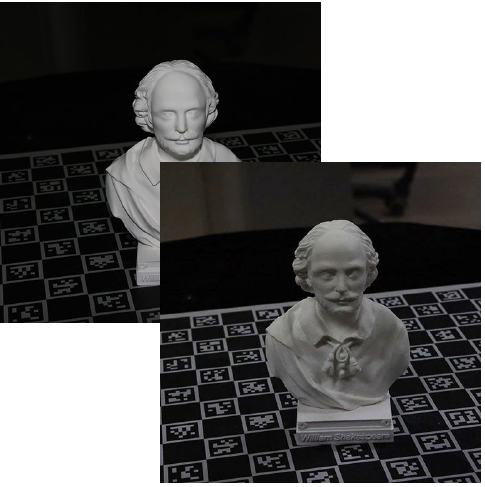} &
    \includegraphics[width=\imgsize\linewidth, trim=60px 2px 60px 4px, clip=true]{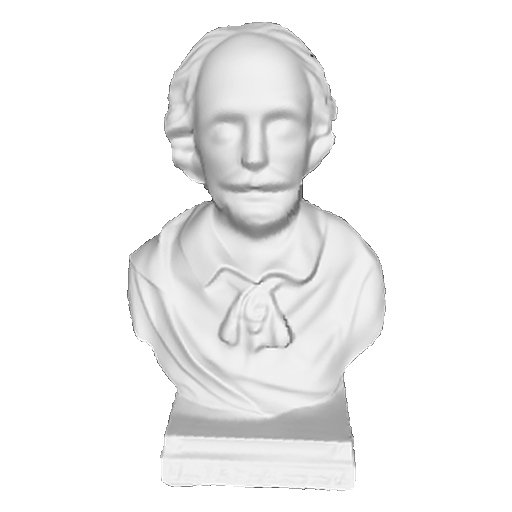} &
    \includegraphics[width=\imgsize\linewidth, trim=60px 2px 60px 4px, clip=true]{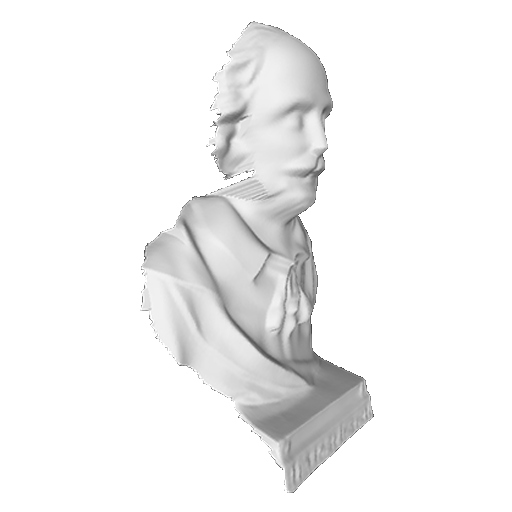}  &
    \includegraphics[width=\imgsize\linewidth]{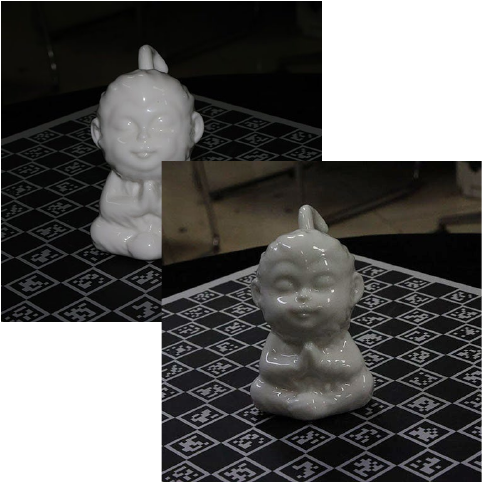} &
    \includegraphics[width=\imgsize\linewidth, trim=60px 2px 60px 4px, clip=true]{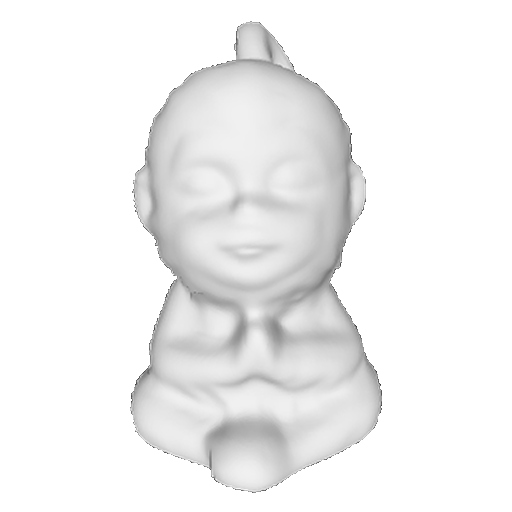} &
    \includegraphics[width=\imgsize\linewidth, trim=60px 2px 60px 4px, clip=true]{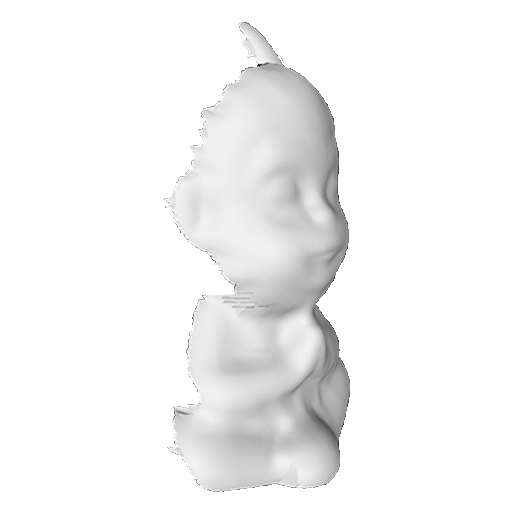} 
    \\
    
    \raisebox{0.1\height}{\rotatebox{90}{\makecell{CLAY~\cite{zhang2024clay}}}} & 
    \includegraphics[width=\imgsize\linewidth]{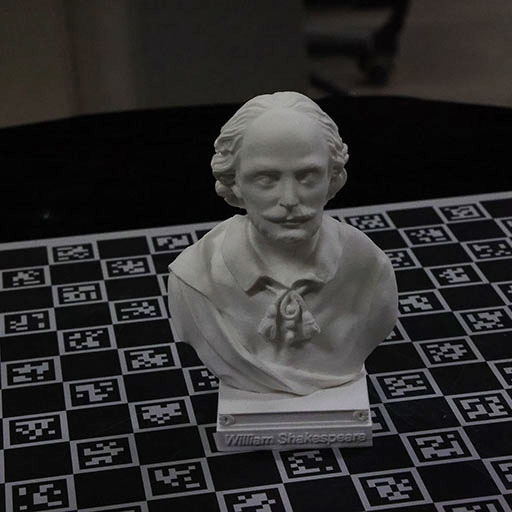} &
    \includegraphics[width=\imgsize\linewidth, trim=60px 2px 60px 4px, clip=true]{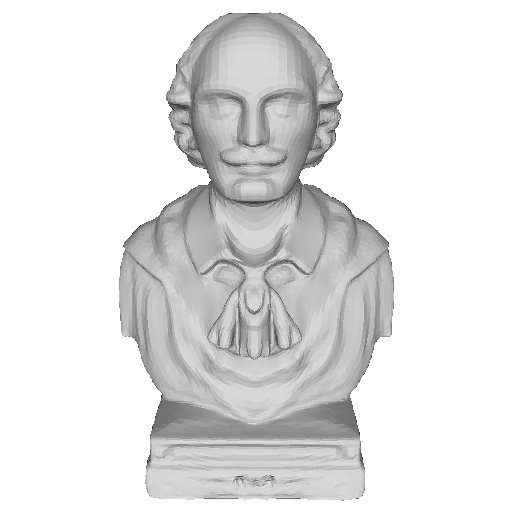} &
    \includegraphics[width=\imgsize\linewidth, trim=60px 2px 60px 4px, clip=true]{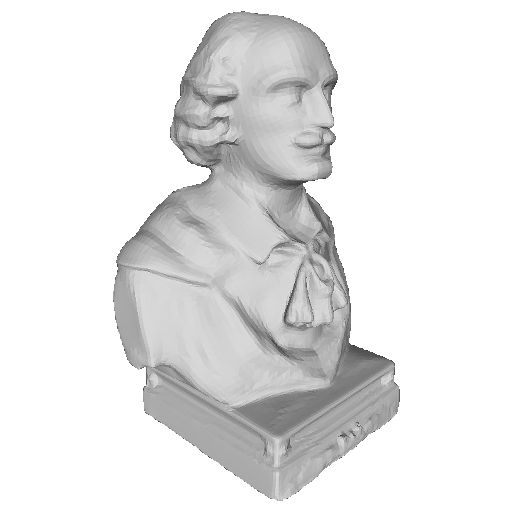}  &
    \includegraphics[width=\imgsize\linewidth]{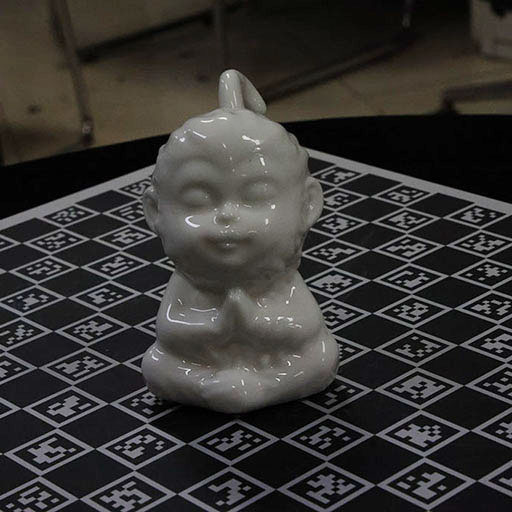} &
    \includegraphics[width=\imgsize\linewidth, trim=60px 2px 60px 4px, clip=true]{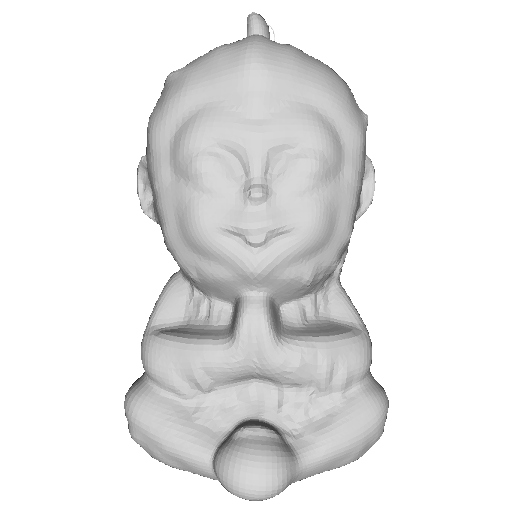} &
    \includegraphics[width=\imgsize\linewidth, trim=60px 2px 60px 4px, clip=true]{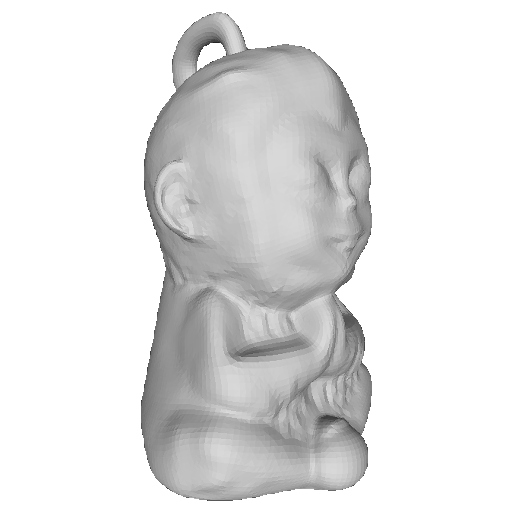}
    \\

    \raisebox{0.0\height}{\rotatebox{90}{\makecell{One-2-3-45++~\cite{liu2024one2345++}}}} & 
    \includegraphics[width=\imgsize\linewidth]{fig/com_aigc/sb_nf.JPG} &
    \includegraphics[width=\imgsize\linewidth, trim=60px 2px 60px 4px, clip=true]{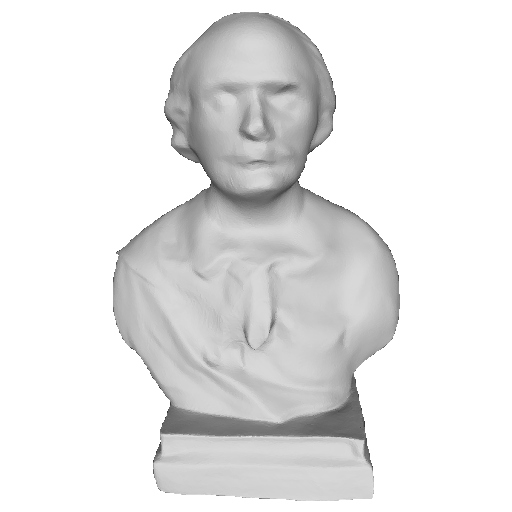} &
    \includegraphics[width=\imgsize\linewidth, trim=60px 2px 60px 4px, clip=true]{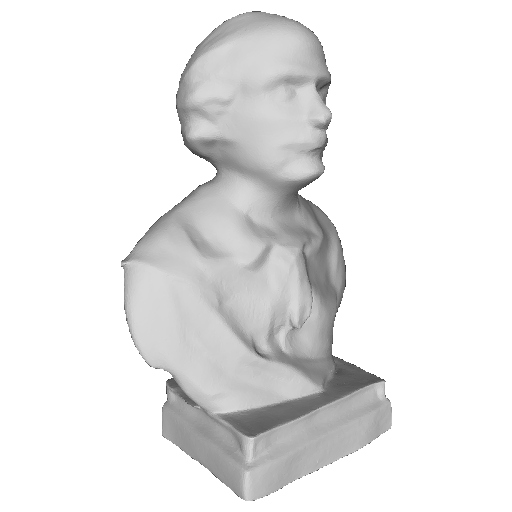}  &
    \includegraphics[width=\imgsize\linewidth]{fig/com_aigc/mb_nf.JPG} &
    \includegraphics[width=\imgsize\linewidth, trim=60px 2px 60px 4px, clip=true]{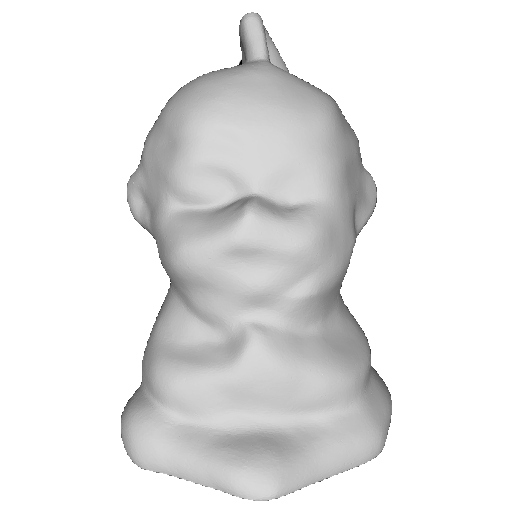} &
    \includegraphics[width=\imgsize\linewidth, trim=60px 2px 60px 4px, clip=true]{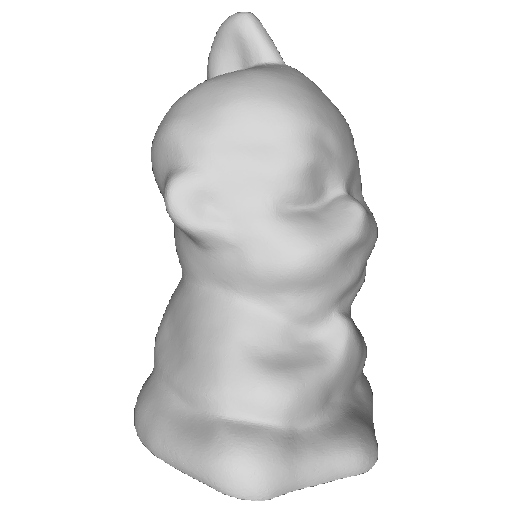}  
    \\

\end{NiceTabular}
\end{adjustbox}
\caption{Mesh generation results of {\sc Shakespeare}~(left side) and {\sc Monkey}~(right side) by image-to-3D methods~\cite{zhang2024clay,liu2024one2345++} and ours. The 3D mesh visualizations are taken as snapshots.} 
\label{fig:com_aigc}
\vspace{-10pt}
\end{figure}

Currently, there are lots of works toward generating 3D shapes from single image~\cite{LRM2024,PF-LRM2024,zhang2024clay,liu2024one2345++}, among which CLAY~\cite{zhang2024clay} and One-2-3-45++~\cite{liu2024one2345++} demonstrate state-of-the-art performances. Therefore, we compare these methods with ours in terms of mesh generation. As shown in \fref{fig:com_aigc}, we show snapshots of reconstructed meshes by image-to-3D methods. Their front view does not align well with the actual appearance of the objects. In our comparison, the key criterion is not the overall plausibility of a completed 3D shape, but whether the visible front-view geometry remains faithful to the observed input. CLAY~\cite{zhang2024clay} and One-2-3-45++~\cite{liu2024one2345++} may generate plausible 3D geometry, but they can also hallucinate or distort visible structures, making their front-view geometry less faithful to the actual observation. Such hallucinated geometry also makes direct alignment-based evaluation less appropriate in our setting. Additionally, their side view is relatively smooth and does not effectively reflect the original details of the surface of the object. On the other hand, we show the mesh integrated by BiNi~\cite{bini2022cao}, whose input is estimated by our method. FlashNormal performs deterministic, pixel-aligned prediction guided by physical shading cues, which is more suitable for recovering visible-surface normals and local details faithfully observed in the input images. Our mesh preserves better surface details and maintains geometric consistency across different views. 

\subsection{Ablation study}

\begin{figure}[!t]
\centering
\begin{adjustbox}{max width=\linewidth}
\addtolength{\tabcolsep}{-0.4em}
\renewcommand{\imgsize}{0.3\textwidth}
\begin{NiceTabular}{cccccc}
 \fontsize{22}{16}\selectfont  {GT normal} & \fontsize{22}{16}\selectfont {coarse normal} & \fontsize{22}{16}\selectfont  {ours} & \fontsize{22}{16}\selectfont  {w/o flash input} & \fontsize{22}{22}\selectfont  {\makecell{w/o curvature-guided \\detail enhancement}} & \fontsize{22}{22}\selectfont  {\makecell{w/o zoomed \\pixels strategy}} \\
    \includegraphics[width=\imgsize, trim=75px 2px 75px 4px, clip=true]{fig/benchmark/pc_gt.jpg} &
    \includegraphics[width=\imgsize, trim=79px 1px 79px 3px, clip=true]{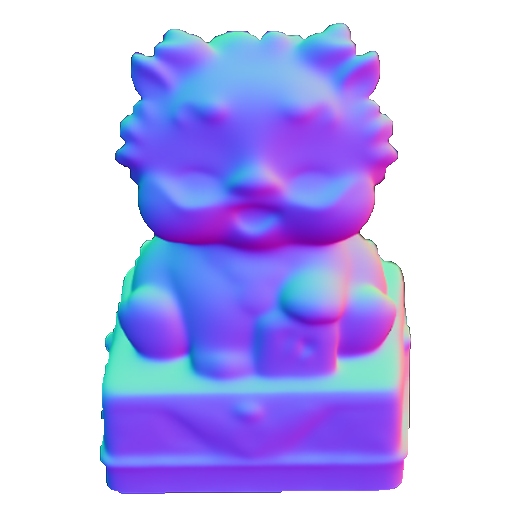} &
    \includegraphics[width=\imgsize, trim=40px 2px 40px 4px, clip=true]{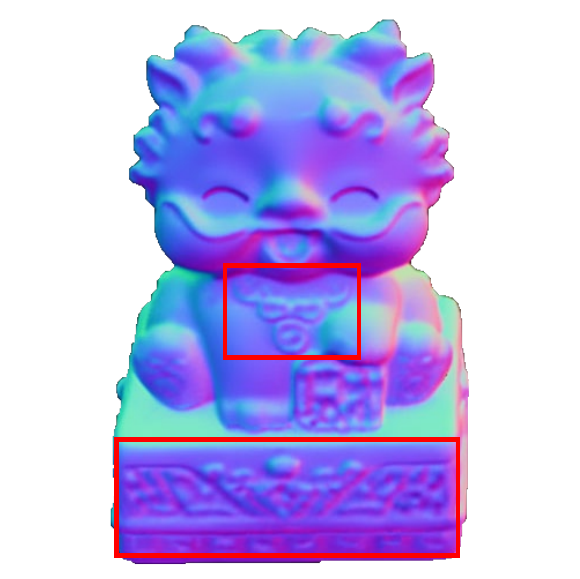} &
    \includegraphics[width=\imgsize, trim=40px 2px 40px 4px, clip=true]{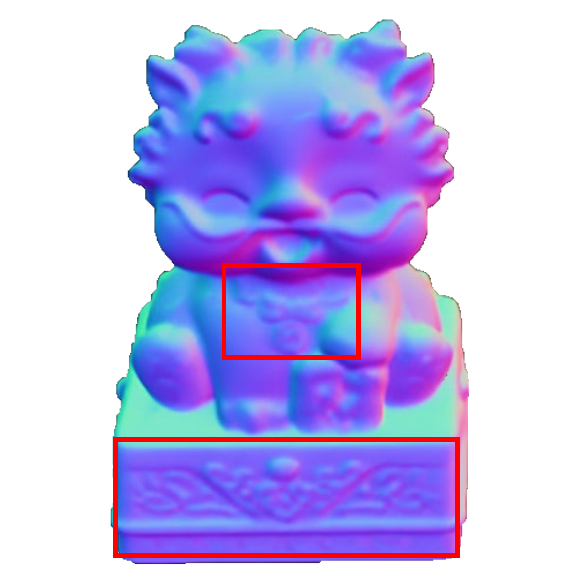} &
    
    \includegraphics[width=\imgsize, trim=40px 2px 40px 4px, clip=true]{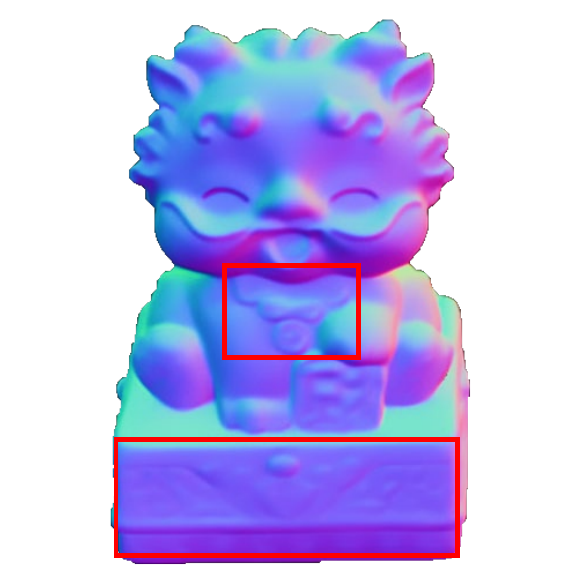} &
    \includegraphics[width=\imgsize, trim=40px 2px 40px 4px, clip=true]{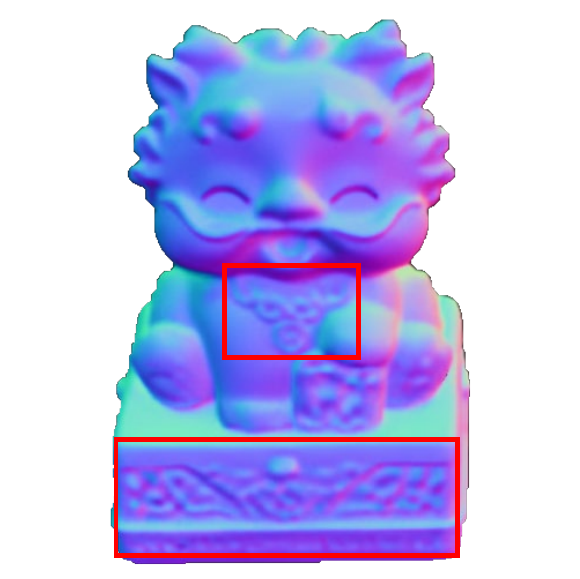} \\
    
    &\fontsize{22}{16}\selectfont MAE=20.31 & \fontsize{22}{16}\selectfont MAE=\textbf{18.95} &\fontsize{22}{16}\selectfont MAE=19.62 &\fontsize{22}{16}\selectfont MAE=19.51  &\fontsize{22}{16}\selectfont MAE=19.34 \\
   
    \includegraphics[width=\imgsize, trim=95px 2px 95px 4px, clip=true]{fig/benchmark/lady_gt.jpg} &
    \includegraphics[width=\imgsize, trim=95px 1px 95px 3px, clip=true]{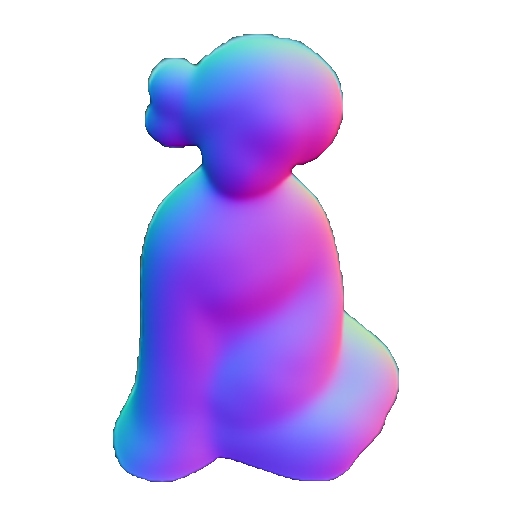} &
    \includegraphics[width=\imgsize, trim=50px 2px 50px 4px, clip=true]{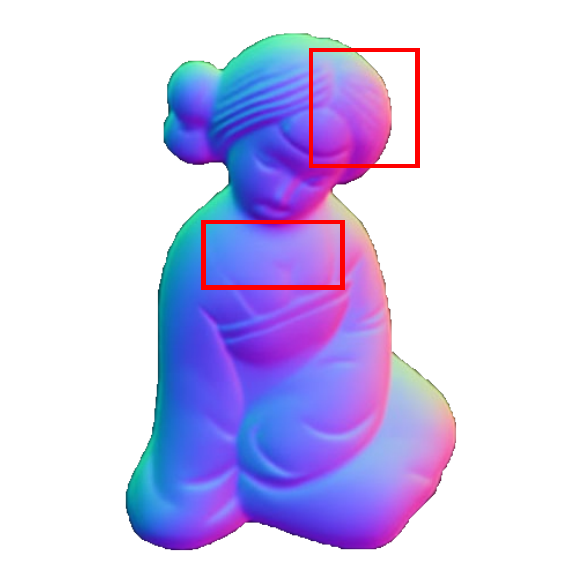} &
    \includegraphics[width=\imgsize, trim=50px 2px 50px 4px, clip=true]{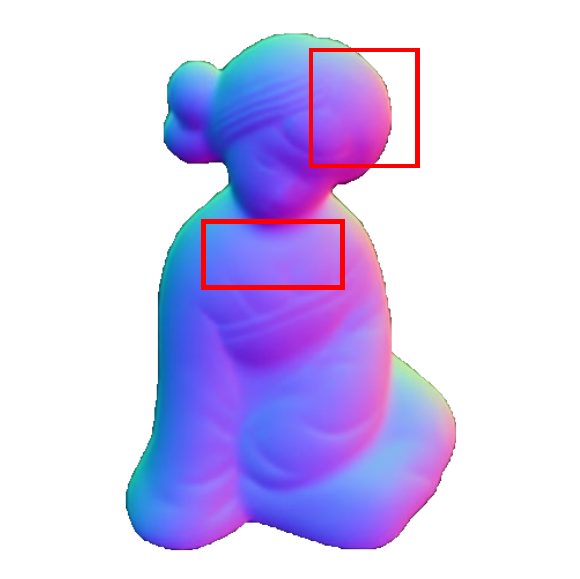} &
    
    \includegraphics[width=\imgsize, trim=50px 2px 50px 4px, clip=true]{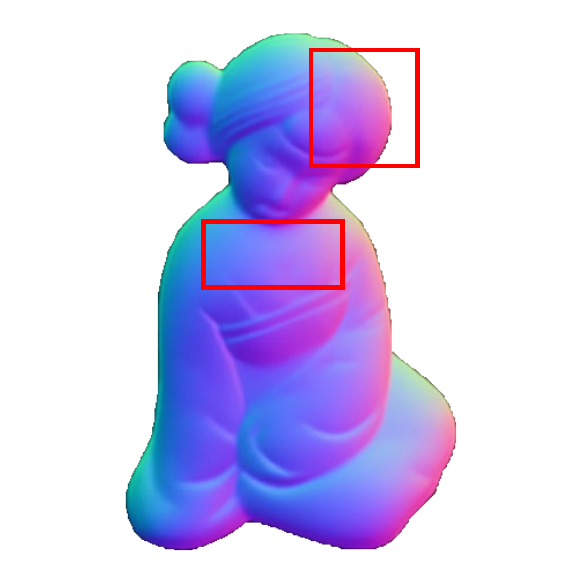} &
    \includegraphics[width=\imgsize, trim=50px 2px 50px 4px, clip=true]{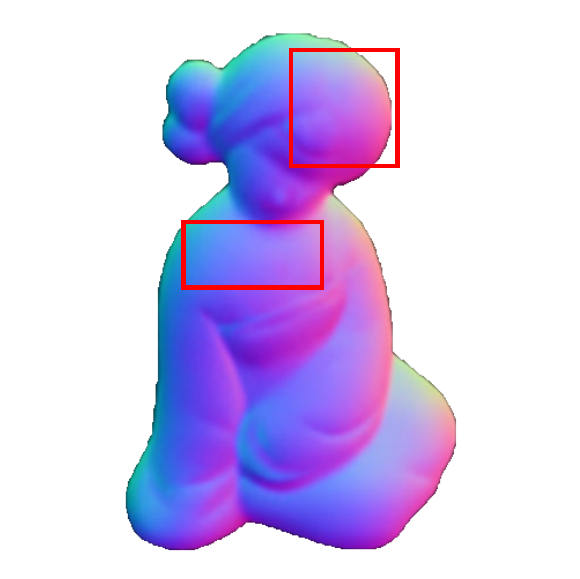} \\
    &\fontsize{22}{16}\selectfont MAE=18.75 &\fontsize{22}{16}\selectfont  MAE=\textbf{17.79} &\fontsize{22}{16}\selectfont MAE=18.56 &\fontsize{22}{16}\selectfont MAE=18.36 &\fontsize{22}{16}\selectfont MAE=18.28 \\

\end{NiceTabular}
\end{adjustbox}
\caption{Ablation on geometric detail recovery from different setups.}

\label{fig:ablation}
\end{figure}

\paragraph{Ablation on geometric detail recovery from different setups} As shown in \fref{fig:ablation}, we demonstrate the effectiveness of flash input, curvature-guided geometry enhancement, and zoomed-pixels strategy in enhancing the accuracy of surface normal estimation with rich detail. When any of these modules is omitted, the MAE value of the estimated surface normals increases, and geometric details are lost, particularly at the neck of {\sc Pinkcat} and the head of {\sc Lady}. \Tref{table:ablation} further reveals that each of the three modules individually reduces the MAE by $3.4\%$, $3.5\%$, and $4.1\%$ on EvalFlash, demonstrating the improved detail estimation ability of our method.

\begin{figure}[!t]
\centering
\begin{adjustbox}{max width=\linewidth}
\addtolength{\tabcolsep}{-0.4em}
\renewcommand{\imgsize}{0.3\textwidth}
\begin{NiceTabular}{ccccc}
     \raisebox{0.08\height}{\rotatebox{90}{\fontsize{20}{16}\selectfont {\sc Drum Poster}}} &
     \includegraphics[width=\imgsize]{fig/qua_com/dd2.jpg} &
     \includegraphics[width=\imgsize]{fig/qua_com/dd1.jpg} &
     \includegraphics[width=\imgsize]{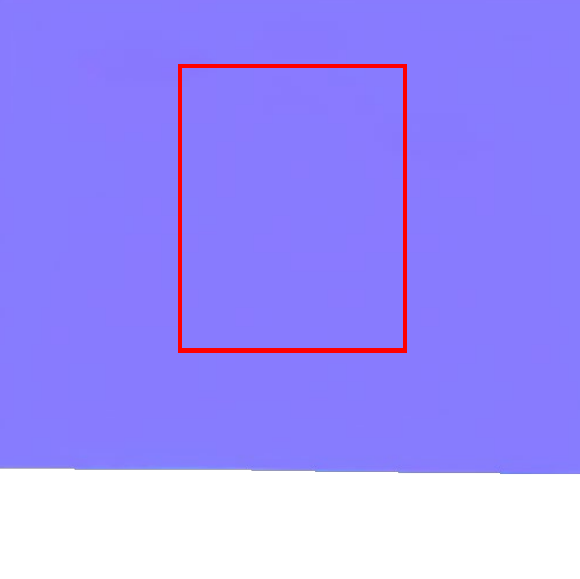} &
     \includegraphics[width=\imgsize]{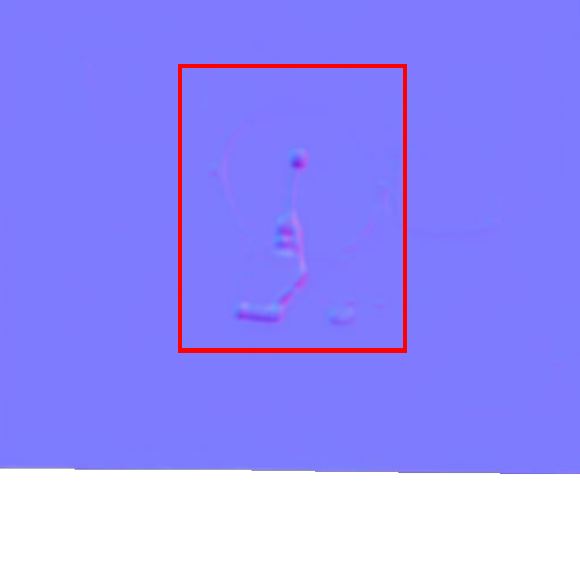}  \\

    \raisebox{0.45\height}{\rotatebox{90}{\fontsize{20}{16}\selectfont {\sc Mirror}}} &
     \includegraphics[width=\imgsize]{fig/qua_com/m2.jpg} &
     \includegraphics[width=\imgsize]{fig/qua_com/m1.jpg} &
     \includegraphics[width=\imgsize]{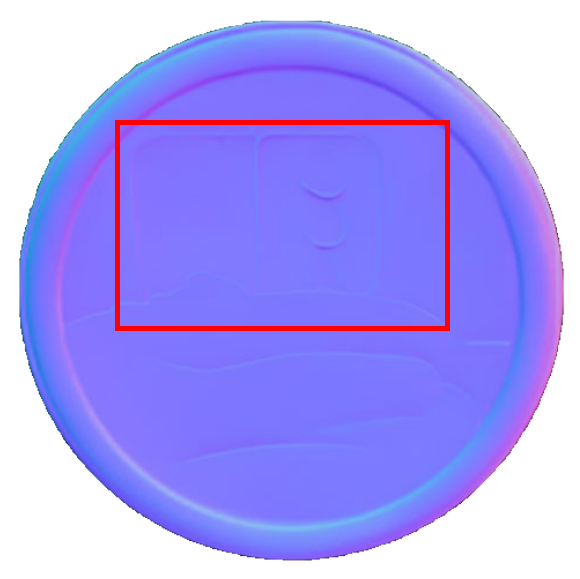} &
     \includegraphics[width=\imgsize]{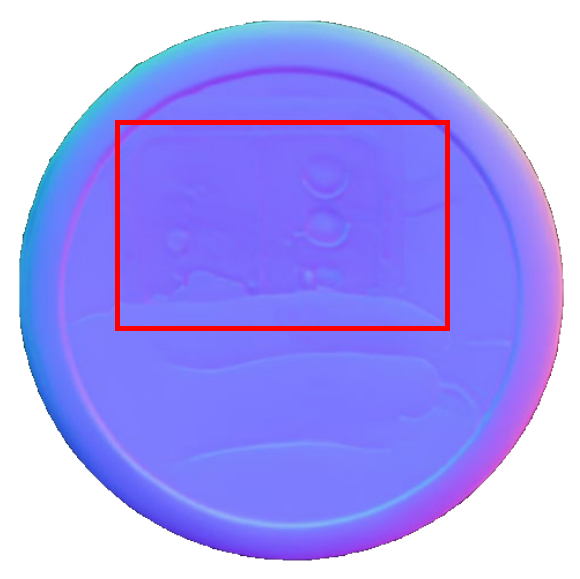}  \\
    
     & \fontsize{22}{16}\selectfont {Flash input} & \fontsize{22}{16}\selectfont {No-flash input} & \fontsize{22}{16}\selectfont {ours} & \fontsize{22}{16}\selectfont {w/o flash input} \\
 \end{NiceTabular}
 \end{adjustbox}
 \caption{Ablation study on flash input.}

 \label{fig:ablation_flash}
 \end{figure}

 \begin{figure}[!t]
 \centering
 \begin{adjustbox}{max width=\linewidth}
 \addtolength{\tabcolsep}{-0.4em}
 \renewcommand{\imgsize}{0.3\textwidth}
 \begin{NiceTabular}{ccc:ccc:ccc}
       \fontsize{22}{16}\selectfont  {coarse normal} && \fontsize{22}{16}\selectfont  {error map}&\fontsize{22}{16}\selectfont  {ours} && \fontsize{22}{16}\selectfont  {error map} & \fontsize{22}{22}\selectfont  {\makecell{w/o coarse shape \\initialization}} && \fontsize{22}{16}\selectfont  {error map}\\
     
     \includegraphics[width=\imgsize, trim=50px 2px 50px 4px, clip=true]{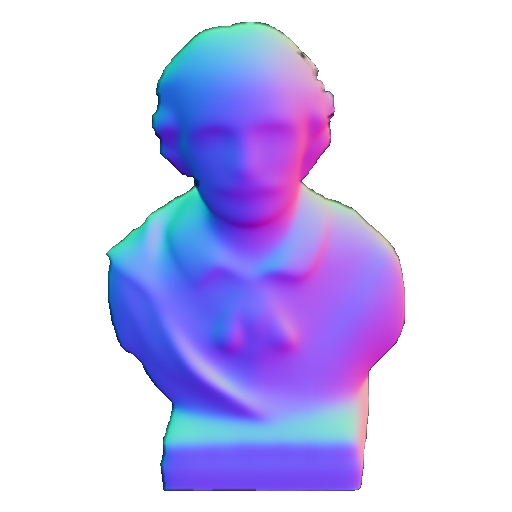} &
     \raisebox{0.45\height}{\rotatebox{90}{\fontsize{20}{16}\selectfont MAE=18.70}} &
     \includegraphics[width=\imgsize, trim=50px 2px 50px 4px, clip=true]{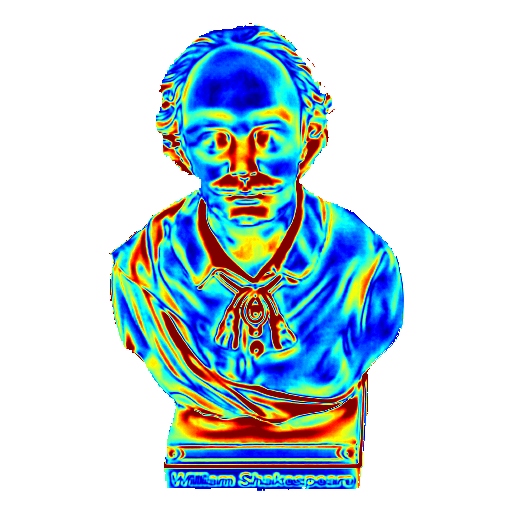} &
     
     \includegraphics[width=\imgsize, trim=50px 2px 50px 4px, clip=true]{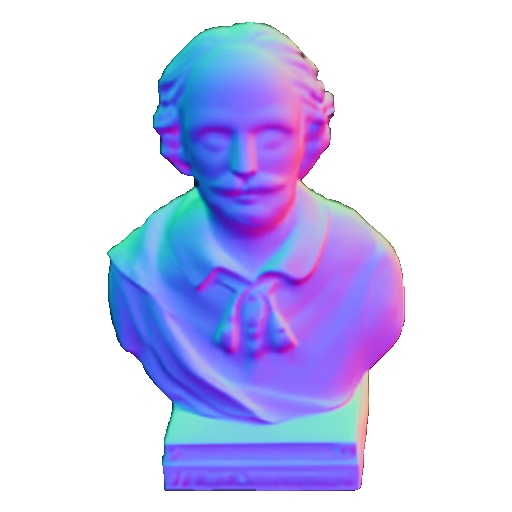} &
     \raisebox{0.45\height}{\rotatebox{90}{\fontsize{20}{16}\selectfont MAE=\textbf{14.98}}} &
     \includegraphics[width=\imgsize, trim=50px 2px 50px 4px, clip=true]{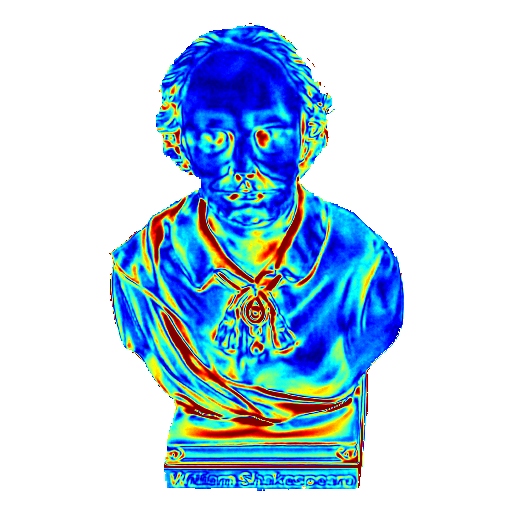} &
     
     \includegraphics[width=\imgsize, trim=50px 2px 50px 4px, clip=true]{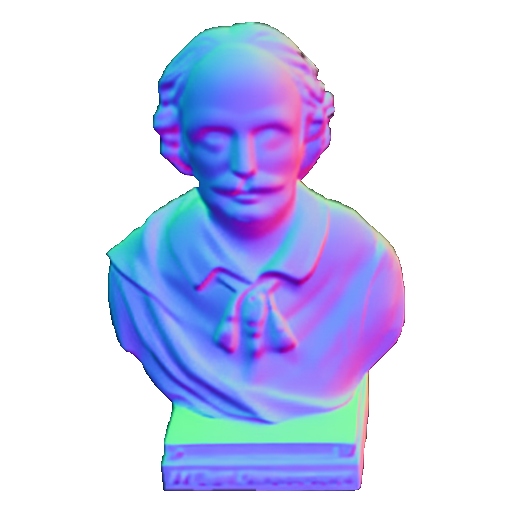} &
     \raisebox{0.45\height}{\rotatebox{90}{\fontsize{20}{16}\selectfont MAE=16.75}} &
     \includegraphics[width=\imgsize, trim=50px 2px 50px 4px, clip=true]{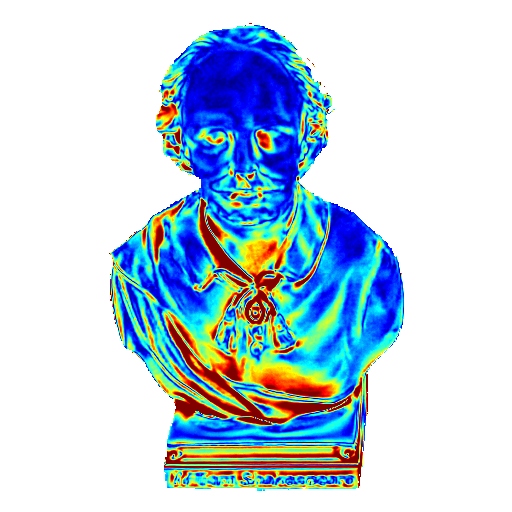}  \\
     
     \includegraphics[width=\imgsize, trim=50px 2px 50px 4px, clip=true]{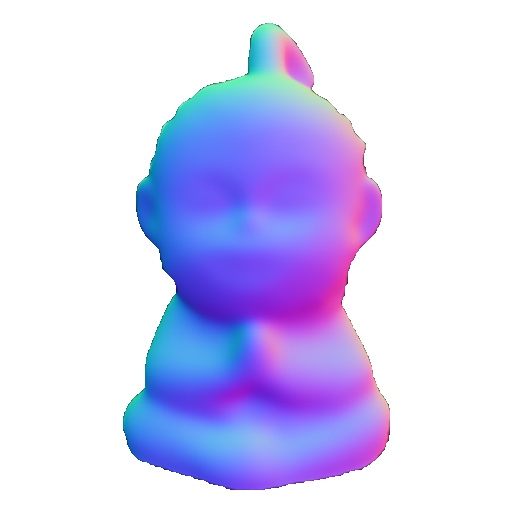} &
     \raisebox{0.45\height}{\rotatebox{90}{\fontsize{20}{16}\selectfont  MAE=15.54}} &
     \includegraphics[width=\imgsize, trim=50px 2px 50px 4px, clip=true]{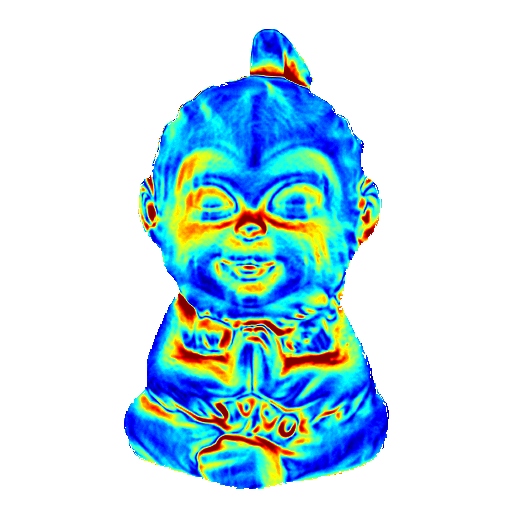} &
     
     \includegraphics[width=\imgsize, trim=50px 2px 50px 4px, clip=true]{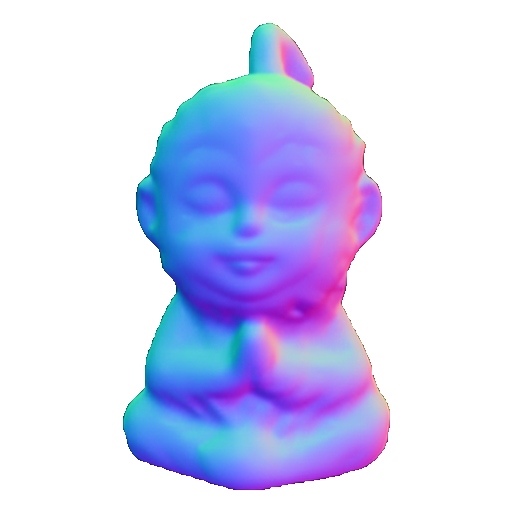} &
     \raisebox{0.45\height}{\rotatebox{90}{\fontsize{20}{16}\selectfont  MAE=\textbf{11.52}}} &
     \includegraphics[width=\imgsize, trim=50px 2px 50px 4px, clip=true]{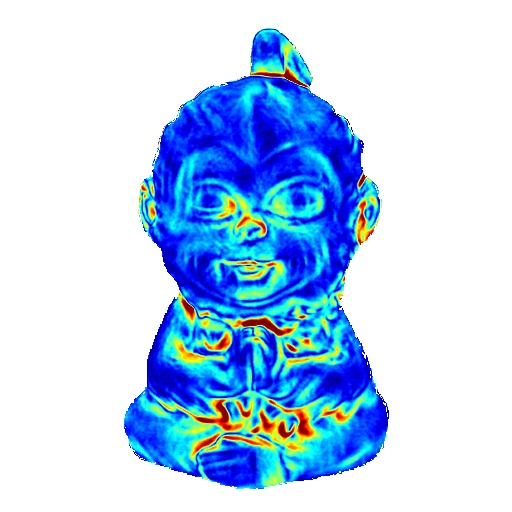} &
     
     \includegraphics[width=\imgsize, trim=50px 2px 50px 4px, clip=true]{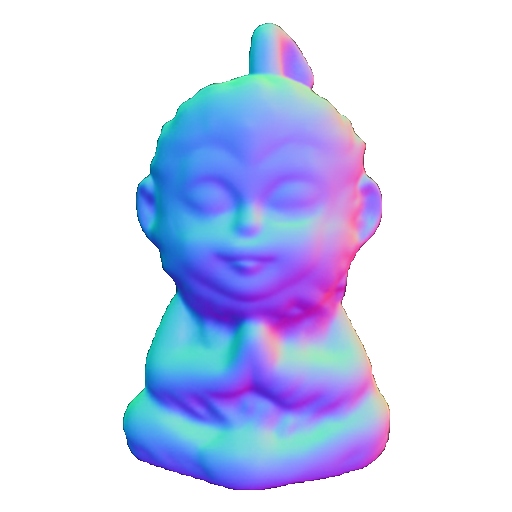} &
     \raisebox{0.45\height}{\rotatebox{90}{\fontsize{20}{16}\selectfont MAE=12.30}} &
     \includegraphics[width=\imgsize, trim=50px 2px 50px 4px, clip=true]{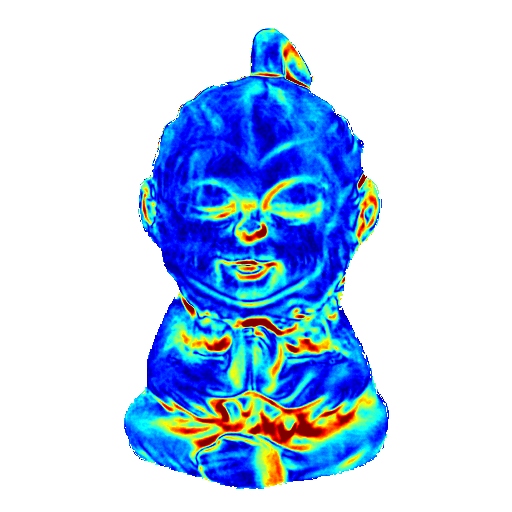}  
     \begin{minipage}{0.020\textwidth} \centering
     	\vspace{-37.5em} \makebox[\textwidth]{ \fontsize{20}{16}\selectfont $\,40^{\circ}$}\\ 
     	\makebox[0.12\textwidth]{}\includegraphics[width=0.9\textwidth]{fig/quan_com/colorbar.jpg} \\ 
     	\makebox[\textwidth]{ \fontsize{20}{16}\selectfont $\,0^{\circ}$}\\
     \end{minipage}
 \end{NiceTabular}
 \end{adjustbox}
 \caption{Ablation study on coarse shape initialization. Though the initial surface normal is lack of detail, it can help enhance the accuracy of estimated surface normal from \ours. }
 \label{fig:ablation_me3d}
\end{figure}

\begin{table}[!htbp]
    \centering
    \caption{Ablation study on \emph{EvalFlash} and \emph{EvalFlash-synth}. We calculated the average values of two testsets.}
    \label{table:ablation}
    \resizebox{0.48\textwidth}{!}{
    \begin{tabular}{ccccccc}
        \toprule 
        &  \multicolumn{3}{c}{\emph{EvalFlash}} & \multicolumn{3}{c}{\emph{EvalFlash-synth}} \\
        Method &  MAE &  RMSE &  MSE &  MAE &  RMSE &  MSE \\
        \midrule
        FlashNormal~(Ours) &\textbf{14.96} &\textbf{0.31} & \textbf{0.03} & \textbf{15.47} & \textbf{0.33} & \textbf{0.04} \\
        \midrule
        w/o flash input &15.48 &0.32 &0.03 &15.88 &0.35 &0.05 \\
        \midrule
        \makecell{w/o curvature-guided \\detail enhancement}  &15.50 &0.33 &0.04 &16.07 &0.35 &0.05 \\
        \midrule
        w/o zoomed pixels strategy &15.60 &0.32 &0.04 &17.96 &0.38 &0.06 \\
        \midrule
        \makecell{w/o global shape \\initialization} &16.33 &0.34 &0.04 &18.23 &0.39 &0.07 \\
        \bottomrule
    \end{tabular}}

\end{table}

\paragraph{Ablation on different input settings}
As shown in \fref{fig:ablation_flash}, flash inputs benefit shape-reflectance ambiguous surfaces, such as the {\sc Drum Poster} and {\sc Mirror}, by using shading variations to differentiate flat surfaces from true 3D structures. Additionally, the last row of \Tref{table:ablation} and \fref{fig:ablation_me3d} demonstrate that providing an easy and fast-to-obtain coarse surface normal initialization accelerates convergence of training process, enhancing the quality of the predicted surface normals when training for the same number~(10) of epochs. These experiments indicate the importance of both flash inputs and coarse shape initialization in the learning process.

\paragraph{Sensitivity to the loss weight $\lambda$}
We additionally evaluated prompt variants by removing ``DSLR'' or ``detailed'' and observed only minor changes, indicating that the prediction is dominated by flash/no-flash visual conditioning rather than prompt wording. The quantitative comparison is reported in Table~\ref{tab:lambda_sensitivity}.

\begin{table}[t]
\centering
\begingroup
\caption{Sensitivity study on the loss weight $\lambda$. We report the average values on EvalFlash and EvalFlash-synth.}
\label{tab:lambda_sensitivity}
\resizebox{0.48\textwidth}{!}{
\begin{tabular}{lcccccc}
\toprule
$\lambda$ & \multicolumn{3}{c}{EvalFlash} & \multicolumn{3}{c}{EvalFlash-synth} \\
\cmidrule(lr){2-4} \cmidrule(lr){5-7}
& MAE & RMSE & MSE & MAE & RMSE & MSE \\
\midrule
10 & 15.46 & 0.34 & 0.11 & 16.12 & 0.35 & 0.05 \\
1000 & 15.23 & 0.33 & 0.05 & 15.98 & 0.34 & 0.05 \\
\addlinespace
\textbf{100 (used in paper)} & \textbf{14.96} & \textbf{0.31} & \textbf{0.03} & \textbf{15.47} & \textbf{0.33} & \textbf{0.04} \\
\bottomrule
\end{tabular}}
\endgroup
\end{table}

\paragraph{Sensitivity to prompt variants}
We additionally evaluated prompt variants by removing ``DSLR'' or ``detailed'' and observed only minor changes, indicating that the prediction is dominated by flash/no-flash visual conditioning rather than prompt wording. The quantitative comparison is reported in Table~\ref{tab:prompt_sensitivity}.
\begin{table}[t]
\centering
\begingroup
\caption{Sensitivity study on prompt variants. We report the average values on EvalFlash and EvalFlash-synth.}
\label{tab:prompt_sensitivity}
\resizebox{0.48\textwidth}{!}{
\begin{tabular}{lcccccc}
\toprule
Prompt variant & \multicolumn{3}{c}{EvalFlash} & \multicolumn{3}{c}{EvalFlash-synth} \\
\cmidrule(lr){2-4} \cmidrule(lr){5-7}
& MAE & RMSE & MSE & MAE & RMSE & MSE \\
\midrule
w/o ``DSLR'' & 15.16 & 0.31 & \textbf{0.03} & 15.73 & 0.34 & \textbf{0.04} \\
w/o ``detailed'' & 15.21 & 0.32 & \textbf{0.03} & 15.66 & 0.34 & 0.05 \\
\addlinespace
\textbf{full prompt (used in paper)} & \textbf{14.96} & \textbf{0.31} & \textbf{0.03} & \textbf{15.47} & \textbf{0.33} & \textbf{0.04} \\
\bottomrule
\end{tabular}}
\endgroup
\end{table}

\subsection{Analyses regarding zoomed-pixels for inputting image to E2E-FT}
\begin{figure}[!t]
\centering
\begin{adjustbox}{max width=\linewidth}
\addtolength{\tabcolsep}{-0.4em}
\renewcommand{\imgsize}{0.3\textwidth}
\begin{NiceTabular}{cccccc}
    &\LARGE {FLash input} &\LARGE {No-flash input} &\LARGE {\ee} & 
    \LARGE {\makecell{ FlashNormal\\(Ours)}} \\
    \raisebox{0.17\height}{\rotatebox{90}{\fontsize{18}{16}\selectfont{\makecell{{\sc Buddha} \\ w/o zoomed \\pixels strategy}}}} &
    \includegraphics[width=\imgsize]{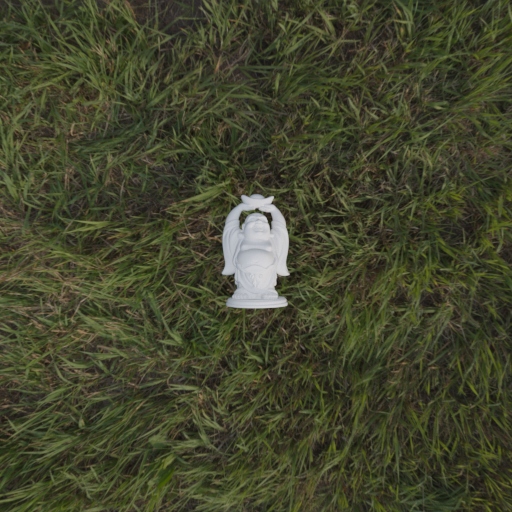} &
    \includegraphics[width=\imgsize]{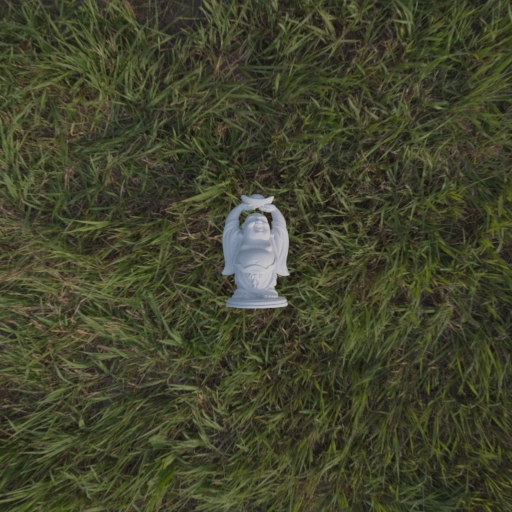} &
    \includegraphics[width=\imgsize]{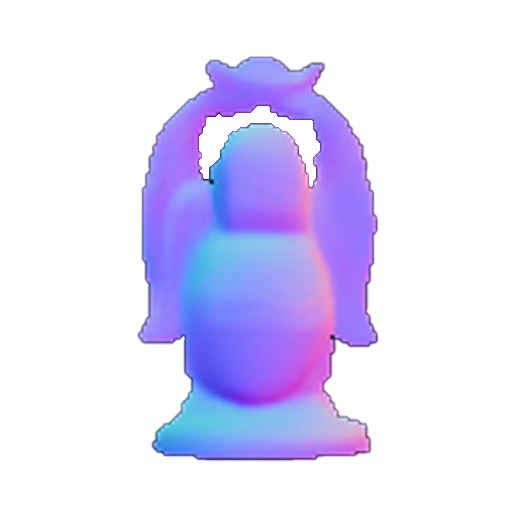} &
    \includegraphics[width=\imgsize]{fig/no.drawio.pdf}
    \\
    & & &\LARGE {MAE=30.01} \\
    \raisebox{0.17\height}{\rotatebox{90}{\fontsize{18}{16}\selectfont{\makecell{{\sc Buddha}\\ w/ zoomed \\pixels strategy}}}} &
    \includegraphics[width=\imgsize]{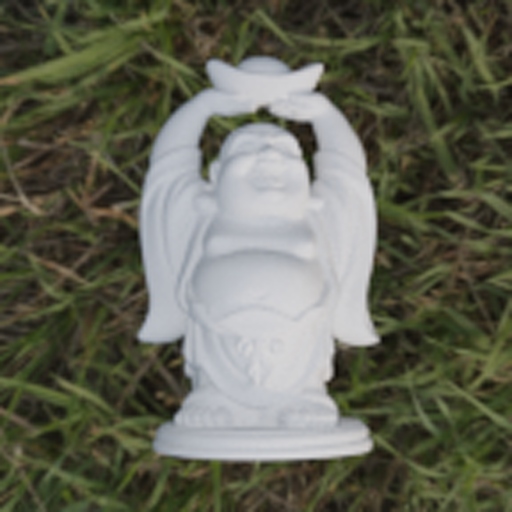} &
    \includegraphics[width=\imgsize]{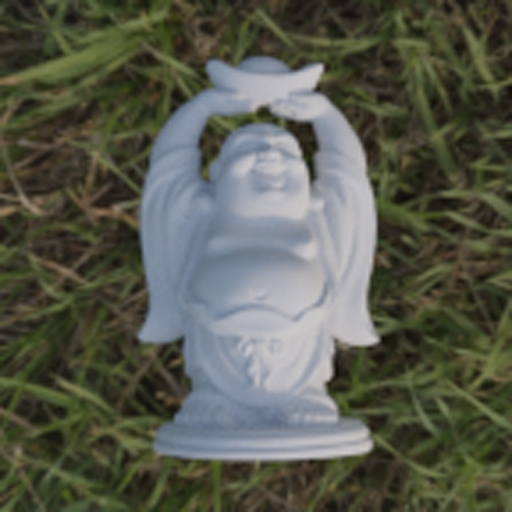} &
    \includegraphics[width=\imgsize]{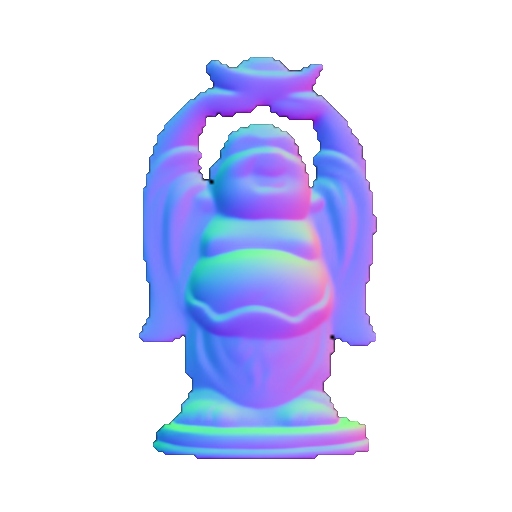} &
    \includegraphics[width=\imgsize]{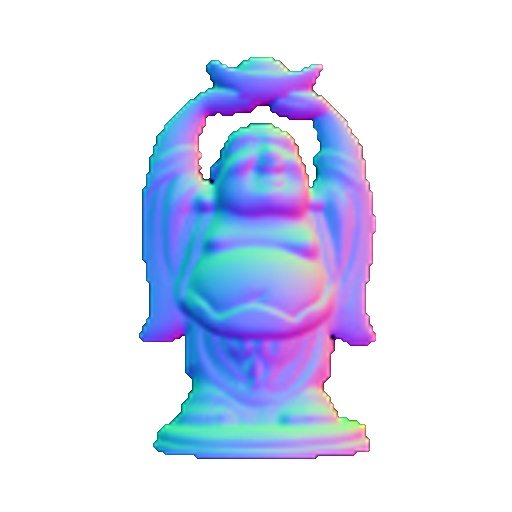} \\
    & & &\LARGE {MAE=18.90} & 
    \LARGE {MAE=\textbf{16.16}} \\
    \raisebox{0.17\height}{\rotatebox{90}{\fontsize{18}{16}\selectfont{\makecell{{\sc Pineapple} \\ w/o zoomed \\pixels strategy}}}} 
    & \includegraphics[width=\imgsize]{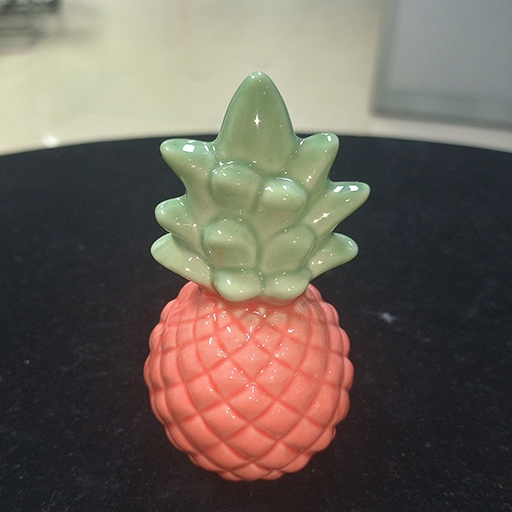} &
    \includegraphics[width=\imgsize]{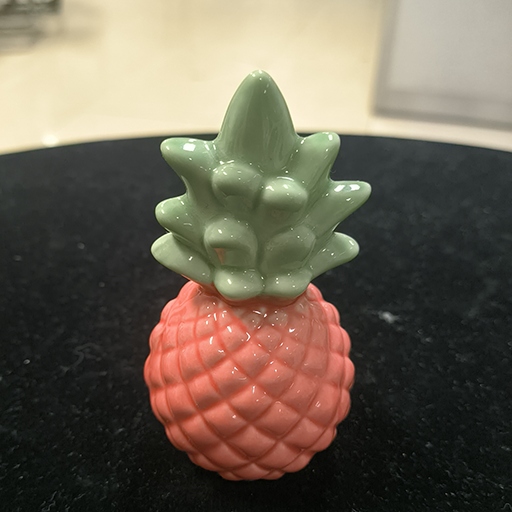} &
    \includegraphics[width=\imgsize]{fig/quan_com/pine_e2e_crop.jpg} &
    \includegraphics[width=\imgsize]{fig/no.drawio.pdf}
    \\
    & & &\LARGE {MAE=14.45} \\
    \raisebox{0.17\height}{\rotatebox{90}{\fontsize{18}{16}\selectfont{\makecell{{\sc Pineapple}\\ w/ zoomed \\pixels strategy}}}} 
    & \includegraphics[width=\imgsize]{fig/benchmark/pine12.jpg} &
    \includegraphics[width=\imgsize]{fig/benchmark/pine11.jpg} &
    \includegraphics[width=\imgsize]{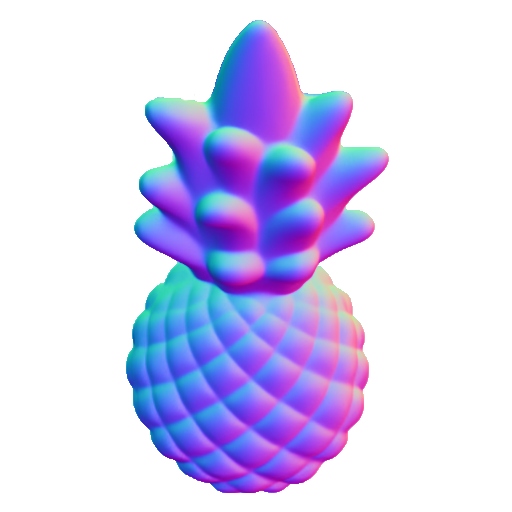} &
    \includegraphics[width=\imgsize]{fig/quan_com/pine_me_crop.jpg}\\
    & & &\LARGE {MAE=13.64} & 
    \LARGE {MAE=\textbf{11.13}} \\
\end{NiceTabular}
\end{adjustbox}
\caption{Comparison between \ee and ours about zoomed pixels strategy. Valid regions are resized to the same resolution.}
\label{fig:com_sn}
\vspace{-10pt}
\end{figure}

While zoomed pixels is a simple yet effective method for fitting variously captured images to networks, in the main paper, we stick to the default settings of our peer methods without applying this method. Therefore, to more comprehensively demonstrate the superiority of our proposal, we here show that even zoomed pixels~(adding bounding box) for inputting image of \ee, its performance is still inferior to our method. As shown in \fref{fig:com_sn}, the two columns on the left are the input RGB images for our proposal, and we input the no-flash image with and without a bounding box into \ee, respectively. It can be observed that the quality of estimated surface normals is effectively improved after adding the bounding box. However, its MAE value is still higher than ours, both in synthetic data {\sc Buddha} and real data {\sc Pineapple}. We attribute this phenomenon to the fact that \ee has not zoomed pixels during training, making the algorithm less sensitive to the presence of bounding boxes. 

\subsection{Robustness analyses}
\paragraph{Robustness analyses \wrt different camera setups}
\begin{figure}[!t]
\centering
\begin{adjustbox}{max width=\linewidth}
\addtolength{\tabcolsep}{-0.4em}
\renewcommand{\imgsize}{0.18}
\begin{NiceTabular}{cccccc}
    & {\Large $f$=24mm} & 
    {\Large $f$=85mm} & 
    {\Large $z$=10m} & 
    {\Large $z$=3m} 
     \\

    \raisebox{0.18\height}{\rotatebox{90}{\Large \makecell{Flash Image}}} &
    \includegraphics[width=\imgsize\textwidth]{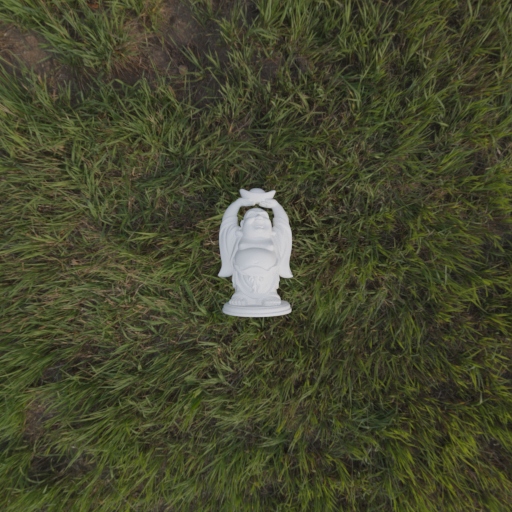} &
    \includegraphics[width=\imgsize\textwidth]{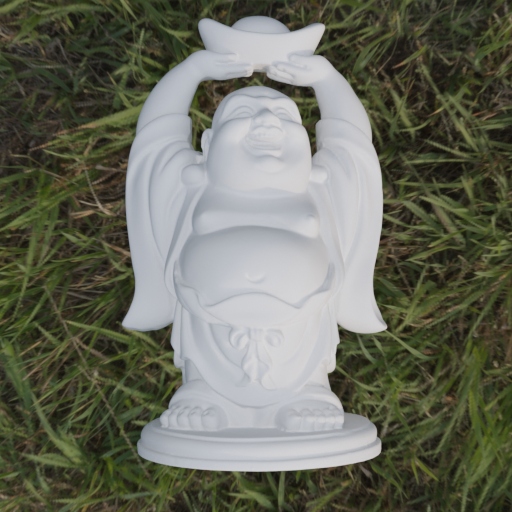} &
    \includegraphics[width=\imgsize\textwidth]{fig/robust_ana/b10_flash.jpg} &
    \includegraphics[width=\imgsize\textwidth]{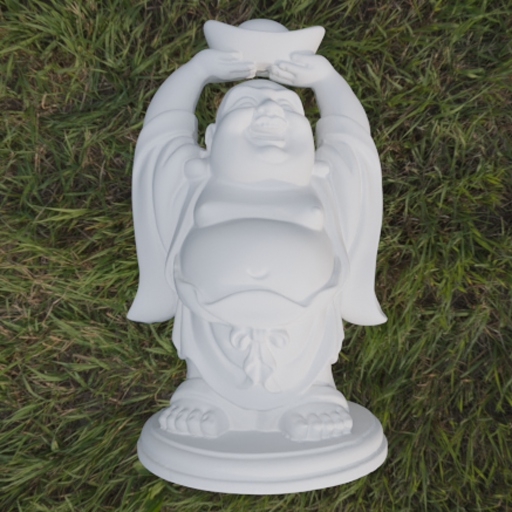} &
    \\
    \raisebox{0.01\height}{\rotatebox{90}{\Large \makecell{No-flash Image}}} &
    \includegraphics[width=\imgsize\textwidth]{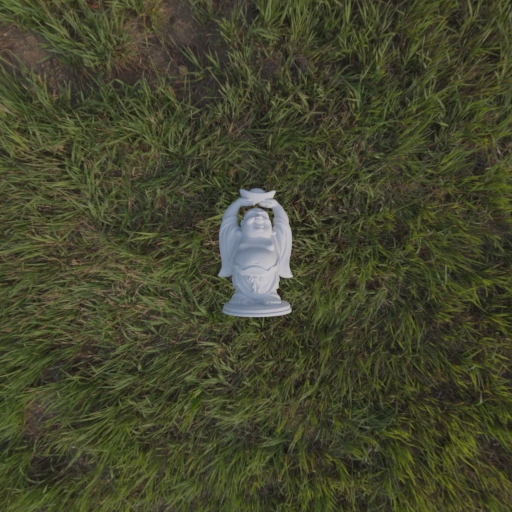} &
    \includegraphics[width=\imgsize\textwidth]{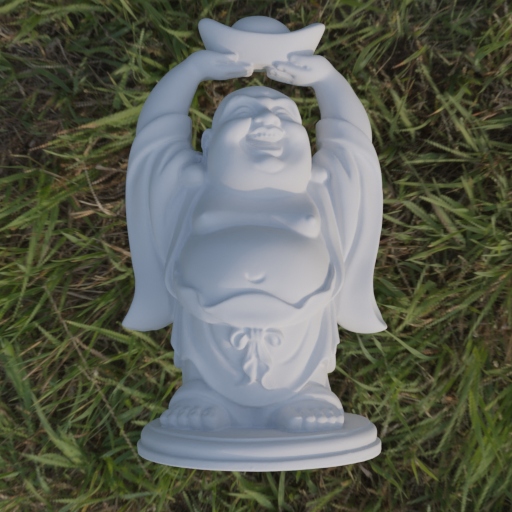} &
    \includegraphics[width=\imgsize\textwidth]{fig/robust_ana/b10.jpg} &
    \includegraphics[width=\imgsize\textwidth]{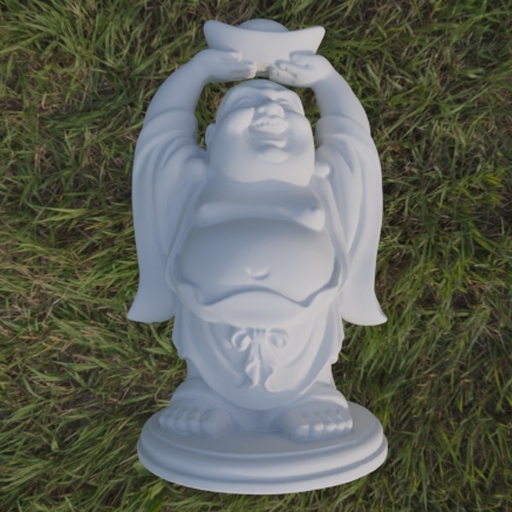} &
    \\

    \raisebox{0.20\height}{\rotatebox{90}{\Large \makecell{FlashNormal (Ours)}}} &
    \includegraphics[width=\imgsize\textwidth, trim=210px 181px 210px 182px, clip=true]{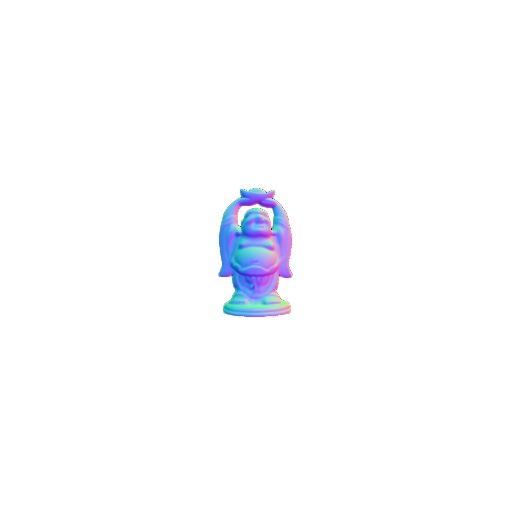} &
    \includegraphics[width=\imgsize\textwidth, trim=100px 2px 100px 4px, clip=true]{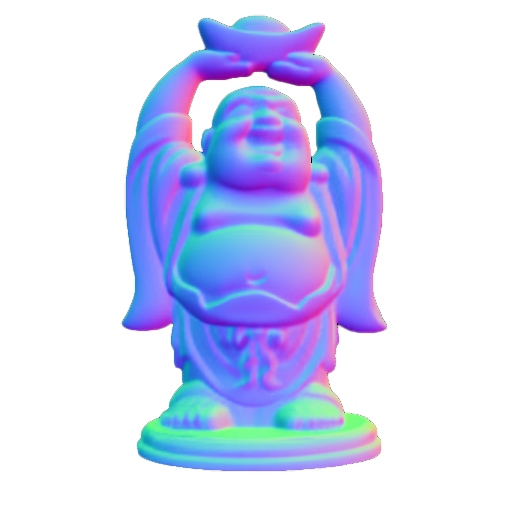} &
    \includegraphics[width=\imgsize\textwidth, trim=218px 197px 218px 192px, clip=true]{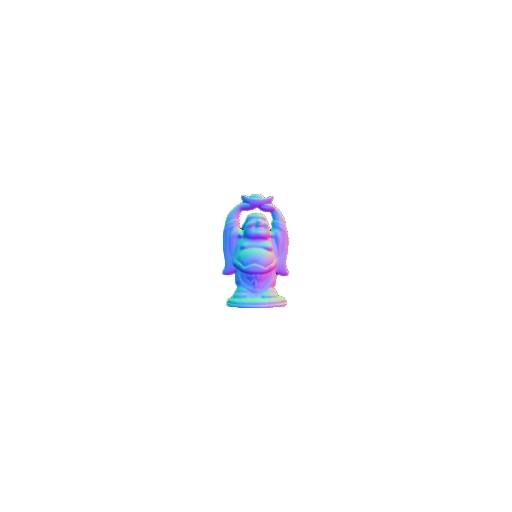} &
    \includegraphics[width=\imgsize\textwidth, trim=100px 2px 100px 4px, clip=true]{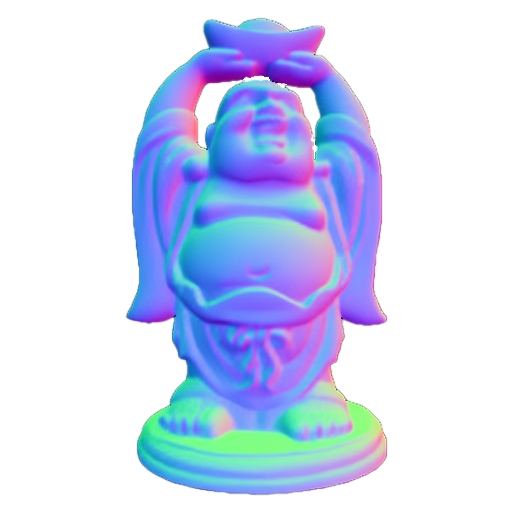} &
    \\
    \raisebox{0.70\height}{\rotatebox{90}{\Large \makecell{\ee}}} &
    \includegraphics[width=\imgsize\textwidth, trim=210px 181px 210px 182px, clip=true]{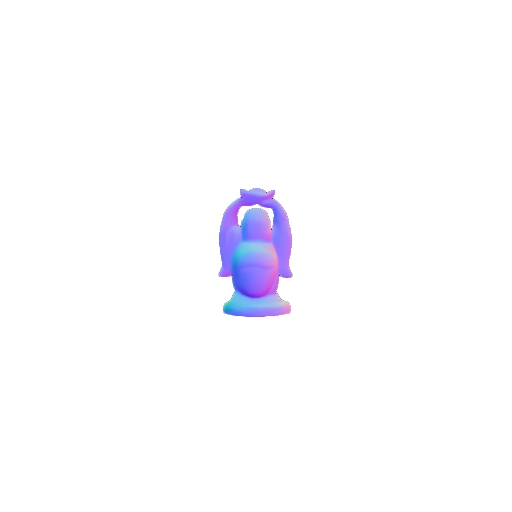} &
    \includegraphics[width=\imgsize\textwidth, trim=100px 2px 100px 4px, clip=true]{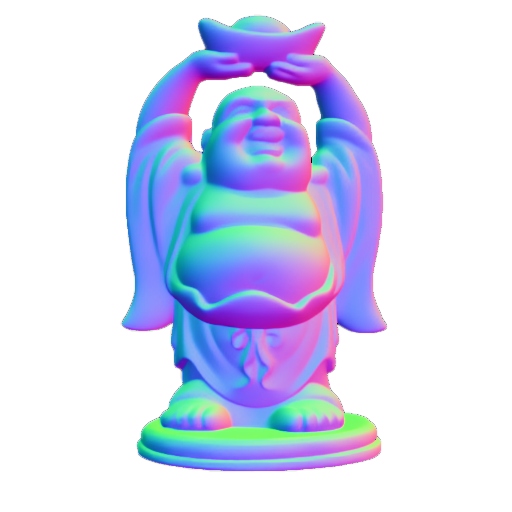} &
    \includegraphics[width=\imgsize\textwidth, trim=218px 197px 218px 192px, clip=true]{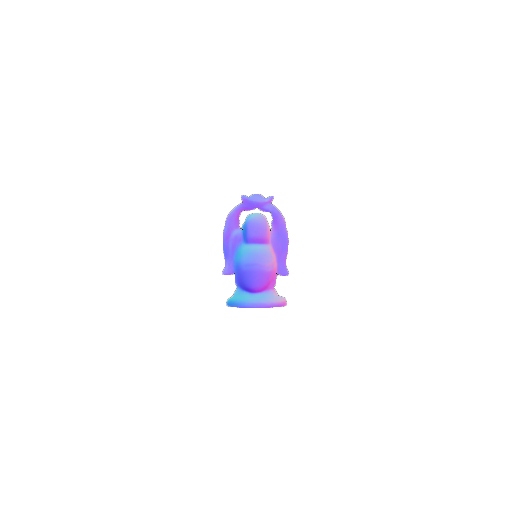} &
    \includegraphics[width=\imgsize\textwidth, trim=100px 2px 100px 4px, clip=true]{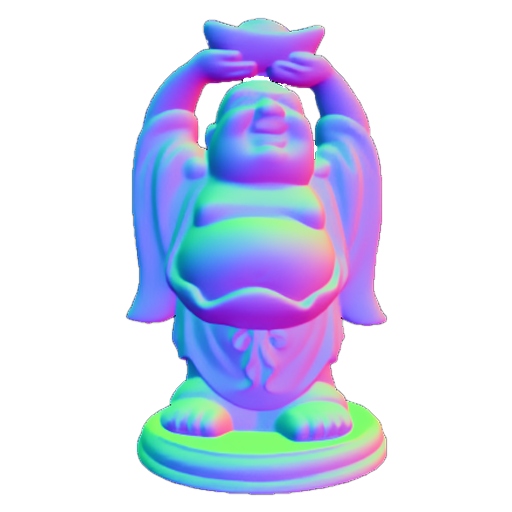} &
  \end{NiceTabular}
\end{adjustbox}
    {\includegraphics[width=0.48\linewidth]{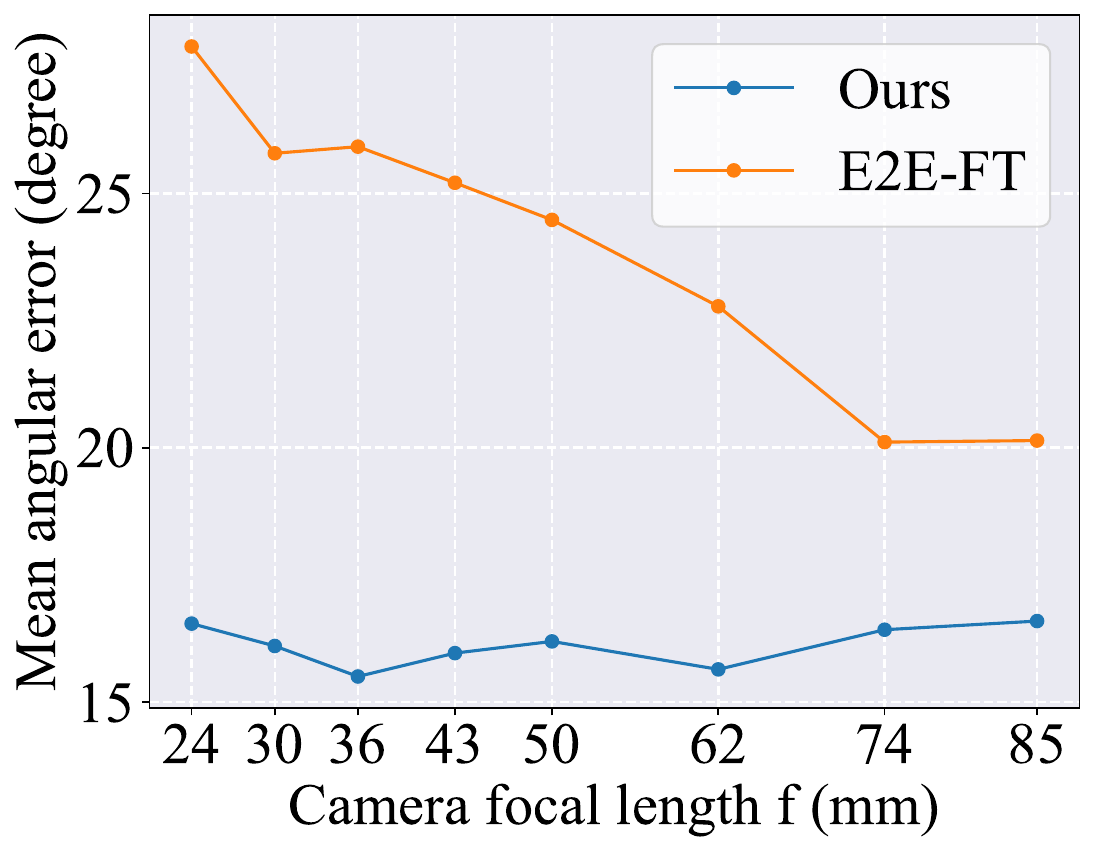}}
  \hfill{\includegraphics[width=0.48\linewidth]{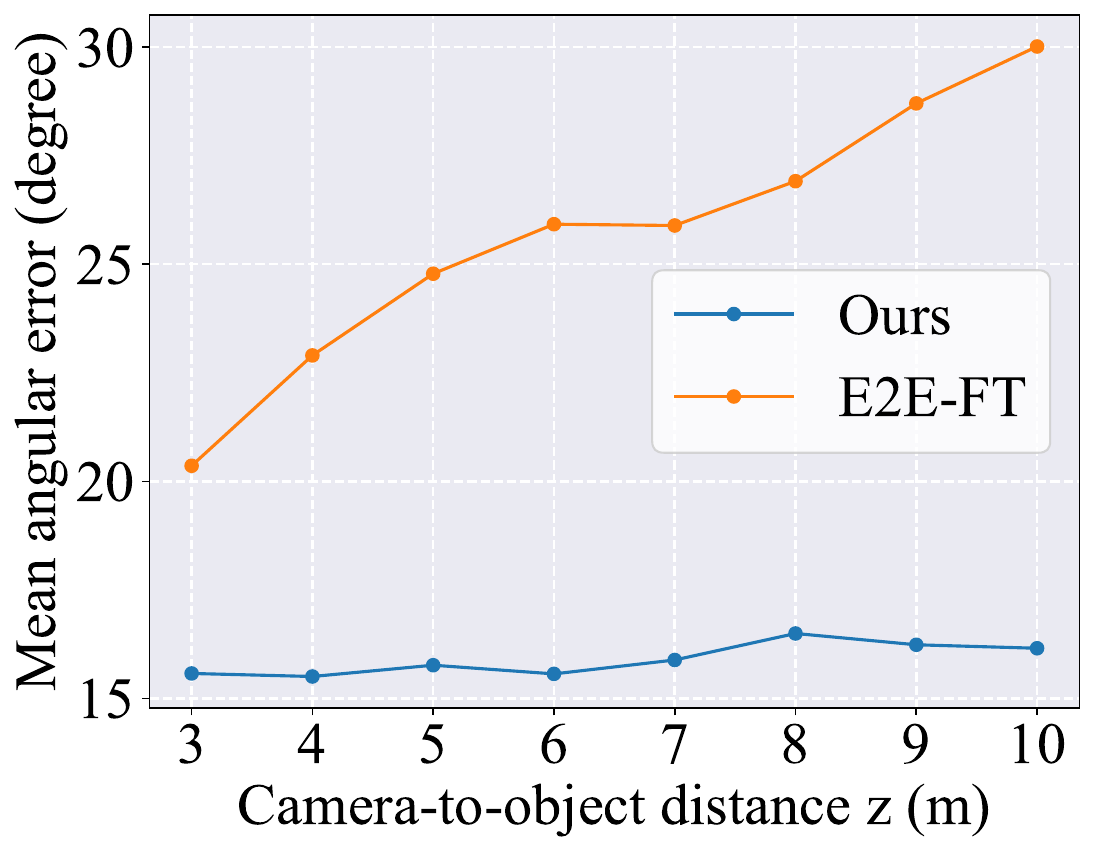}}
\caption{Analysis on the robustness against varying focal length $f$ and camera-to-object distance $z$. The top and middle four rows show the image observations and estimated surface normal maps. Valid regions are resized to the same resolution. The bottom row summarizes the normal estimation errors measured by MAE under varying $f$ and $z$.}
\label{fig:robust_ana}
\vspace{-10pt}
\end{figure}

\begin{figure}[!hbtp]
	\centering
	\includegraphics[width=\linewidth]{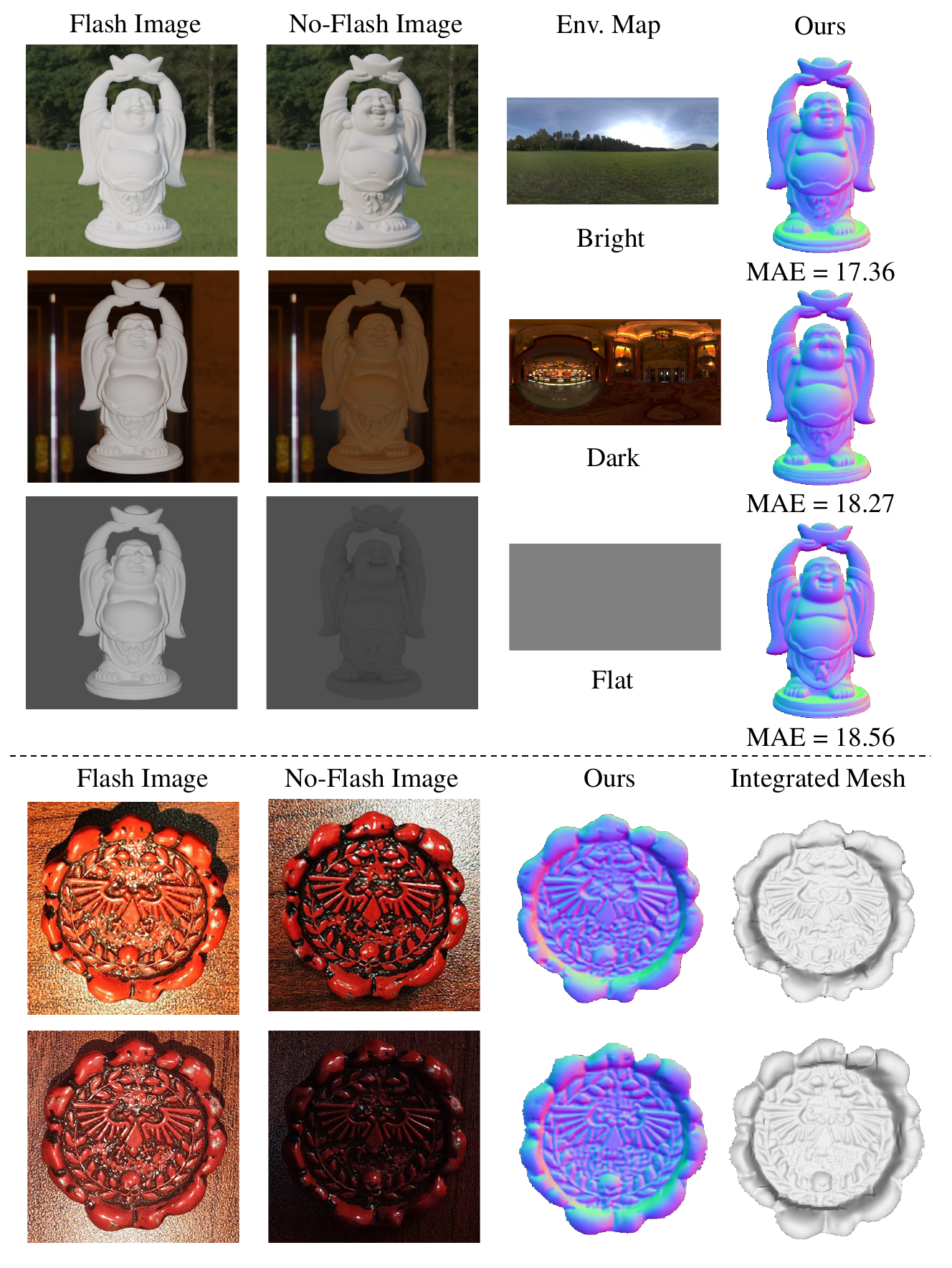}
	\caption{Analysis on the robustness \wrt varying illuminations \& color temperatures. The top three rows show the synthetic experimental results and bottom two rows show the real-world experimental results.}
	\label{fig:robust_light}
\end{figure}

In real-world image acquisition, the camera focal length $f$ and camera-to-object distance $z$ could vary dramatically. To assess the robustness of \ours under these conditions, we conduct tests on images rendered with fixed $z$ but varying $f$, and fixed $f$ but varying $z$. As shown in \fref{fig:robust_ana}, \ee presents blurred predictions when the valid surface area is small, resulting in higher MAE errors. By contrast, our zoomed pixels strategy maintains informative inputs across different values of $f$ and $z$, enabling consistently high-quality surface normal estimations.

\paragraph{Robustness analyses \wrt varying illuminations \& color temperatures}
In real-world scenarios, illumination conditions can vary dramatically. We here show the robustness analyses \wrt varying illuminations \& color temperatures. As shown in \fref{fig:robust_light}, synthetic (top) and real-world (bottom) experimental results show that our proposal can consistently estimate reliable surface normals under ambient light with varying intensities and color temperatures, even under the case of dark and flat illuminations. These results demonstrate the robustness and applicability of our method. 

\begin{figure}[!t]
\centering
\begingroup
\setlength{\tabcolsep}{2pt}
\renewcommand{\arraystretch}{1.02}
\begin{tabular}{cccc}
\makecell{Flash\\input} &
\makecell{No-flash\\input} &
\makecell{Mask\\predicted by\\SAM 2~\cite{ravi2024sam2}} &
\makecell{Recovered\\surface normal} \\
\includegraphics[width=0.23\linewidth]{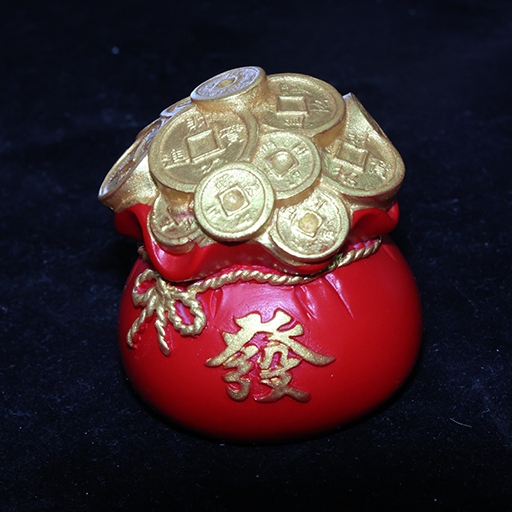} &
\includegraphics[width=0.23\linewidth]{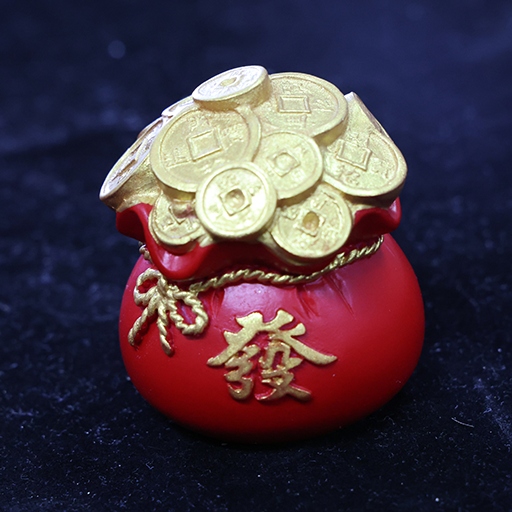} &
\includegraphics[width=0.23\linewidth]{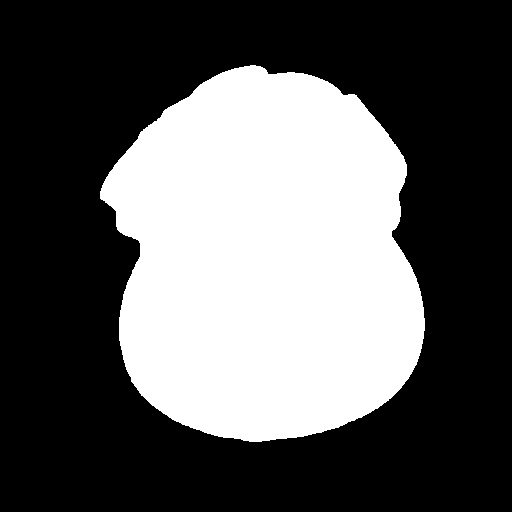} &
\includegraphics[width=0.23\linewidth]{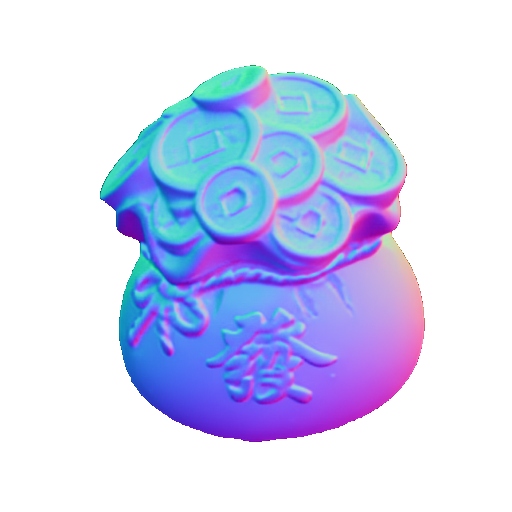} \\
\includegraphics[width=0.23\linewidth]{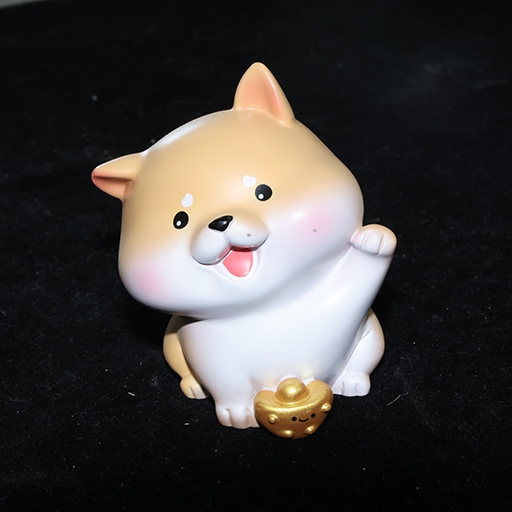} &
\includegraphics[width=0.23\linewidth]{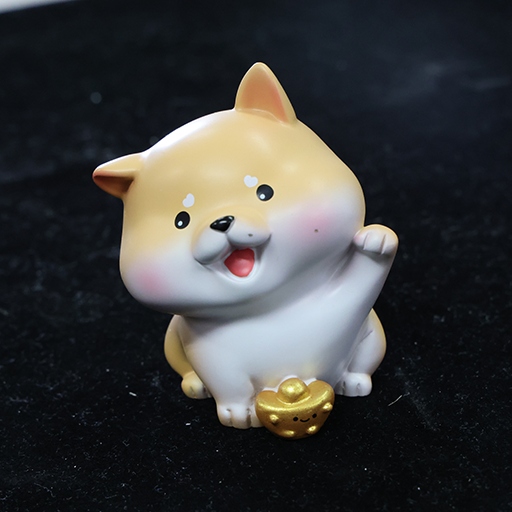} &
\includegraphics[width=0.23\linewidth]{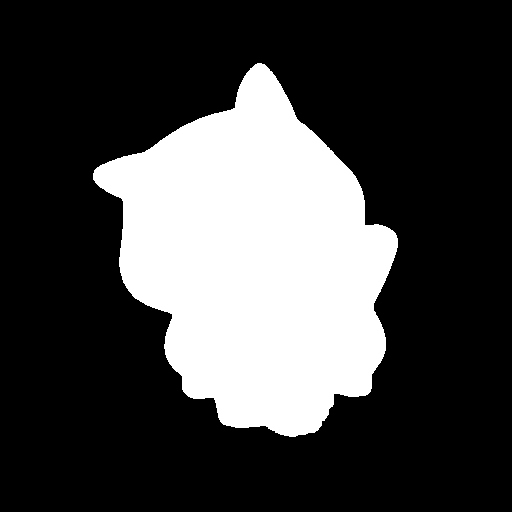} &
\includegraphics[width=0.23\linewidth]{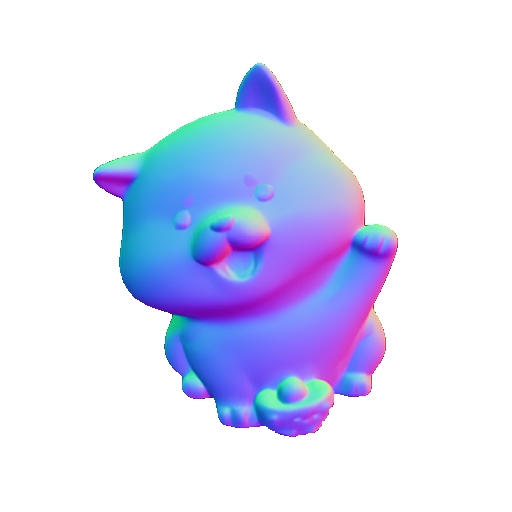} \\
\includegraphics[width=0.23\linewidth]{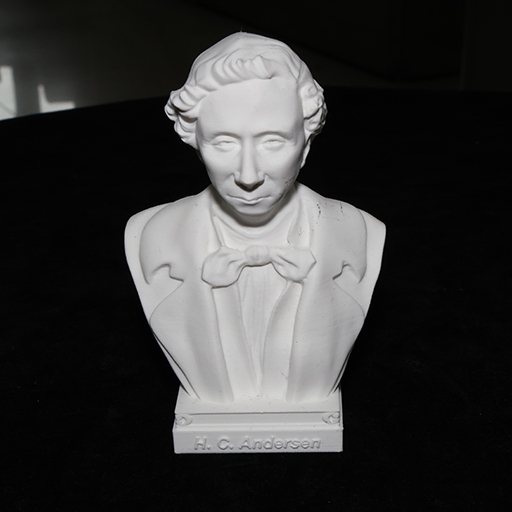} &
\includegraphics[width=0.23\linewidth]{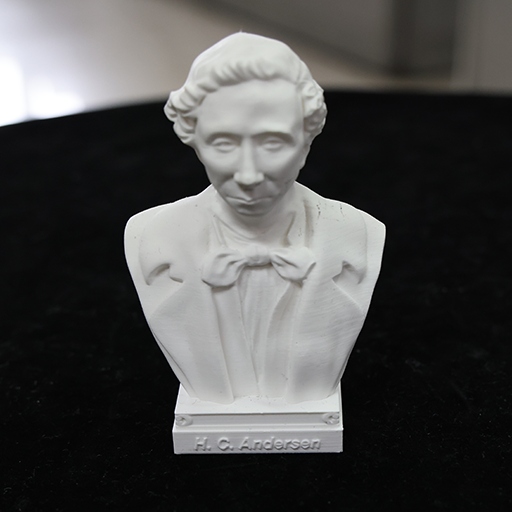} &
\includegraphics[width=0.23\linewidth]{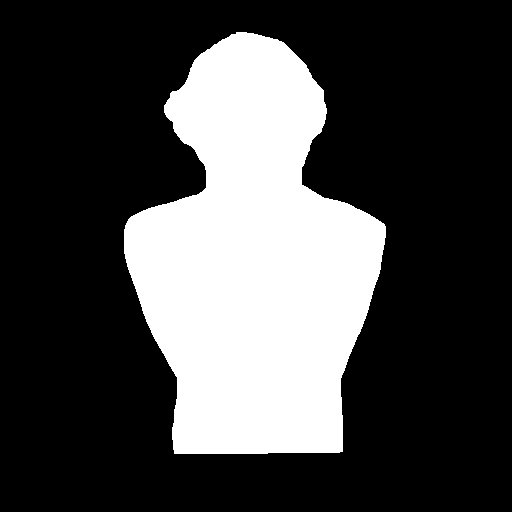} &
\includegraphics[width=0.23\linewidth]{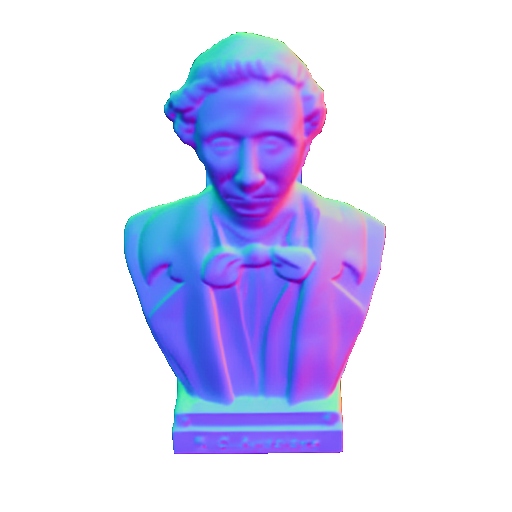} \\
\includegraphics[width=0.23\linewidth]{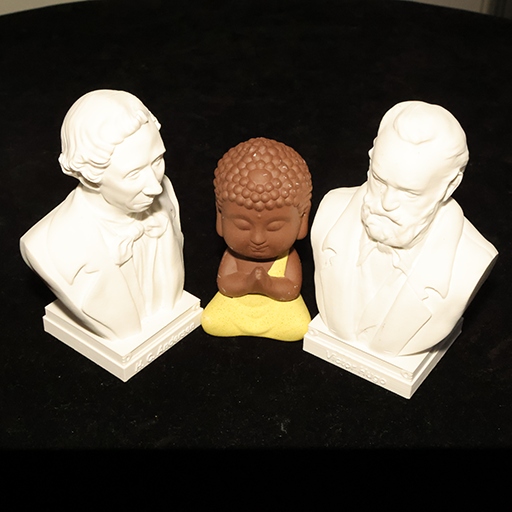} &
\includegraphics[width=0.23\linewidth]{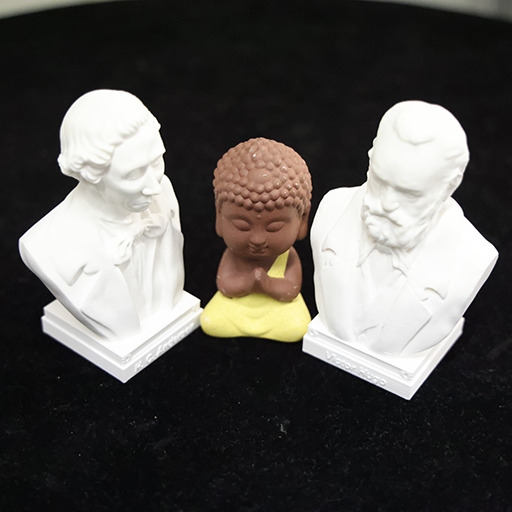} &
\includegraphics[width=0.23\linewidth]{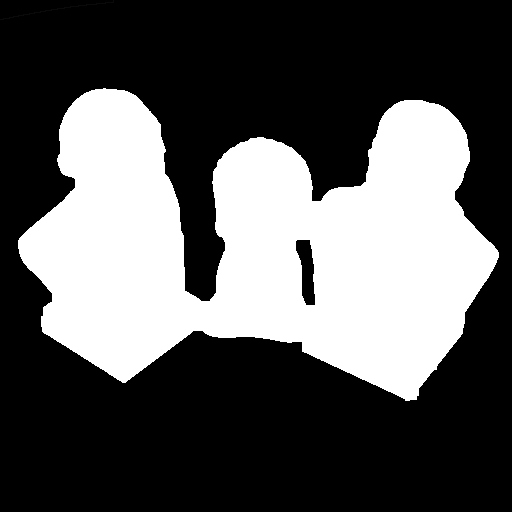} &
\includegraphics[width=0.23\linewidth]{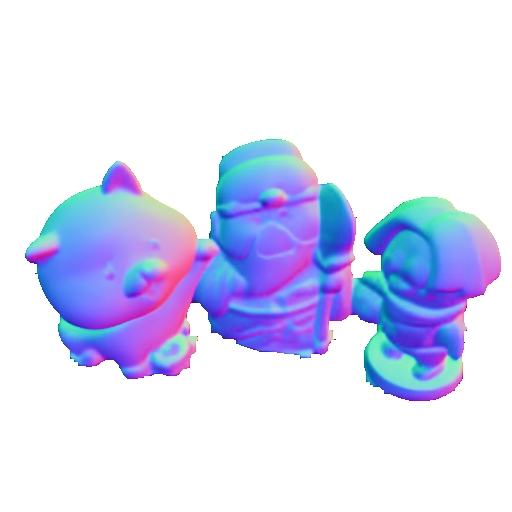} \\
\includegraphics[width=0.23\linewidth]{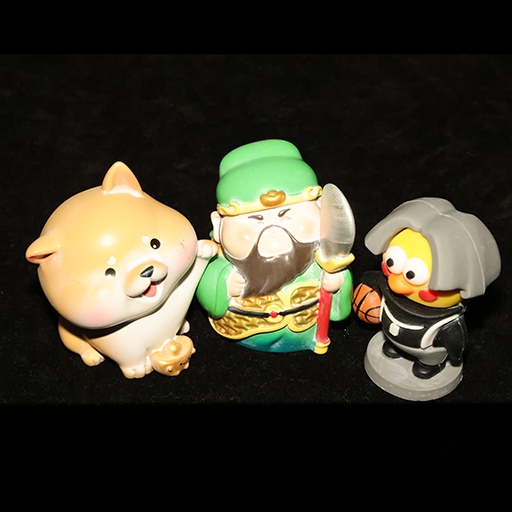} &
\includegraphics[width=0.23\linewidth]{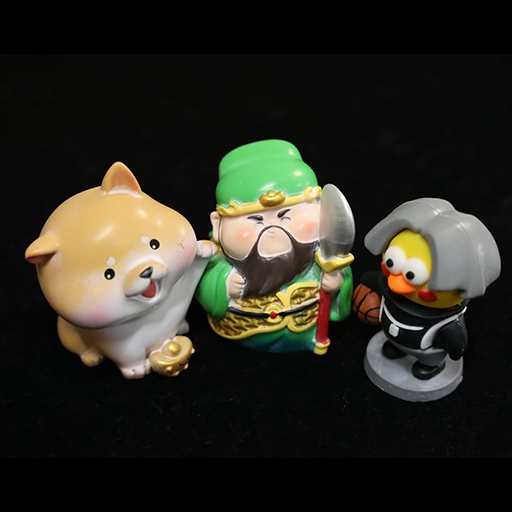} &
\includegraphics[width=0.23\linewidth]{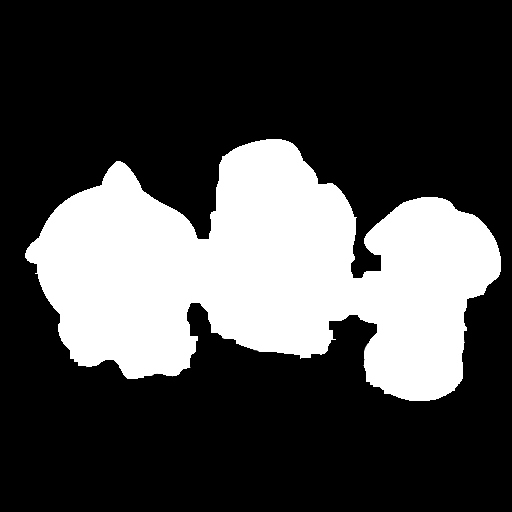} &
\includegraphics[width=0.23\linewidth]{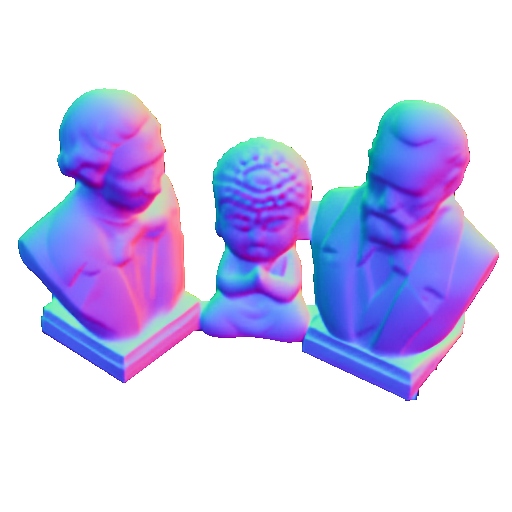} \\
\end{tabular}
\caption{Representative ROI extraction and surface normal prediction results using SAM~2~\cite{ravi2024sam2}. The first three rows are single-object cases, and the last two rows are multi-object cases.}
\label{fig:sam_roi}
\endgroup
\end{figure}

\paragraph{Automatic ROI acquisition with SAM~2~\cite{ravi2024sam2}}
We further add practical experiments on automatic ROI acquisition using SAM~2~\cite{ravi2024sam2}. As shown in Fig.~\ref{fig:sam_roi}, five representative cases, including three single-object examples and two multi-object scenes, demonstrate that the predicted masks can be converted into valid ROIs and lead to coherent surface normal predictions. These examples also show that the method can handle multi-object scenes when the target regions are segmented correctly, while mask failures may still reduce robustness in cluttered cases.

\begin{figure}[!t]
\renewcommand{\imgsize}{0.18}
\begin{adjustbox}{max width=\linewidth}
\addtolength{\tabcolsep}{-0.4em}
\begin{NiceTabular}{cccc}
    {In-the-wild Image} &
    {Delighted Image} & 
    {\makecell{FlashNormal~(Ours)}} &
    {Integrated Mesh}\\
    \includegraphics[width=\imgsize\textwidth]{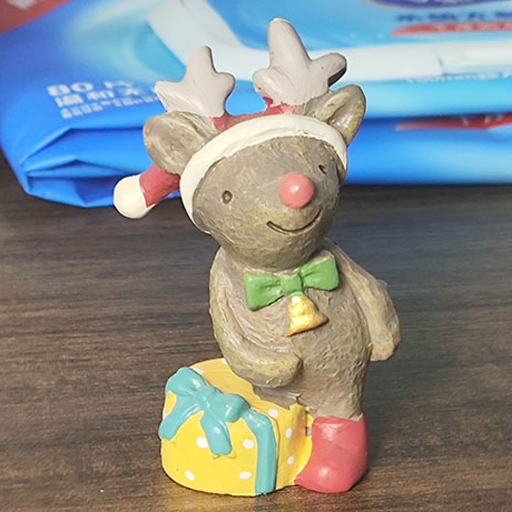} &
    \includegraphics[width=\imgsize\textwidth]{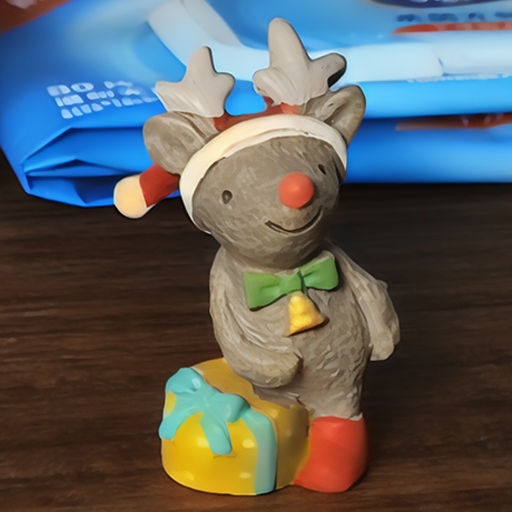} &
    \includegraphics[width=\imgsize\textwidth]{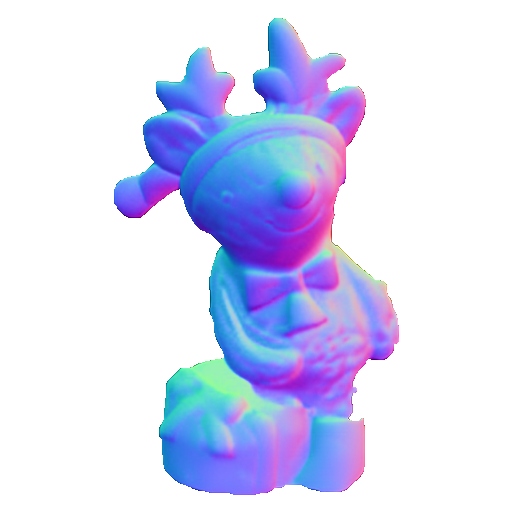} &
    \includegraphics[width=\imgsize\textwidth]{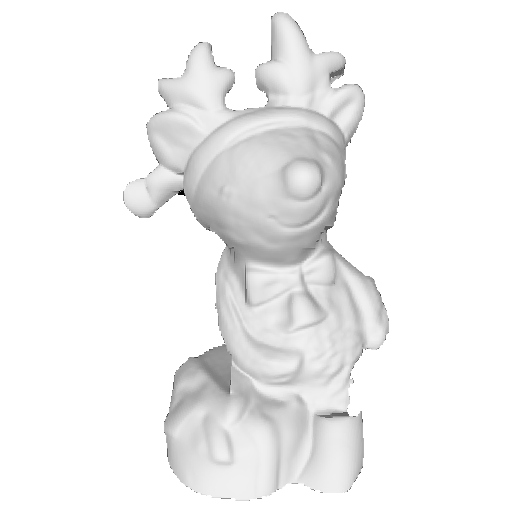} 

\end{NiceTabular}
\end{adjustbox}
\caption{Qualitative result using an in-the-wild image processed by StableDelight~\cite{stabledelightrepo}.}
\label{fig:any_image}
\end{figure}

\subsection{Suiting FlashNormal to single-image input}
While our proposal in general assumes a pair of flash/no-flash images as input, it can still work reasonably well in case of a single ambient-light image. As shown in \fref{fig:any_image}, given an ambient-light image and its algorithmically delighted version via StableDelight~\cite{stabledelightrepo}, our proposal can still estimate high-quality surface normal maps.

\subsection{Application: Multi-view normal integration for 3D reconstruction}

\begin{figure}[!t]
\renewcommand{\imgsize}{0.2}
\begin{adjustbox}{max width=\linewidth}
\addtolength{\tabcolsep}{-0.4em}
\begin{NiceTabular}{ccccccc}
    &
    \multicolumn{2}{c}{Front view} &
    \multicolumn{2}{c}{Side view} & 
    \multicolumn{2}{c}{Back view} \\
    \raisebox{2.9\height}{\rotatebox{90}{\makecell{GT}}} &
    \includegraphics[width=\imgsize\linewidth, trim=80px 2px 80px 4px, clip=true]{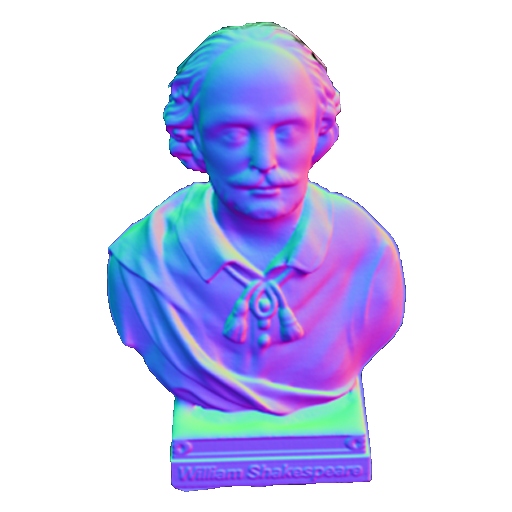} &
    \includegraphics[width=\imgsize\linewidth, trim=80px 2px 80px 4px, clip=true]{fig/application/sb_gtmesh11.jpg} & 
    \includegraphics[width=\imgsize\linewidth, trim=80px 2px 80px 4px, clip=true]{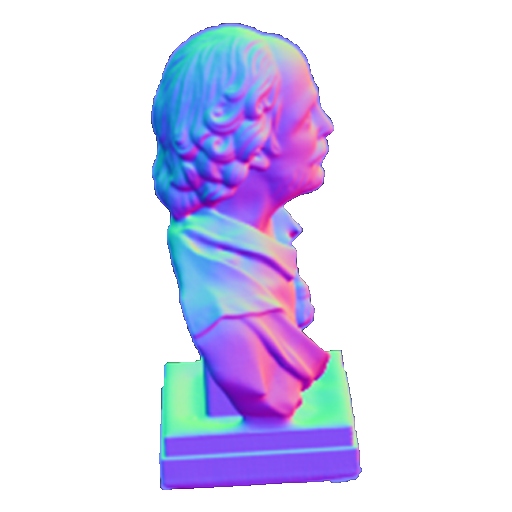} &
    \includegraphics[width=\imgsize\linewidth, trim=80px 2px 80px 4px, clip=true]{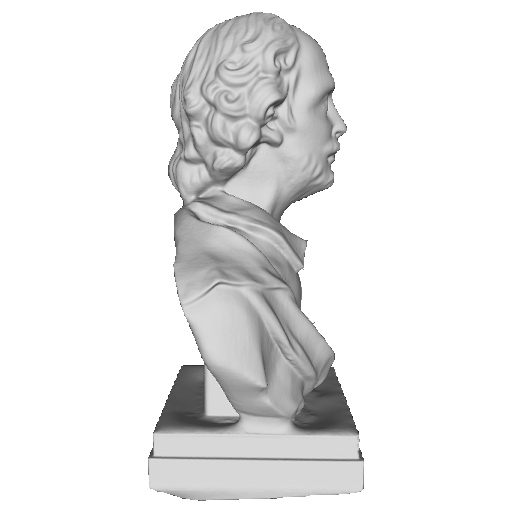}	&
    \includegraphics[width=\imgsize\linewidth, trim=80px 2px 80px 4px, clip=true]{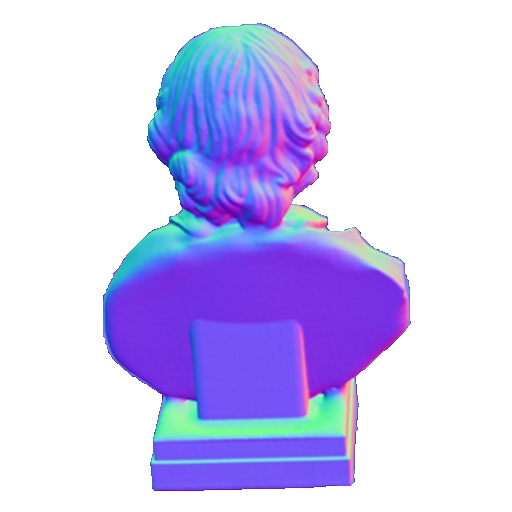} &
    \includegraphics[width=\imgsize\linewidth, trim=80px 2px 80px 4px, clip=true]{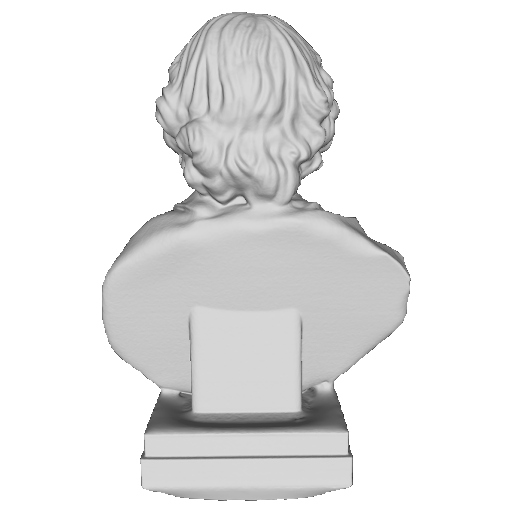} 
    \\
    
    \raisebox{0.2\height}{\rotatebox{90}{\makecell{FlashNormal\\(Ours)}}} &
    \includegraphics[width=\imgsize\linewidth, trim=80px 2px 80px 4px, clip=true]{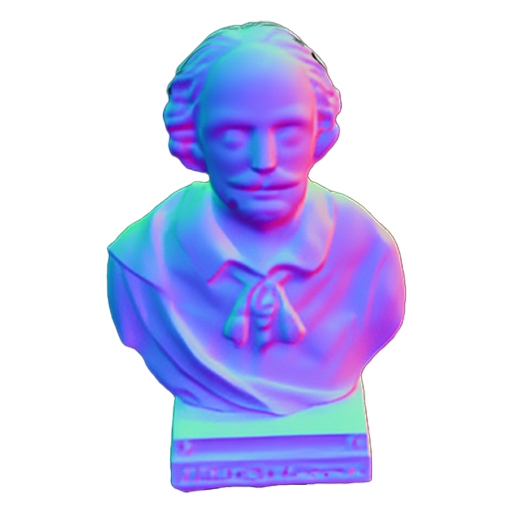} & 
    \includegraphics[width=\imgsize\linewidth, trim=80px 2px 80px 4px, clip=true]{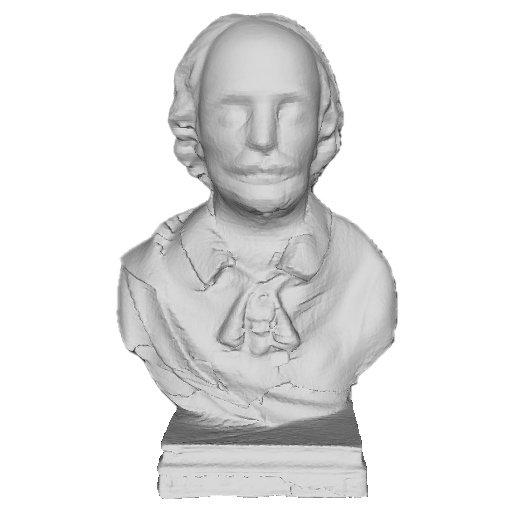} &
    \includegraphics[width=\imgsize\linewidth, trim=80px 2px 80px 4px, clip=true]{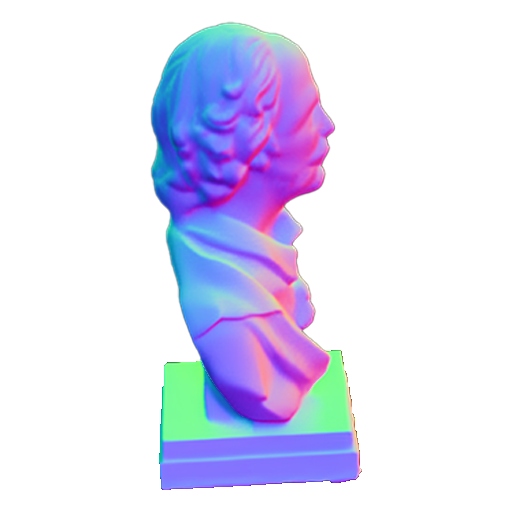}	&
    \includegraphics[width=\imgsize\linewidth, trim=80px 2px 80px 4px, clip=true]{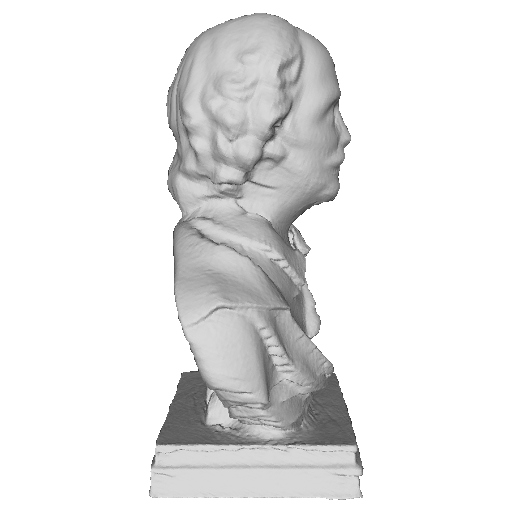} &
    \includegraphics[width=\imgsize\linewidth, trim=80px 2px 80px 4px, clip=true]{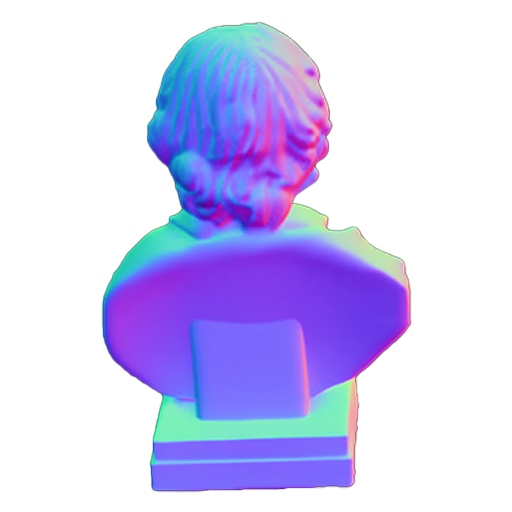} &
    \includegraphics[width=\imgsize\linewidth, trim=80px 2px 80px 4px, clip=true]{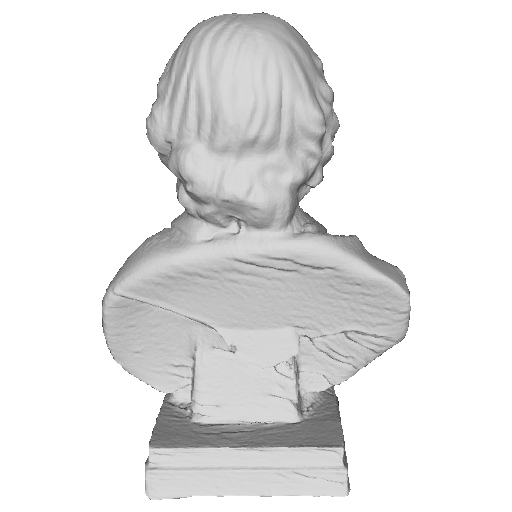} \\
    &MAE=15.30 &&MAE=18.50 &&MAE=14.41 \\
    
    \raisebox{0.15\height}{\rotatebox{90}{\makecell{E2E-FT\cite{martingarcia2024diffusione2eft}}}} &
    \includegraphics[width=\imgsize\linewidth, trim=80px 2px 80px 4px, clip=true]{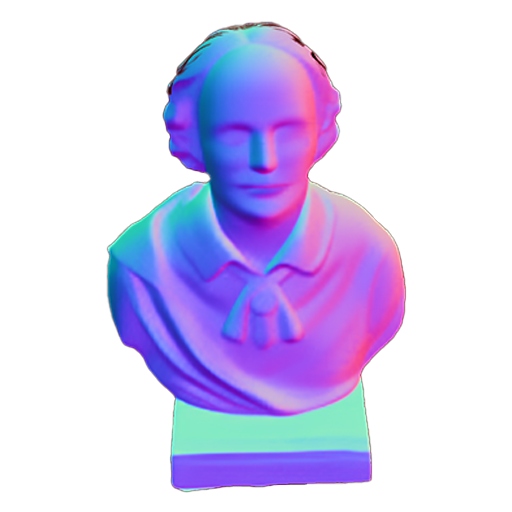} & 
    \includegraphics[width=\imgsize\linewidth, trim=80px 2px 80px 4px, clip=true]{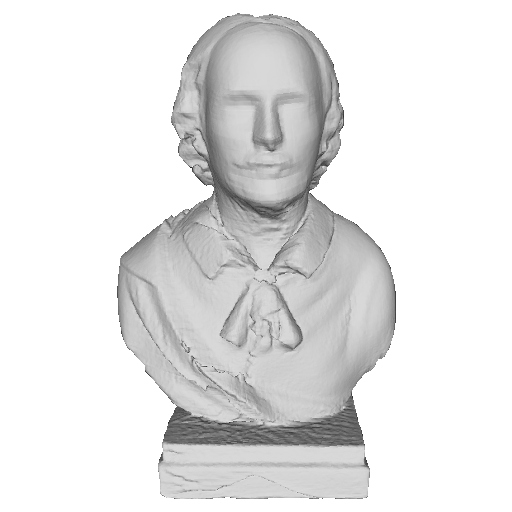} &
    \includegraphics[width=\imgsize\linewidth, trim=80px 2px 80px 4px, clip=true]{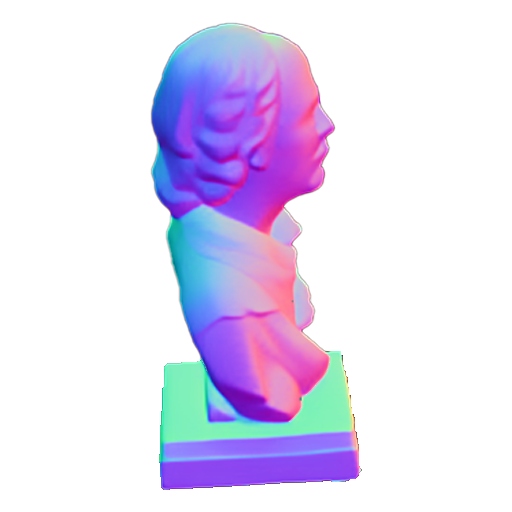} &
    \includegraphics[width=\imgsize\linewidth, trim=80px 2px 80px 4px, clip=true]{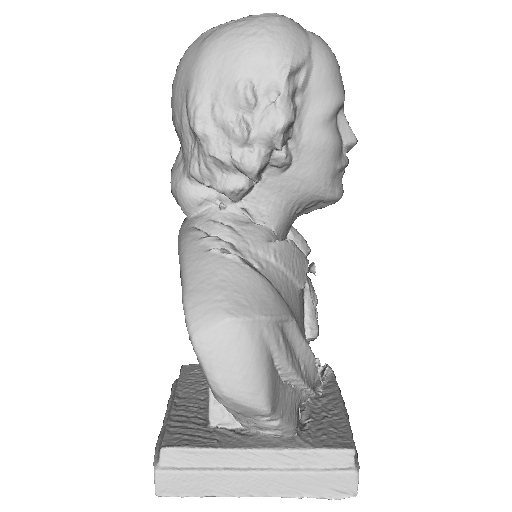} &
    \includegraphics[width=\imgsize\linewidth, trim=80px 2px 80px 4px, clip=true]{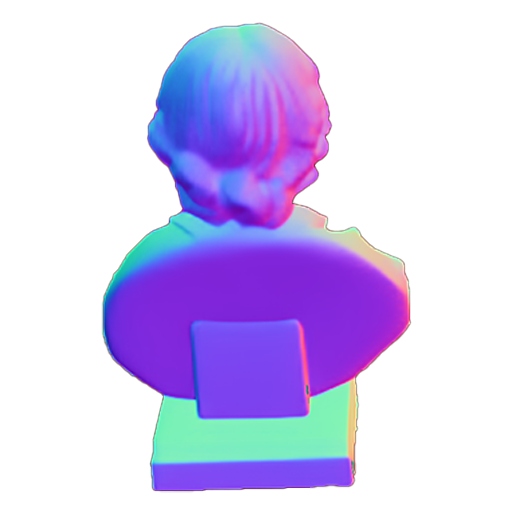} &
    \includegraphics[width=\imgsize\linewidth, trim=80px 2px 80px 4px, clip=true]{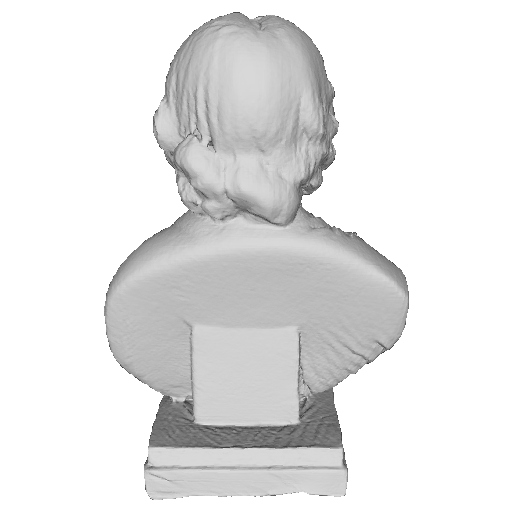} \\
    &MAE=20.73 &&MAE=21.93 &&MAE=17.02 \\
    \hdashline
    \raisebox{2.2\height}{\rotatebox{90}{\makecell{GT}}} &
    \includegraphics[width=\imgsize\linewidth, trim=60px 2px 60px 4px, clip=true]{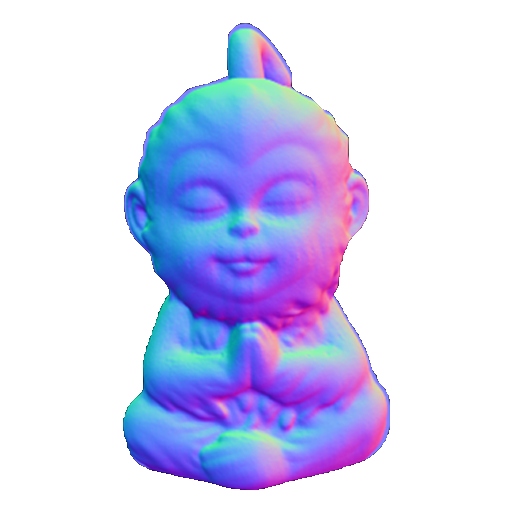} &
    \includegraphics[width=\imgsize\linewidth, trim=60px 2px 60px 4px, clip=true]{fig/application/mb_gtmesh11.jpg} & 
    \includegraphics[width=\imgsize\linewidth, trim=50px 2px 40px 4px, clip=true]{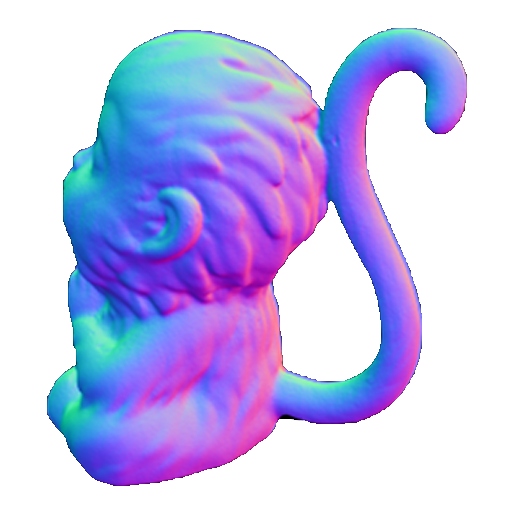} &
    \includegraphics[width=\imgsize\linewidth, trim=50px 2px 40px 4px, clip=true]{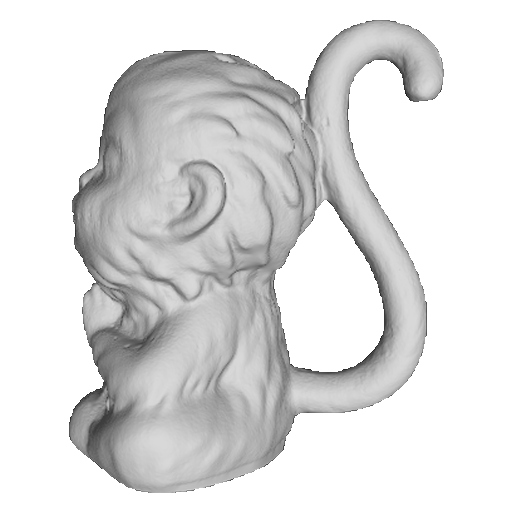}	&
    \includegraphics[width=\imgsize\linewidth, trim=60px 2px 60px 4px, clip=true]{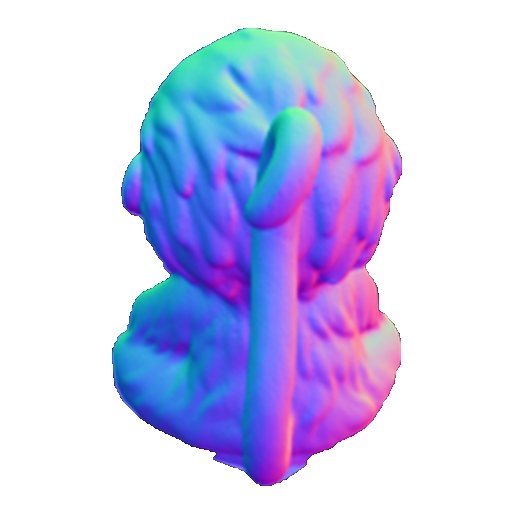} &
    \includegraphics[width=\imgsize\linewidth, trim=60px 2px 60px 4px, clip=true]{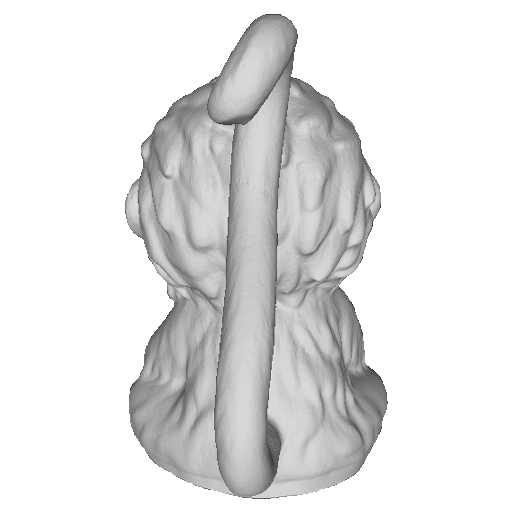} 
    \\
    
    \raisebox{0.2\height}{\rotatebox{90}{\makecell{FlashNormal\\(Ours)}}} &
    \includegraphics[width=\imgsize\linewidth, trim=60px 2px 60px 4px, clip=true]{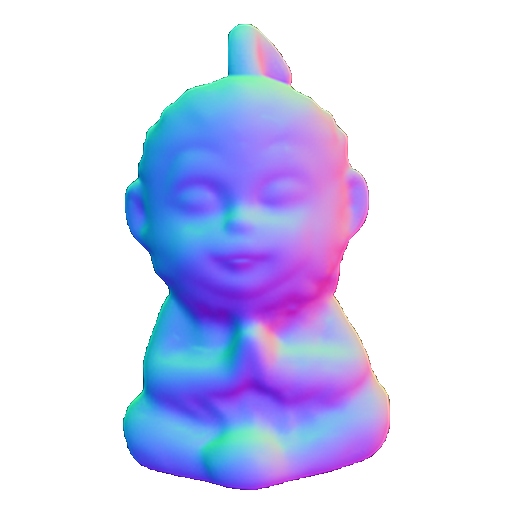} & 
    \includegraphics[width=\imgsize\linewidth, trim=60px 2px 60px 4px, clip=true]{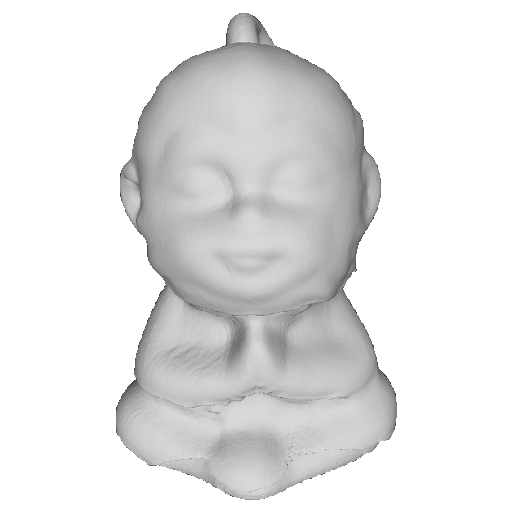} &
    \includegraphics[width=\imgsize\linewidth, trim=50px 2px 40px 4px, clip=true]{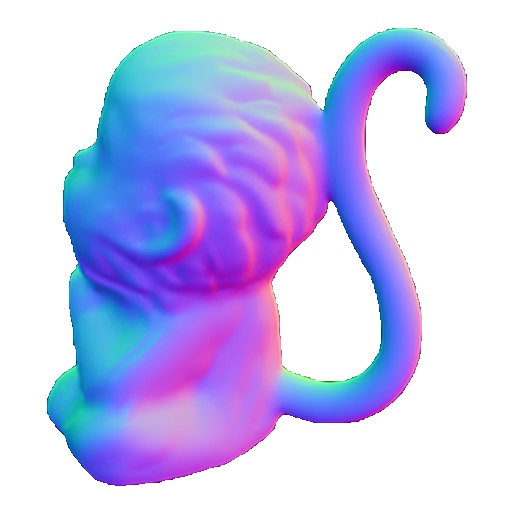}	&
    \includegraphics[width=\imgsize\linewidth, trim=50px 2px 40px 4px, clip=true]{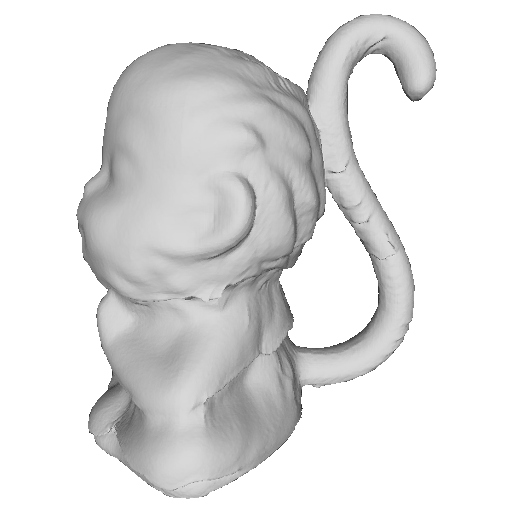} &
    \includegraphics[width=\imgsize\linewidth, trim=60px 2px 60px 4px, clip=true]{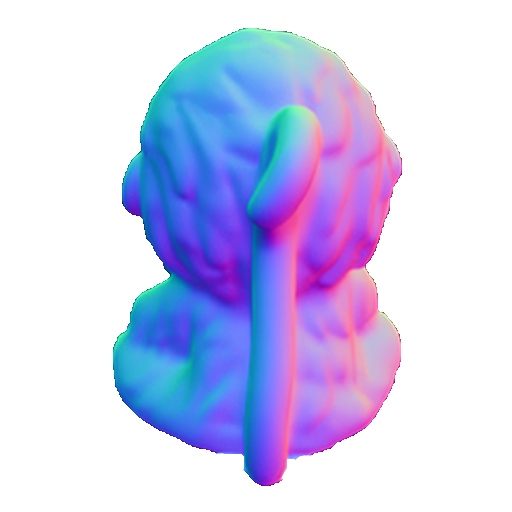} &   
    \includegraphics[width=\imgsize\linewidth, trim=60px 2px 60px 4px, clip=true]{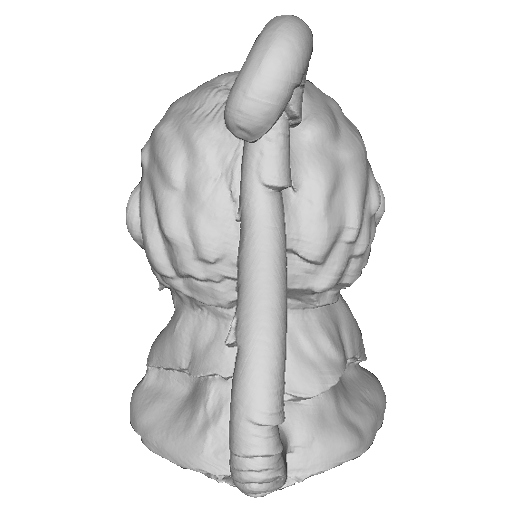} \\
    & MAE=12.76 && MAE=11.15 && MAE=14.16 \\
    
    \raisebox{0.0\height}{\rotatebox{90}{\makecell{\ee}}} &
    \includegraphics[width=\imgsize\linewidth, trim=60px 2px 60px 4px, clip=true]{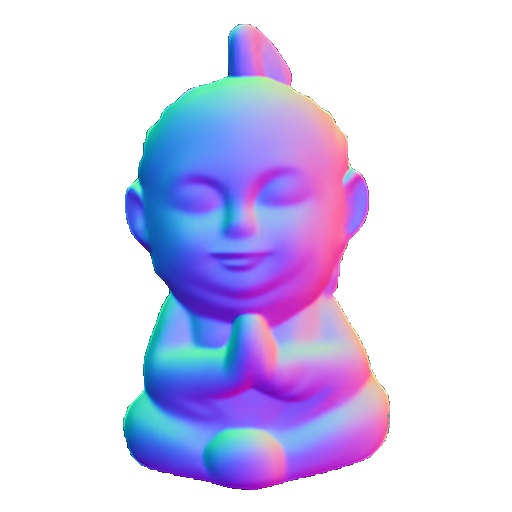} & 
    \includegraphics[width=\imgsize\linewidth, trim=60px 2px 60px 4px, clip=true]{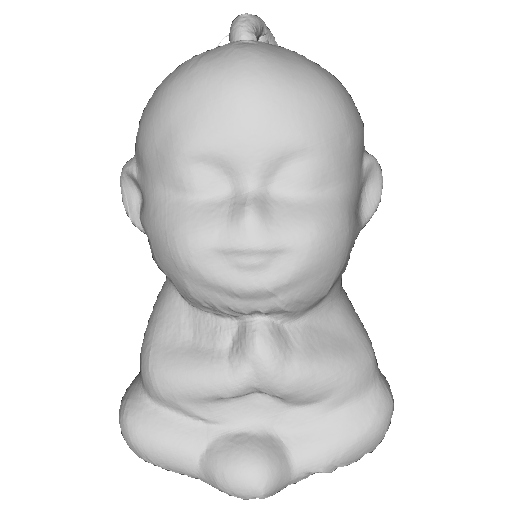} &
    \includegraphics[width=\imgsize\linewidth, trim=50px 2px 40px 4px, clip=true]{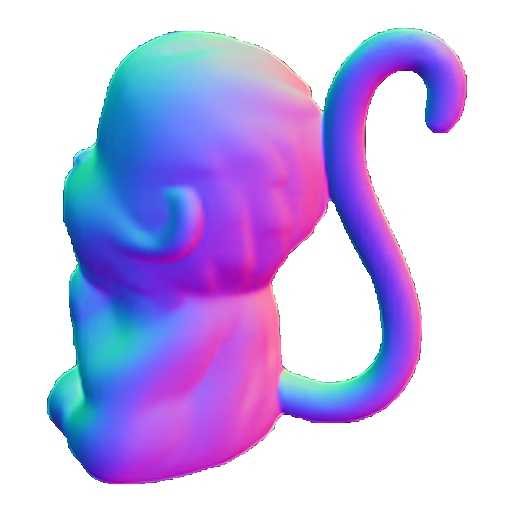} &
    \includegraphics[width=\imgsize\linewidth, trim=50px 2px 40px 4px, clip=true]{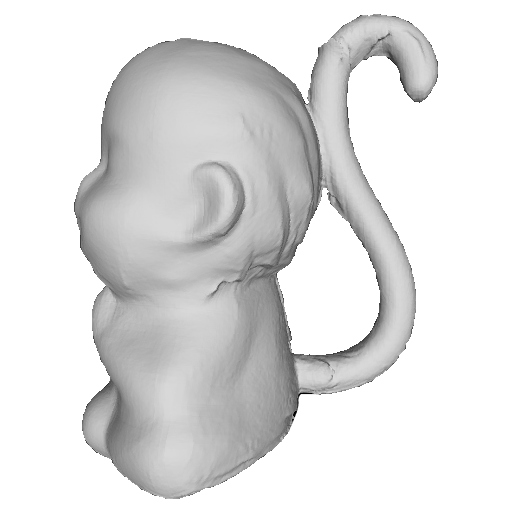} &
    \includegraphics[width=\imgsize\linewidth, trim=60px 2px 60px 4px, clip=true]{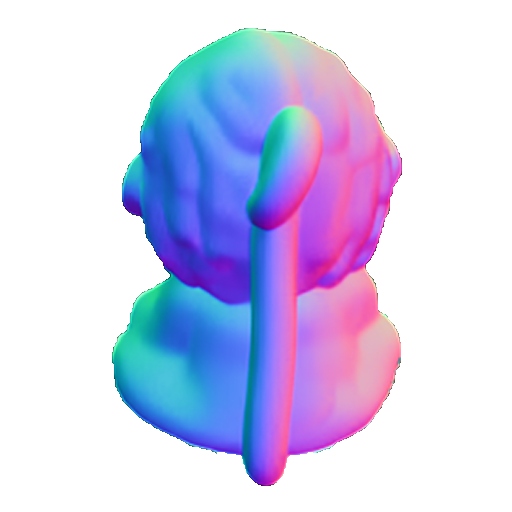} &
    \includegraphics[width=\imgsize\linewidth, trim=60px 2px 60px 4px, clip=true]{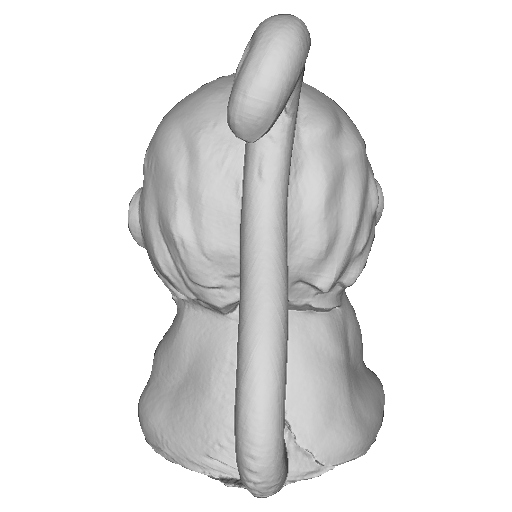} \\
    & MAE=17.44 && MAE=16.58 && MAE=20.14 
    
\end{NiceTabular}
\end{adjustbox}
\caption{3D reconstruction via integrating multi-view surface normal maps estimated by \ee and ours. The 3D mesh visualizations are taken as snapshots.}
\label{fig:app}
\end{figure}

Beyond single-view surface normal estimation, our method excels in multi-view normal integration for complete 3D reconstruction. As shown in \fref{fig:app}, we apply our method to two real-world objects, {\sc Shakespeare} and {\sc Monkey} to estimate surface normal maps of $17$ distinct views. These maps are integrated using SuperNormal~\cite{supernormal2024cao}, yielding detailed, geometrically consistent meshes. Compared to \ee, for {\sc Shakespeare}, our method produces richer geometric features and a more complete reconstruction of its head. Additionally, using the same method as EvalFlash, we generate $17$ GT normal maps and compute an average multi-view MAE of $17.62$, outperforming \ee~($19.51$) by $9.7\%$. Similarly, for {\sc Monkey}, our estimated surface normals maintain greater detail and geometric consistency across views, yielding reconstructed meshes with richer geometric features compared to \ee, whose results are smoother. We also compute the average multi-view MAE for {\sc Monkey}, achieving an MAE of $12.86$, compared to \ee's $17.83$, reflecting a $27.8\%$ improvement. These results demonstrate the superior geometric consistency, stability, and overall quality of 3D reconstruction produced by our method. 

\section{Limitations and Future Works}
\begin{figure}[!t]
\centering
\begin{adjustbox}{max width=\linewidth}
\addtolength{\tabcolsep}{-0.4em}
\renewcommand{\imgsize}{0.4\linewidth}
\begin{NiceTabular}{cccc}
    & No-flash input & Flash input & FlashNormal~(Ours)\\
    \raisebox{1.0\height}{\rotatebox{90}{\sc Rabbit}} &
    \includegraphics[width=\imgsize]{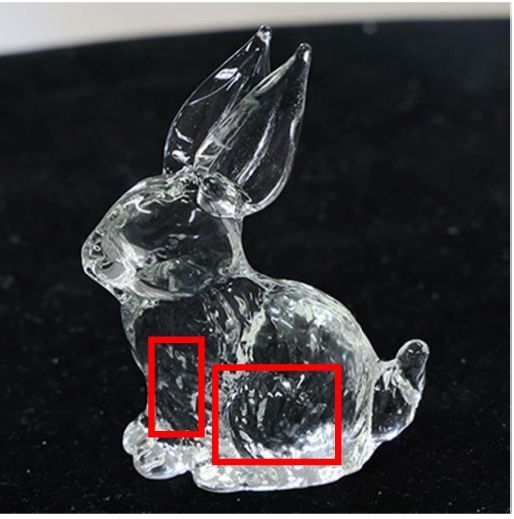} &
    \includegraphics[width=\imgsize]{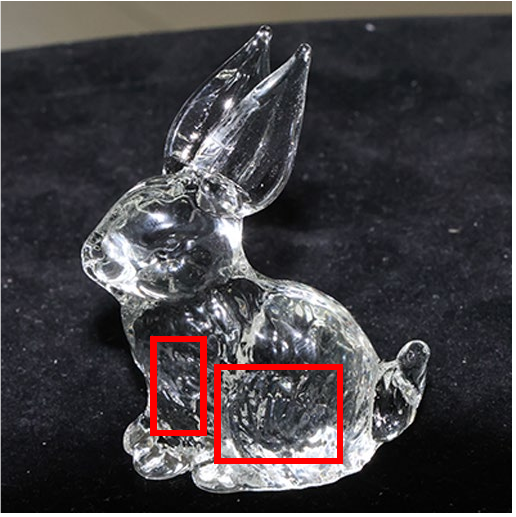} &
    \includegraphics[width=\imgsize]{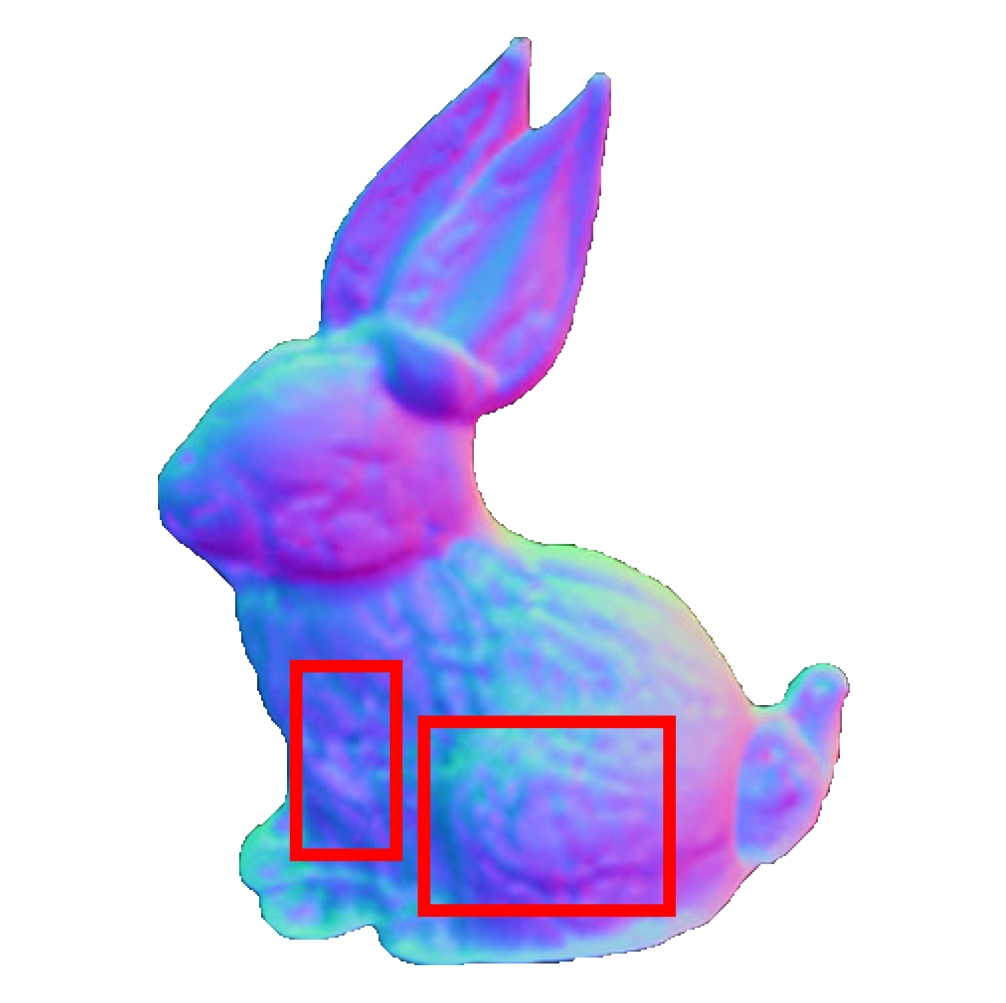} 
    \\
    \raisebox{1.9\height}{\rotatebox{90}{\sc Owl}} &
    \includegraphics[width=\imgsize]{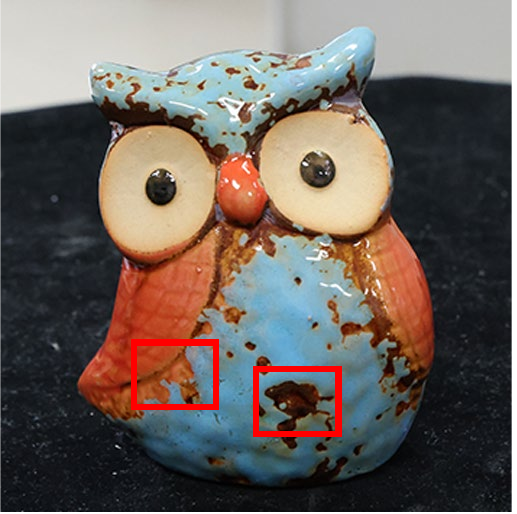} &
    \includegraphics[width=\imgsize]{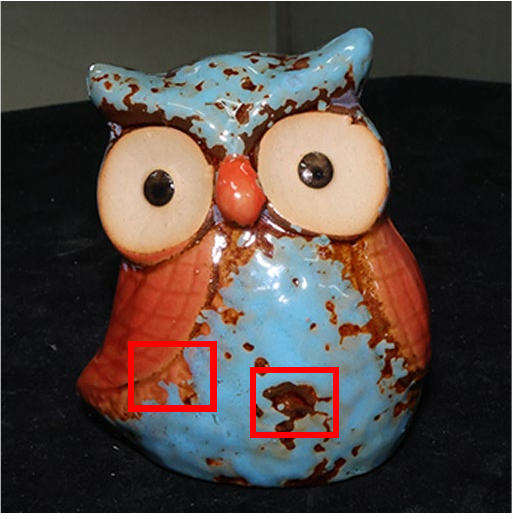} &
    \includegraphics[width=\imgsize]{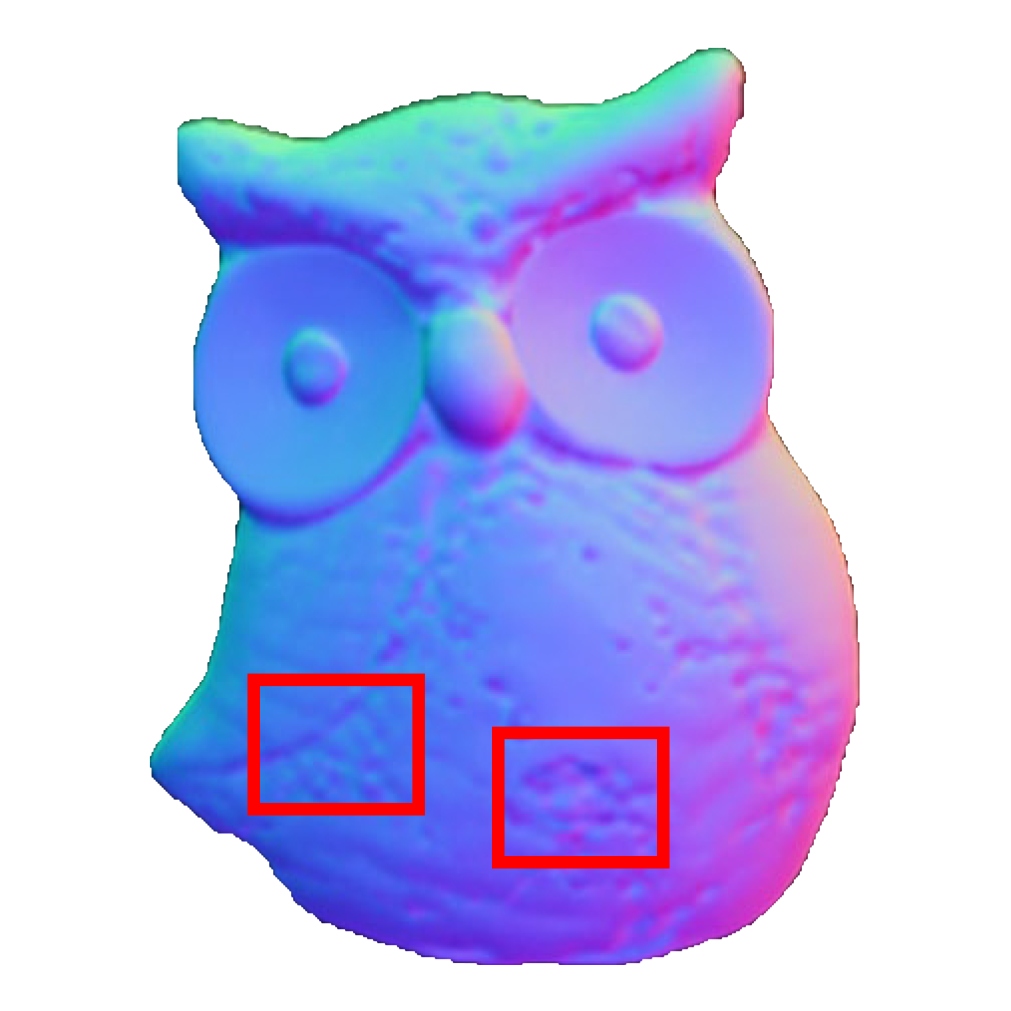} 
\end{NiceTabular}
\end{adjustbox}
\caption{Failure cases from our proposal. Red boxes highlight areas where surface normals are incorrect.}
\label{fig:failure_case}
\end{figure}

Although our proposal can predict surface normals with rich details and reduce shape-reflectance ambiguity in general, it still struggles in certain extreme cases. For example, as shown in the top row of \fref{fig:failure_case}, our method shows substandard performance when dealing with transparent objects due to their unique reflective properties, which make it difficult to clearly distinguish the surface of the object. Additionally, the relatively low number of transparent objects in our training dataset limits the prior knowledge our method has about the GT surface normals of such objects. Furthermore, as shown in the bottom row of \fref{fig:failure_case}, given flat objects with highly complex and realistic textures, our approach tends to misinterpret them as 3D. This occurs because, after introducing flash input, the shading variations in textured areas become minimal, making it difficult to distinguish textures from actual surface structures. In future work, we aim to address these limitations by augmenting the training dataset with more diverse data.

In the future, we will explore surface normal estimation for objects with special materials, as discussed earlier. We also plan to extend the flash/no-flash setup to scene-level applications, and study its effects in complex environments. Additionally, we will investigate how to leverage this setup to improve the quality and temporal consistency of surface normal estimation in dynamic scenes.

\section{Conclusion}

We introduce \ours, a robust and practical surface normal estimator leveraging flash/no-flash image pairs. By integrating diffusion priors, curvature-guided geometry enhancement, and the zoomed-pixels strategy, \ours achieves fine-grained normal estimation while effectively mitigating shape-reflectance ambiguity. To support training and evaluation, we provide \traindata, a large-scale photorealistic dataset, and \ourdata, the first real-world dataset tailored to flash/no-flash-based surface normal estimation. Extensive experiments demonstrate that \ours outperforms state-of-the-art methods, delivering significant improvements in surface normal estimation accuracy and downstream multi-view 3D reconstruction quality.

\section*{Acknowledgment}
This work was supported by Beijing Major Science and Technology Project No. Z251100007125021, Hebei Natural Science Foundation Project No. 242Q0101Z, Beijing-Tianjin-Hebei Basic Research Funding Program No. F2024502017, and the National Natural Science Foundation of China (Grant Nos. 62472044, U24B20155, and U23B2052).

\bibliographystyle{IEEEtran}
\bibliography{tcsvt}

\begin{IEEEbiography}[{\includegraphics[width=1in,height=1.25in,clip,keepaspectratio]{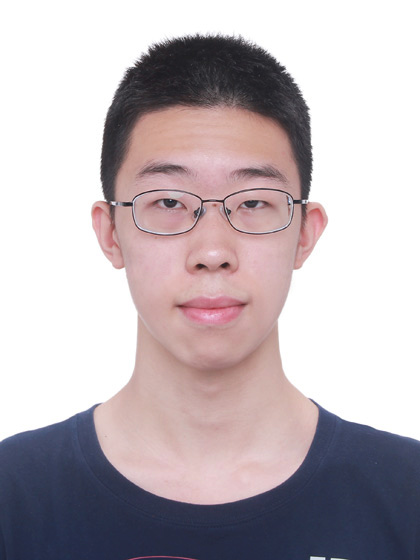}}]{Ruiyang Chen}
		received the B.S degree in computer science and technology from Beijing University of Technology, Beijing, China, in 2025. He is currently pursuing the M.S. degree with the College of Artificial Intelligence, Beijing University of Posts and Telecommunications. His research interests include deep learning and computer vision.
\end{IEEEbiography}
\vspace{-10pt}
\begin{IEEEbiography}[{\includegraphics[width=1in,height=1.25in,clip,keepaspectratio]{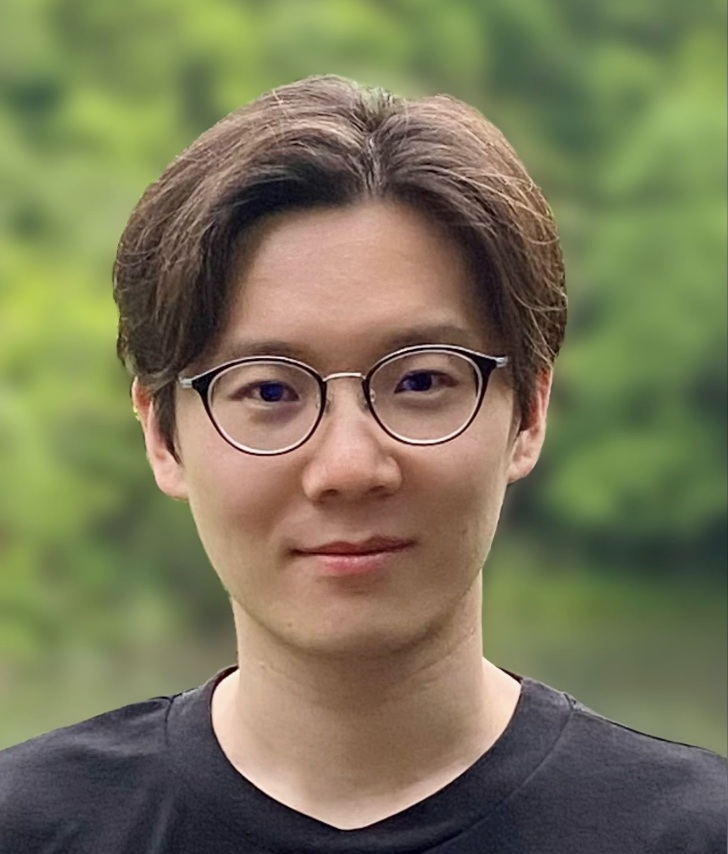}}]{Feiran Li}
	received his Ph.D. in Information Science from Osaka University in 2023. Before that, He obtained M.Eng. in Robotics from Nara Institute of Science $\&$ Technology in 2019, and B.Eng. in Mechanical Engineering from East China University of Science $\&$ Technology in 2017. His research interests lie in computational photography and geometric computer vision.
\end{IEEEbiography}
\vspace{-10pt}
\begin{IEEEbiography}[{\includegraphics[width=1in,height=1.25in,clip,keepaspectratio]{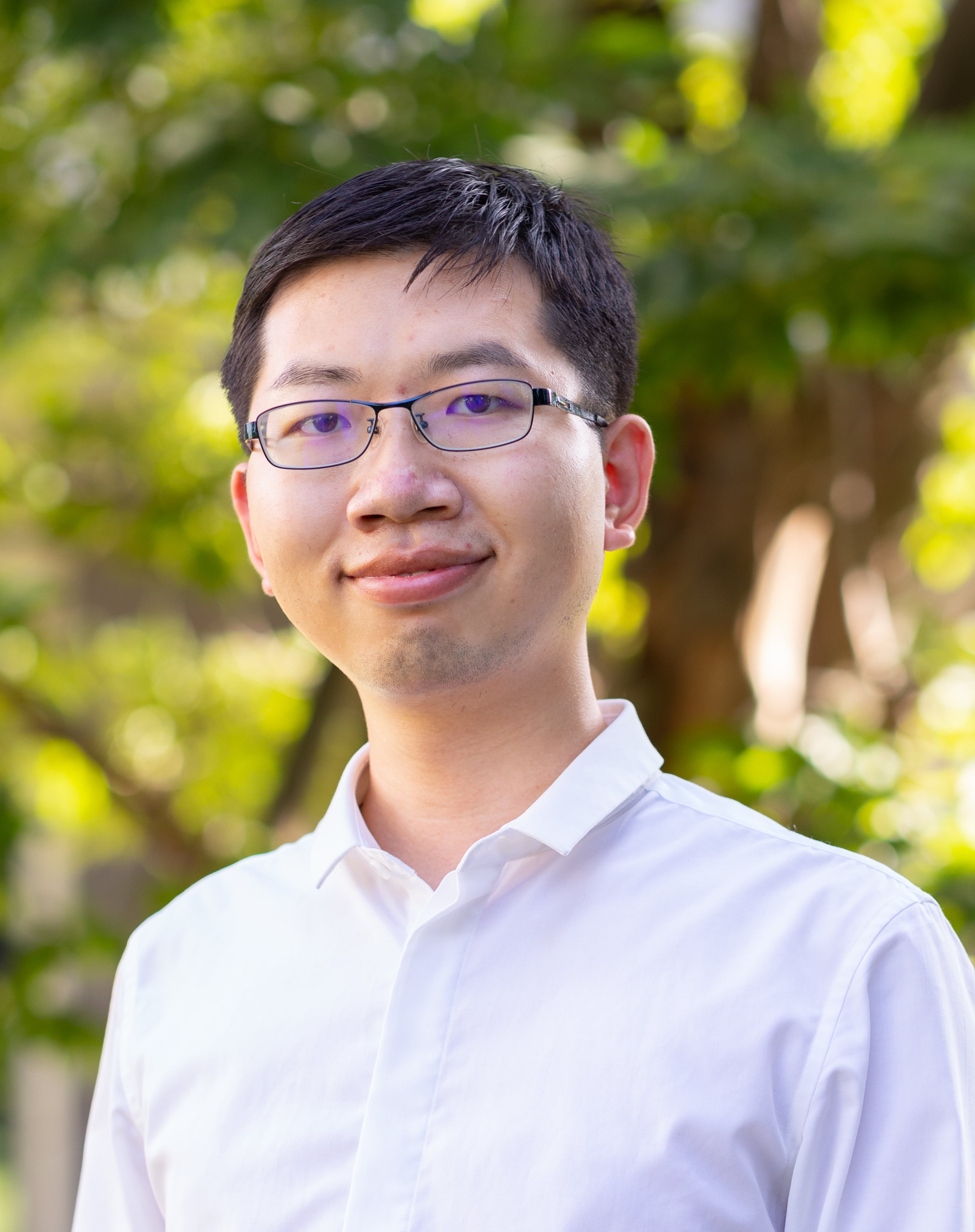}}]{Heng Guo}
	(Member, IEEE) received the BE and ME degrees from University of Electronic Science and Technology, and the PhD degree from Osaka University, in 2015, 2018, and 2022. He is currently a specially-appointed research Professor at Beijing University of Posts and Telecommunications~(BUPT). Before joining BUPT, he was a specially-appointed assistant professor at Osaka University from 2022 to 2023. His research interest includes computational photography, physics-based computer vision, and 3D reconstruction.
\end{IEEEbiography}
\vspace{-10pt}
\begin{IEEEbiography}[{\includegraphics[width=1in,height=1.25in,clip,keepaspectratio]{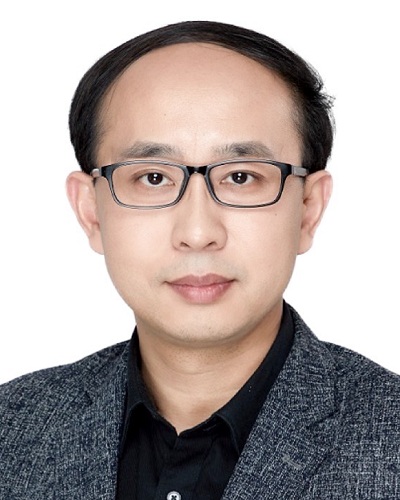}}]{Zhanyu Ma}
	(Senior Member, IEEE) received the Ph.D. degree in electrical engineering from
	the KTH Royal Institute of Technology, Sweden, in 2011. He has been a Professor with Beijing University of Posts and Telecommunications, Beijing, China, since 2019. From 2012 to 2013, he was a Post-Doctoral Research Fellow with the School of Electrical Engineering, KTH. He was an Associate Professor with Beijing	University of Posts and Telecommunications,	Beijing, China, from 2014 to 2019. His research interests include pattern recognition and machine learning fundamentals, with a focus on applications in computer vision and multimedia signal processing.
\end{IEEEbiography}
\end{document}